\documentclass[11pt]{article}
\usepackage[utf8]{inputenc}
\usepackage{xcolor}
\usepackage[top=1.0in, bottom=1.0in, left=1.0in, right=1.0in]{geometry}
\usepackage{enumitem}
\usepackage{graphicx}
\usepackage{tabularx}
\usepackage{tipa}
\usepackage{multirow}
\usepackage{appendix}
\usepackage{longtable}
\usepackage{float}
\usepackage{wrapfig}
\usepackage{titlesec}
\usepackage{xspace}
\usepackage{amsmath}
\usepackage[T1]{fontenc}
\usepackage{comment}
\usepackage[backend=biber,style=numeric,sorting=none]{biblatex}
\usepackage{xurl}

\usepackage[colorlinks = true,  
            linkcolor = blue,
            urlcolor  = blue,
            citecolor = blue,
            anchorcolor = blue]{hyperref}

\usepackage[acronym,nonumberlist,nogroupskip,nopostdot,style=long]{glossaries}
\makeglossaries

\newacronym{NSF}{NSF}{National Science Foundation}
\newacronym{AI}{AI}{artificial intelligence}
\newacronym{MPS}{MPS}{(NSF Directorate for) Mathematical and Physical Sciences}
\newacronym{AST}{AST}{(NSF Division of) Astronomical Sciences}
\newacronym{CHE}{CHE}{(NSF Division of) Chemistry}
\newacronym{DMR}{DMR}{(NSF Division of) Materials Research}
\newacronym{DMS}{DMS}{(NSF Division of) Mathematical Sciences}
\newacronym{PHY}{PHY}{(NSF Division of) Physics}
\newacronym{STEM}{STEM}{science, technology, engineering, and mathematics}
\newacronym{ML}{ML}{machine learning}
\newacronym{DFT}{DFT}{density functional theory}
\newacronym{CIF}{CIF}{Crystallographic Information Framework}
\newacronym{ICSD}{ICSD}{Inorganic Crystal Structure Database}
\newacronym{PDB}{PDB}{Protein Data Bank}
\newacronym{GAN}{GAN}{generative adversarial network}
\newacronym{VAE}{VAE}{variational autoencoder}
\newacronym{NF}{NF}{normalizing flow}
\newacronym{SBI}{SBI}{simulation-based inference}
\newacronym{LLM}{LLM}{large language model}
\newacronym{AGI}{AGI}{artificial general intelligence}
\newacronym{RL}{RL}{reinforcement learning}
\newacronym{SAT}{SAT}{Boolean satisfiability problem}
\newacronym{SMT}{SMT}{satisfiability modulo theories}
\newacronym{SciML}{SciML}{scientific machine learning}
\newacronym{HNN}{HNN}{Hamiltonian neural network}
\newacronym{LNN}{LNN}{Lagrangian neural network}
\newacronym{DOE}{DOE}{Department of Energy}
\newacronym{MoDL}{MoDL}{Mathematical and Scientific Foundations of Deep Learning}
\newacronym{NITMB}{NITMB}{National Institute for Theory and Mathematics in Biology}
\newacronym{CSSI}{CSSI}{Cyberinfrastructure for Sustained Scientific Innovation}
\newacronym{CISE}{CISE}{(NSF Directorate for) Computer and Information Science and Engineering}
\newacronym{IAIFI}{IAIFI}{Institute for Artificial Intelligence and Fundamental Interactions}
\newacronym{SGD}{SGD}{stochastic gradient descent}
\newacronym{NAIRR}{NAIRR}{National AI Research Resource}
\newacronym{API}{API}{application programming interface}
\newacronym{MaaS}{MaaS}{Model as a Service}
\newacronym{GPU}{GPU}{graphics processing unit}
\newacronym{HPC}{HPC}{high-performance computing}
\newacronym{DLM}{DLM}{data lifecycle management}
\newacronym{LHC}{LHC}{Large Hadron Collider}
\newacronym{PINN}{PINN}{physics-informed neural network}
\newacronym{UQ}{UQ}{uncertainty quantification}
\newacronym{CS}{CS}{computer science}
\newacronym{RAG}{RAG}{retrieval augmented generation}
\newacronym{RAGA}{RAGA}{retrieval augmented generation assessment}
\newacronym{GNN}{GNN}{graphic neural network}
\newacronym{ODE}{ODE}{ordinary differential equation}
\newacronym{ADS}{ADS}{Astrophysics Data System}
\newacronym{CDS}{CDS}{Strasbourg astronomical Data Center}
\newacronym{GRMHD}{GRMHD}{general-relativistic magnetohydrodynamic}
\newacronym{MLIP}{MLIP}{machine-learned interatomic potential}
\newacronym{KG}{KG}{knowledge graph}
\newacronym{DNN}{DNN}{deep neural network}
\newacronym{PDE}{PDE}{partial differential equation}
\newacronym{SDE}{SDE}{stochastic differential equation}
\newacronym{MFAI}{MFAI}{Mathematical Foundations of AI}
\newacronym{RMT}{RMT}{random matrix theory}
\newacronym{NTK}{NTK}{neural tangent kernel}
\newacronym{RKHS}{RKHS}{reproducing kernel Hilbert space}
\newacronym{TDA}{TDA}{topological data analysis}
\newacronym{LIGO}{LIGO}{Laser Interferometer Gravitational-wave Observatory}
\newacronym{PNT}{PNT}{positioning, navigation, and timing}
\newacronym{EIC}{EIC}{Electron-Ion Collider}
\newacronym{NN}{NN}{neural network}
\newacronym{QFT}{QFT}{quantum field theory}
\newacronym{LISA}{LISA}{Laser Interferometer Space Antenna}
\newacronym{HL-LHC}{HL-LHC}{High-luminosity LHC}
\newacronym{DUNE}{DUNE}{Deep Underground Neutrino Experiment}
\newacronym{NISQ}{NISQ}{noisy intermediate-scale quantum}
\newacronym{BDT}{BDT}{boosted decision tree}
\newacronym{CNN}{CNN}{convolutional neural network}
\newacronym{NASA}{NASA}{National Aeronautics and Space Administration}
\newacronym{CCAS}{CCAS}{Center for Computer-Assisted Synthesis}
\newacronym{LoRA}{LoRA}{Low Rank Adaptation}
\newacronym{PI}{PI}{principal investigator}
\newacronym{NeurIPS}{NeurIPS}{Conference on Neural Information Processing Systems}
\newacronym{DARPA}{DARPA}{Defense Advanced Research Projects Agency}
\newacronym{RSE}{RSE}{Research Software Engineer}

\DeclareRobustCommand{\Sec}[1]{Sec.~\ref{#1}}
\DeclareRobustCommand{\Secs}[2]{Secs.~\ref{#1} \& \ref{#2}}
\DeclareRobustCommand{\Secss}[3]{Secs.~\ref{#1}, \ref{#2} \& \ref{#3}}
\DeclareRobustCommand{\Secsss}[4]{Secs.~\ref{#1}, \ref{#2}, \ref{#3}, \& \ref{#4}}

\begin{document}

\begin{center}
    {\LARGE \textbf{The Future of Artificial Intelligence and the \\Mathematical and Physical Sciences (AI+MPS)} \\} ~\\

\large \textit{Community Paper from the NSF Future of AI+MPS Workshop \\ Cambridge, Massachusetts --- March 24--26, 2025}
\end{center} 

\vspace{12pt}

\renewcommand{\thefootnote}{\fnsymbol{footnote}}

\noindent \textbf{Organizers:}

Andrew Ferguson (Materials Research, University of Chicago)

Marisa LaFleur\footnotemark[1] (MIT) 

Lars Ruthotto (Mathematical Sciences, Emory University)

Jesse Thaler\footnotemark[1] (Physics, MIT) 

Yuan-Sen Ting (Astronomical Sciences, The Ohio State University)

Pratyush Tiwary (Chemistry, University of Maryland)

Soledad Villar (Mathematical Sciences, Johns Hopkins University)\\

\footnotetext[1]{Corresponding authors: \href{mailto:mlafleur@mit.edu}{mlafleur@mit.edu} and \href{mailto:jthaler@mit.edu}{jthaler@mit.edu}}

\renewcommand{\thefootnote}{\arabic{footnote}}

\noindent \textbf{Participants:}

E.\ Paulo Alves (University of California, Los Angeles); Jeremy Avigad (Carnegie Mellon University); Simon Billinge (Columbia University); Camille Bilodeau (University of Virginia); Keith Brown (Boston University); Emmanuel Candes (Stanford University); Arghya Chattopadhyay (University of Puerto Rico Mayaguez); Bingqing Cheng (Unversity of California, Berkeley); 
Jonathan Clausen (Strategic Analysis, Inc.); Connor Coley (MIT); Andrew Connolly (University of Washington); Fred Daum (Raytheon); Sijia Dong (Northeastern University); Chrisy Xiyu Du (University of Hawai`i at Mānoa); Cora Dvorkin (Harvard University); Cristiano Fanelli (William \& Mary); Eric B.\ Ford (The Pennsylvania State University); Luis Manuel Frutos (University of Alcala); Nicolás García Trillos (University of Wisconsin-Madison); Cecilia Garraffo (Center for Astrophysics | Harvard \& Smithsonian); Robert Ghrist (University of Pennsylvania); Rafael Gomez-Bombarelli (MIT DMSE); Gianluca Guadagni (University of Virginia); Sreelekha Guggilam (Texas A\&M University, Corpus Christi); Sergei Gukov (Caltech); Juan B. Gutiérrez (University of Texas at San Antonio); Salman Habib (Argonne National Laboratory); Johannes Hachmann (University at Buffalo, The State University of New York); Boris Hanin (Princeton University); Philip Harris (MIT); Murray Holland (JILA and University of Colorado Boulder); Elizabeth Holm (University of Michigan); Hsin-Yuan Huang (Caltech); Shih-Chieh Hsu (University of Washington); Nick Jackson (University of Illinois, Urbana-Champaign); Olexandr Isayev (Carnegie Mellon University); Heng Ji (University of Illinois Urbana-Champaign, Amazon Scholar); Aggelos Katsaggelos (Northwestern University); Jeremy Kepner (MIT Lincoln Laboratory Supercomputing Center); Yannis Kevrekidis (Johns Hopkins University); Michelle Kuchera (Davidson College); J.\ Nathan Kutz (University of Washingon); Branislava Lalic (University of Novi Sad); Ann Lee (Carnegie Mellon University); Matt LeBlanc (Brown University); Josiah Lim (Johns Hopkins University); Rebecca Lindsey (University of Michigan, Ann Arbor); Yongmin Liu (Northeastern University); Peter Y.\ Lu (University of Chicago); Sudhir Malik (University of Puerto Rico Mayaguez); Vuk Mandic (University of Minnesota); Vidya Manian (University of Puerto Rico, Mayaguez); Emeka P. Mazi (Georgia State University); Pankaj Mehta (Boston University); Peter Melchior (Princeton University); Brice Ménard (Johns Hopkins University); Jennifer Ngadiuba (Fermilab); Stella Offner (University of Texas at Austin); Elsa Olivetti (MIT); Shyue Ping Ong (University of California San Diego); Christopher Rackauckas (JuliaHub, Pumas-AI, MIT); Philippe Rigollet (MIT);
Chad Risko (University of Kentucky); Philip Romero (Duke University); Grant Rotskoff (Stanford University); Brett Savoie (University of Notre Dame); Uros Seljak (University of California, Berkeley); David Shih (Rutgers University); Gary Shiu (University of Wisconsin-Madison); Dima Shlyakhtenko (UCLA); Eva Silverstein (Stanford University); Taylor Sparks (University of Utah); Thomas Strohmer (University of California, Davis); Christopher Stubbs (Harvard); Stephen Thomas (AMD and Lehigh University); Suriyanarayanan Vaikuntanathan (University of Chicago); Rene Vidal (University of Pennsylvania); Francisco Villaescusa-Navarro (Flatiron Institute); Gregory Voth (The University of Chicago); Benjamin Wandelt (Johns Hopkins University); Rachel Ward (University of Texas at Austin); Melanie Weber (Harvard University); Risa Wechsler (Stanford University); Stephen Whitelam (Lawrence Berkeley National Laboratory); Olaf Wiest (University of Notre Dame);  Mike Williams (MIT); Zhuoran Yang (Yale University); Yaroslava G.\ Yingling (North Carolina State University); Bin Yu (University of California, Berkeley); Shuwen Yue (Cornell University); Ann Zabludoff (University of Arizona); Huimin Zhao (University of Illinois, Urbana-Champaign); Tong Zhang (University of Illinois, Urbana-Champaign) 

\vspace{12pt}

\noindent \textbf{Abstract:} 

This community paper developed out of the NSF Workshop on the Future of Artificial Intelligence (AI) and the Mathematical and Physics Sciences (MPS), which was held in March 2025 with the goal of understanding how the MPS domains (Astronomy, Chemistry, Materials Research, Mathematical Sciences, and Physics) can best capitalize on, and contribute to, the future of AI. We present here a summary and snapshot of the MPS community's perspective, as of Spring/Summer 2025, in a rapidly developing field. The link between AI and MPS is becoming increasingly inextricable; now is a crucial moment to strengthen the link between AI and Science by pursuing a strategy that proactively and thoughtfully leverages the potential of AI for scientific discovery and optimizes opportunities to impact the development of AI by applying concepts from fundamental science. To achieve this, we propose activities and strategic priorities that: (1) enable AI+MPS research in both directions; (2) build up an interdisciplinary community of AI+MPS researchers; and (3) foster education and workforce development in AI for MPS researchers and students. We conclude with a summary of suggested priorities for funding agencies, educational institutions, and individual researchers to help position the MPS community to be a leader in, and take full advantage of, the transformative potential of AI+MPS. 

\newpage

\setcounter{tocdepth}{3}
\tableofcontents

\newpage

\section*{Preface}
\phantomsection
\addcontentsline{toc}{section}{Preface}

\glsresetall

The workshop on the Future of \glslink{AI}{Artificial Intelligence (AI)}\glsunset{AI} and the \glslink{MPS}{Mathematical and Physical Sciences (MPS)}\glsunset{MPS}, which was held in March 2025 with support from the U.S.~\gls{NSF}, was conceived in Summer 2024. The \gls{MPS} domains have long used and developed techniques in machine learning, statistics, and data science to drive scientific innovation. Starting about a decade ago, the rise of deep learning enabled exciting new strategies to analyze scientific datasets and perform scientific calculations.  Today, the growing availability of powerful AI tools is poised to fundamentally change the ways scientists pursue groundbreaking discoveries. AI is becoming ubiquitous, in the sciences and in society more generally, and there are no signs of AI progress slowing down.  In fact, AI appears to be advancing at an ever accelerating pace, and it is essential for science to keep up, both in understanding how to take advantage of these tools and in contributing to their development.

How can the MPS domains best capitalize on, and contribute to, the future of AI? Answering this question was the goal of the workshop, which brought together MPS researchers across the Astronomical Sciences (\acrshort{AST}), Chemistry (\acrshort{CHE}), Materials Research (\acrshort{DMR}), Mathematical Sciences (\acrshort{DMS}), and Physics (\acrshort{PHY}) domains, along with experts from computer science and AI more broadly. Through online surveys and in-person conversations, attendees discussed what aspects of AI are driving scientific innovation, what barriers there are to wider use of AI by MPS researchers, what existing AI capabilities are being underutilized in our fields, and what new AI capabilities are needed to further advance the sciences.
The workshop highlighted the ways that MPS researchers are contributing to the ``Science of AI,'' which can lead to an improved understanding of how AI systems work and subsequently help AI scale beyond current capabilities. 
Another key topic of the workshop was how AI is changing the way we need to educate and train the next generation of MPS researchers, and how MPS researchers can best ensure that AI is considered reliable, both in scientific research and in broader societal applications. This document is a summary and snapshot of our thinking, as of Spring/Summer 2025, in a rapidly developing field. The initial draft was put together based on pre-workshop surveys and the discussions at the workshop, but subsequent drafts have incorporated feedback from workshop attendees and the community at large.

This report also highlights the past achievements and tremendous opportunities to continue and strengthen the longstanding partnership between federal funding agencies and academic research institutions in the United States. For many decades, federal agencies like the \gls{NSF}, \gls{DOE}, and \gls{NASA} have been essential supporters of scientific research, in the MPS domains and beyond. Recent investments in programs like the NSF AI Institutes have contributed significantly to the adoption and development of AI techniques.
U.S.\ leadership in AI is being propelled through expertise and discoveries from foundational research---particularly through academic, national lab, and industry partnerships---as well as through the rapid scaling and commercialization of AI by for-profit companies.
Basic research funding and training have produced an AI workforce that has established the U.S. as a hub for these for-profit companies, with a deep pool of talent from which to draw. It is through the pursuit of cutting-edge AI research by early-career scientists that we can continue to cultivate an AI-enabled  \acrshort{STEM} workforce. Sustained support for academic research thus is essential to maintain U.S. leadership in AI. 

\newpage

\section*{Executive Summary}
\phantomsection
\addcontentsline{toc}{section}{Executive Summary}

\glsresetall

\Gls{AI} is a rapidly growing field with pervasive applications across a variety of scientific disciplines. The link between AI and the Mathematical and Physical Sciences (\acrshort{MPS}) is becoming increasingly inextricable, as evidenced by the 2024 Nobel Prizes in Physics and Chemistry. The researchers were recognized, respectively, for pioneering work in the development of foundational methods underpinning modern AI technologies and for the applications of AI tools for computational protein design and structure prediction. These, and many other innovations, have been made possible by fusing MPS insights with AI capabilities, to the benefit of both.  \textbf{Now is a crucial moment to strengthen the link between AI and Science} by pursuing a strategy that proactively and thoughtfully leverages the potential of AI for scientific discovery and optimizes opportunities to impact the development of AI by applying concepts from fundamental science. 

\subsubsection*{Strategic Vision}

The opportunities for AI+MPS are vast, while of course resources are finite. It is not the intention of this white paper to prioritize specific activities or research areas, but we advocate that if incentives, funding, and resources are targeted toward \textbf{building a mutually beneficial bridge between MPS and AI research}, innovation and discovery will follow.  This means prioritizing activities that: 

\begin{enumerate}
\item \textbf{Enable AI+MPS research in both directions}: AI is a powerful tool for advancing scientific discovery \emph{and} MPS insights are essential for enhancing AI innovation and understanding; 
\item \textbf{Build up an interdisciplinary community of AI+MPS researchers}: Cross-disciplinary AI collaborations facilitate knowledge sharing and broad applications \emph{and} MPS domain-specific science necessitates novel AI development for various scales and rigor; and 
\item \textbf{Foster education and workforce development in AI for MPS researchers and students}: Broad AI education and training is needed for researchers at the AI+MPS intersection \emph{and} MPS has unique opportunities to integrate AI literacy into all levels of education. 
\end{enumerate}
Such a strategy that recognizes the ``two-way street'' between science and AI, as described in \Sec{sec:vision}, will serve to establish MPS as a leader in the next wave of AI adoption, development, and innovation.

\subsubsection*{Key Opportunities}
With the strategic vision in mind, this white paper presents opportunities for funding entities, university leadership, and researchers themselves to consider in building the future of AI+MPS:

\begin{itemize}
    \item \textbf{Develop a roadmap for MPS to lead the way in advancing AI+Science} by leveraging the substantial historical precedent for the virtuous cycle between MPS domain research and AI technological innovation (\Secs{sec:mutual-innovation}{sec:future}).
    \item \textbf{Establish diverse funding streams} to support interdisciplinary research at multiple scales, which would provide the flexibility and thoughtful investment needed to address the disruption caused by AI and remain at the cutting edge of research (\Sec{sec:funding}).
    \item \textbf{Pursue the ``Science of AI''} to apply scientific insights toward developing a deeper understanding of AI models and more robust, interpretable, and accurate AI tools for broad applications, which MPS is uniquely situated to do (\Sec{sec:science-of-ai}).
    \item \textbf{Develop scalable AI infrastructures}, including computing resources, the archiving of data, and the creation of software pipelines, which are required to accommodate the rapid growth of AI across the MPS domains (\Sec{sec:infrastructure}).
    \item \textbf{Facilitate cross-disciplinary collaborations} through activities such as workshops and exchanges that promote knowledge transfer within and beyond MPS (\Sec{sec:facilitating}). 
    \item \textbf{Advance key AI techniques for science} that have broad applicability and demonstrated impact across the MPS domains (\Sec{sec:techniques}). 
    \item \textbf{Leverage AI for conducting research}, including as a co-pilot and through self-driving labs  (\Sec{sec:conducting-research}).
    \item \textbf{Educate and train an AI+MPS workforce}, which is essential for continued progress in both scientific discovery and AI innovation; in particular, the MPS community would benefit from hiring and supporting early- and mid-career researchers at the AI+MPS intersection. On a longer time scale it will be important to revisit educational curricula and approaches, especially for undergraduate and graduate students (\Sec{sec:education}).
    \item \textbf{Empower AI innovation by reducing barriers} for people to access and use AI tools, such as by streamlining resources, enabling peer education, and strengthening communication, all of which is increasingly important as AI becomes pervasive across MPS (\Sec{sec:empower-ai}). 
    \item \textbf{Accelerate scientific discovery} in the MPS domains and innovation in AI by incorporating domain knowledge into AI approaches (\Sec{sec:domains}).
\end{itemize} 
This document concludes (\Sec{sec:summary}) with a summary of the key considerations for funding agencies, educational institutions, and individual researchers interested in advancing the future of AI+MPS.

\vfill

\section*{Note Added}
\label{sec:note-added}

As we were finalizing this document, the White House released ``Winning the Race: America's
AI Action Plan'' \cite{WhiteHouse2025}.
This action plan echoes many of the findings of this document, in particular its recommendations to:
\textit{Encourage Open-Source and Open-Weight AI} (\Sec{sec:infrastructure}), 
\textit{Empower American Workers in the Age of AI} (\Sec{sec:education}),
\textit{Invest in AI-Enabled Science} (\Secsss{sec:funding}{sec:techniques}{sec:conducting-research}{sec:domains}),
\textit{Build World-Class Scientific Datasets} (\Sec{sec:infrastructure}),
\textit{Advance the Science of AI} (\Sec{sec:science-of-ai}),
\textit{Invest in AI Interpretability, Control, and Robustness Breakthroughs} (\Sec{sec:science-of-ai}), and 
\textit{Build an AI Evaluations Ecosystem} (\Sec{sec:infrastructure}).

\newpage

\section{Introduction}
\label{sec:AI+MPS}

\glsresetall

The nature of intelligence, the dynamics of the physical world, and the structures of mathematics have long inspired human curiosity.
Through centuries of curiosity-driven research, we have gained insights into scientific domains, which have not only \textbf{enriched our understanding of the world around us} and our place in it, but also \textbf{sparked transformative technological advances}. 
A century ago, curiosity about atoms and their orbitals led to the development of quantum mechanics, which eventually led to the technology of the transistor, which is now the core technology behind modern computing.
In the reverse direction, the steam engine offered a practical technological advance, but it was only through fundamental research into the limits of thermodynamics that its power could be fully harnessed.
This demonstrates that \textbf{advances in science and technology require continued investments in fundamental research}, even when the exact outcome is not known or anticipated.

\textbf{Today’s \gls{AI} revolution is fueled by decades of basic research} across multiple disciplines in the \glslink{MPS}{Mathematical and Physical Sciences (MPS)}, which have also served as important catalysts for AI developments by providing challenging problems, datasets, and novel application contexts that stimulate innovation. The recent 2024 Nobel Prizes, for example, recognized pioneering work in the development of foundational methods underpinning modern AI technologies (Physics) \cite{NobelPrize2024Physics} and the applications of AI tools for computational protein design and structure prediction (Chemistry) \cite{NobelPrize2024Chemistry}. 
Indeed, the collection and careful curation of data (often taking decades) by MPS researchers have enabled breakthroughs with AI.

A few notable examples of recent AI-based advances from across the MPS domains are listed below, where the acronym indicates the corresponding division within the MPS Directorate of the U.S.\ \gls{NSF}: 
\begin{itemize}
    \item \textbf{Astronomy (\acrshort{AST})} has driven progress in real-time control systems, scalable image processing, anomaly detection, and \gls{UQ}, motivated by the analysis of massive sky surveys and rare astronomical events.
    \item \textbf{Chemistry (\acrshort{CHE})} has advanced AI through the development of graph neural networks and generative models for molecular design, significantly accelerating tasks such as materials and chemical matter discovery.  
    \item \textbf{Materials Science (\acrshort{DMR})} has spurred innovations in inverse design, autonomous experimentation frameworks, and multiscale surrogate modeling, enhancing the discovery and optimization of new materials. 
    \item \textbf{Mathematics and Statistics (\acrshort{DMS})} have laid foundational groundwork for AI and continue to provide the theoretical tools necessary to analyze and interpret modern AI systems at a deeper level, as well as serve as both a challenge and litmus test for pushing the boundaries of logic, abstraction, and reasoning in AI models. 
    \item \textbf{Physics (\acrshort{PHY})} has advanced real-time AI algorithms and incorporated symmetries into neural networks, to efficiently process the deluge of data from accelerator-based experiments. 
\end{itemize}

Of course, the nature of a revolution is that it is disruptive and not always cohesive.
The goal of this document is to show how coordinated efforts can help the MPS domains best capitalize on, and contribute to, the future of AI.
We outline a strategic vision for AI+MPS in \Sec{sec:vision}, which also provides a roadmap for the rest of this white paper.
(An additional roadmap appears in \Sec{sec:summary}, which highlights key considerations for funding agencies,
educational institutions, and individual AI+MPS researchers.)
Then, in \Sec{sec:mutual-innovation}, we establish a common understanding/language across MPS for what we mean when we talk about AI, and summarize the virtuous cycle of AI+MPS so far, setting a baseline for how AI is advancing MPS and how MPS is advancing AI.
Finally, in \Sec{sec:future}, we consider the potential for transformational outcomes that could result from dedicated AI+Science research.

\subsection{Strategic Vision for AI+MPS}
\label{sec:vision}

By addressing their unique domain challenges, scientists have continuously expanded the practical capabilities of AI, both for scientific applications and more broadly.
In surveying the MPS community, we found many commonalities across disciplines in their approaches to AI, which point to opportunities for collaboration. By working together to develop techniques and explore research opportunities, we can make substantial progress toward advancing AI+MPS as a research domain in its own right. To achieve this, it will be important to \textbf{bolster the AI+MPS field through funding, infrastructure, collaboration, and education}, while also identifying \textbf{focused efforts within domains where AI will have the biggest impact.}

The symbiosis between AI development and scientific discovery is the foundation upon which to build future strategies for AI+MPS. By targeting incentives, funding, and resources toward \textbf{building a mutually beneficial bridge between MPS and AI research}, innovation and discovery will follow. A strategy that recognizes the ``two-way street'' between science and AI will help establish MPS as a leader in the next wave of AI adoption, development, and innovation. 

\subsubsection{Enable AI+MPS Research in Both Directions}

MPS researchers are already leveraging AI to accelerate scientific discovery and using scientific principles to drive AI understanding (\Sec{sec:mutual-innovation}), and there is significant potential for further innovation in both directions (\Sec{sec:future}). It is therefore important to recognize, when considering strategies for enabling AI+MPS research, that the benefits extend to both MPS and AI, such that: 

\begin{itemize}
    \item \textbf{AI is a powerful tool for advancing scientific discovery}, as demonstrated by historical successes. Looking forward, there are opportunities to leverage key AI techniques that already show promise for scientific applications (\Sec{sec:techniques}) and to use AI itself as a partner for conducting research (\Sec{sec:conducting-research}). Within and across MPS domains, researchers can take advantage of AI to accelerate their research and provide new insights (\Sec{sec:domains}).
    
    \item \textbf{MPS insights are essential for enhancing AI innovation and understanding}, which indicates a need for further research in the ``Science of AI'' (\Sec{sec:science-of-ai}) that will generate AI innovations, enhance our understanding of AI, and support the development of robust AI techniques. Such research will also empower AI innovation (\Sec{sec:empower-ai}) by helping make AI research more cost-efficient, improving the scientific integrity of AI, and communicating the impact of AI broadly. 
\end{itemize}

\noindent With this mutual innovation in mind, \textbf{resources should be allocated to efforts with impact in both MPS domains and AI development}, supporting AI for scientific discovery \emph{and} scientific insights for AI innovation.

\subsubsection{Build an Interdisciplinary AI+MPS Community} 

The development of AI techniques and research into AI is pervasive across MPS disciplines. Encouraging collaboration and creating opportunities for knowledge sharing are essential to build on existing successes and achieve AI innovations with impactful applications. Building an interdisciplinary community of researchers who are interested in AI should put equal emphasis on collaboration across disciplines and on domain-specific research, because:  

\begin{itemize}
    \item \textbf{Cross-disciplinary AI collaborations facilitate knowledge sharing and broad applications}, reducing duplicate effort and amplifying the impact of AI+MPS research. Such collaborations can be encouraged by funding various types of interdisciplinary efforts (\Sec{sec:funding}) and providing scalable resources (\Sec{sec:infrastructure}). The community itself can also facilitate collaborations through workshops, conferences, and other events (\Sec{sec:facilitating}).
    \item \textbf{MPS domain-specific science leads to novel AI development for various scales and rigor}, which can inspire developments in other domains even beyond MPS. MPS researchers have a key role to play in AI innovation by leveraging the solutions they are developing to address domain-specific problems (\Sec{sec:domains}). It is therefore equally important to build AI-enabled communities within domains, enable knowledge sharing across MPS domains, and bring AI researchers interested in scientific applications into MPS communities to collaborate.
\end{itemize}
To support interdisciplinary work in AI, it is essential to \textbf{identify ways to facilitate interdisciplinary research collaboration and establish community hubs} for knowledge exchange, both within and across MPS domains.

\subsubsection{Foster AI+MPS Education and Workforce Development}

Expertise and discoveries developed from basic research, particularly in academic, national lab, and industry partnerships, as well as the rapid scaling and commercialization of AI by for-profit companies, are driving U.S.\ leadership in AI. Encouraging the pursuit of cutting-edge AI research by early-career scientists can help cultivate an AI-enabled \acrshort{STEM} workforce and help sustain U.S.\ leadership in AI. It is therefore important to consider strategies for education and workforce development that equip MPS researchers with the knowledge and tools necessary to navigate the AI revolution. By establishing a pipeline of students who are intellectually nimble critical thinkers, capable of recognizing AI’s potential and deploying it to solve complex problems, we can better adapt and evolve alongside rapidly evolving AI technologies. This means:

\begin{itemize}
    \item \textbf{Broad AI education and training is needed for researchers at the AI+MPS intersection }in order to build an AI-enabled STEM workforce in both academia and industry. This includes training for senior researchers who may be using AI in their research for the first time (\Sec{sec:faculty}), as well as for early-career researchers with varying levels of experience using AI (\Secs{sec:postdocs}{sec:graduate}).
    \item \textbf{MPS has unique opportunities to integrate AI literacy into all levels of education}, which will not only strengthen the future AI+MPS talent pool (\Sec{sec:undergrad}), but will also help shape public perception of AI (\Sec{sec:public}). Establishing credentials in AI+MPS could attract more students and researchers to the discipline (\Sec{sec:graduate}), given the increasing focus on related skills for jobs in both academia and industry. At the same time, MPS domain science education remains essential for developing the experience and intuition necessary to use AI productively.
\end{itemize}

The unprecedented scope of upskilling required affects virtually everyone in MPS fields. By proactively integrating AI into education and training---\textbf{developing both AI-related training for MPS researchers and leveraging MPS education to strengthen AI literacy}---the MPS community can more effectively demonstrate and maintain its leadership in AI.

\subsection{Mutual Innovation in AI+MPS}
\label{sec:mutual-innovation}

Strengthening the link between AI and MPS has the potential to significantly impact both scientific discovery and AI development. By establishing a common language and understanding the historical precedent for the virtuous cycle between MPS and technological advances, we can take a proactive and thoughtful approach to AI+MPS innovation. 

\subsubsection{AI in the Context of MPS}
\label{sec:ai-context}

The term ``artificial intelligence'' is understood in diverse ways across different communities, reflecting its rich historical evolution. Similarly, \glslink{ML}{``machine learning'' (ML)}\glsunset{ML} is interpreted variously as a subset, a synonym, or a complementary approach to AI---hence the common use of ``AI/ML.'' While AI is often associated with mimicking human intelligence, classical \gls{ML} approaches like linear regression learn in ways that are quite distinct from human cognition. 

\textbf{We advocate treating AI as a broad umbrella} that encompasses a wide range of computational and statistical innovations (including symbolic/rule-based systems and \gls{ML}) that are relevant for advancing the MPS domains.
By encouraging innovation under this broad umbrella, we can incentivize AI+MPS researchers to apply the most effective strategies to their problems, rather than relying on trends.
While we will not attempt to provide a complete taxonomy of AI, here we explain some terminology that captures the range of AI algorithms and applications that have had an impact on MPS research.

\begin{itemize}

    \item \textbf{Shallow versus deep \gls{ML}}: Shallow ML encompasses classical statistical methods such as linear/logistic regression and straightforward generalizations like \glspl{BDT}. These methods can be particularly useful for processing sparse data sets where stronger assumptions are needed to achieve robust results.

    In contrast, deep ML refers to \glspl{DNN}---including multi-layer feedforward networks, \glspl{CNN}, and transformers---that are highly adaptable and capable of identifying complex features in data sets. The rise of deep ML sparked the current wave of AI innovations, and understanding how these algorithms achieve such effective performance remains an open research question within the ``Science of AI'' (\Sec{sec:understand-ai-behavior}). 

    \item \textbf{Supervised versus unsupervised learning}: Supervised learning seeks to train regression or classification models to predict data labels. Unsupervised learning operates on unlabeled data to extract patterns, clusters, or low-dimensional structure, and is helpful for data discovery and outlier detection, which can be critical to account for ``unknown-unknowns'' and provide a (relatively) unbiased approach to data exploration. Semi-supervised learning blends these concepts to operate on partially labeled data sets. Self-supervised learning operates on unlabeled data, but where the training paradigm creates its own ``supervisory signal,'' often by predicting one part of the data from another (e.g., masked language modeling). It has been an important paradigm in training \glspl{LLM} where the data corpora are too large to admit human labeling of all training instances. In weakly-supervised learning, coarse-grained or inaccurate labels are available, while in self-supervised learning, labels are automatically generated during pretest task training and transfer learning is applied to the downstream training task of interest.
    
    \item \textbf{Discriminative versus generative models}: \Gls{ML} models generally can be broken into two categories: discriminative models map observations to labels and make predictions, while generative models learn the underlying distribution and manifold of the data and may or may not be conditioned on labels. Discriminative models have made significant inroads in MPS, from classifying galaxy morphologies and astronomical transients to regression tasks for physical parameter estimation. These models excel at tasks like reconstructing particle collision events or predicting material properties. 

Generative models aim to learn a process that can generate new data points that resemble a given set of samples, even though the true underlying distribution that produced the original data is unknown. Examples of generative models are \glspl{GAN}, \glspl{VAE}, autoregressive models, \glspl{NF}, and diffusion models \cite{Yang2023}. Recently, generative models have driven much of the progress across multiple domains and in multiple applications, including anomaly detection, surrogate modeling, and \gls{SBI} (\Sec{sec:sbi}). 

\item \textbf{Deterministic versus stochastic models}:  Deterministic and stochastic models differ in how they handle uncertainty and randomness.
Deterministic models produce the same output every time given the same input and parameters; there is no inherent randomness in their predictions.
By contrast, stochastic models incorporate probabilistic elements like noise to yield random outputs for the same inputs; such models are often better suited for capturing uncertainty and variability in data.
Even when the goal of machine learning is to describe a probabilistic process, one can work with either deterministic or stochastic models: a deterministic model would simply output the probability $p(x)$ for a given outcome $x$, while a stochastic model would map input noise to different $x$ values according to that probability distribution.

\item \textbf{Reasoning versus agentic models:} Modern AI systems are increasingly demonstrating sophisticated reasoning and agentic capabilities. Reasoning models are designed to tackle complex intellectual tasks by breaking them down into intermediate steps, evaluating multiple perspectives, and forming coherent chains of thought---abilities particularly valuable for mathematical and scientific problem-solving. While reasoning focuses on the internal cognitive process, agentic models are designed to take autonomous actions to accomplish goals, often involving multiple steps of planning, tool use, and interaction with their environment (including computational resources, databases, or experimental equipment). These models can formulate hypotheses, design experiments, analyze results, and iterate on scientific workflows with varying degrees of autonomy. The distinction between reasoning and agency represents a spectrum rather than a binary classification, with state-of-the-art systems increasingly combining both capabilities to serve as effective research assistants or even semi-autonomous scientific collaborators (\Sec{sec:ai-co-pilot}).

\item \textbf{Foundation models:} Foundation models are large-scale AI systems trained on vast, diverse datasets. They serve as general-purpose bases for adaptation to a wide range of downstream tasks through fine-tuning or prompting. These models, exemplified by \glspl{LLM}, vision transformers, and multimodal systems, are characterized by their scale, generality, and emergent capabilities that weren't explicitly programmed. The term ``foundation'' reflects their role as versatile platforms upon which domain-specific applications can be built. In the MPS context, domain-specific foundation models trained on scientific literature, experimental data, and simulations can provide powerful starting points for specialized scientific applications, enabling knowledge transfer across related problems and reducing the data requirements for new tasks compared to training from scratch (\Sec{sec:foundation_models}).

    \item \textbf{\glsreset{LLM}\Glspl{LLM}:} The term ``generative models'' has taken on prominence in the context of \glspl{LLM}. Starting from a ``prompt'' as input, these algorithms output rich text responses that can describe scientific concepts, generate hypotheses, or even suggest experimental designs.
    While \glspl{LLM} have demonstrated impressive capabilities, the debate continues about their fundamental nature. Some researchers still maintain they can be viewed as ``stochastic parrots'' that primarily recombine training data in sophisticated ways. Others see them as a potential pathway toward \glslink{AGI}{``artificial general intelligence'' (AGI)}\glsunset{AGI}, pointing to their ability to reason about novel scientific questions and integrate knowledge across domains. Genuine successes in this area have been reliant upon logic-based reasoning engines (e.g., Lean for formal math proofs \cite{Lean}) to perform reasoning with the rigor typically required in MPS, while \glspl{LLM} have been limited to translating into appropriate syntax and/or invoking such logic engines as agents. 

    \item \textbf{\glsreset{RL}\Gls{RL}:} \Gls{RL} is a type of machine learning in which an agent learns to make decisions by interacting with an environment in order to maximize the expected cumulative reward over time, typically called the return. In this framework for sequential decision-making, an agent: (i) observes the current state of the environment; (ii) chooses an action based on a policy (or strategy); (iii) receives a reward and possibly transitions to a new state; and (iv) learns from this feedback to improve future decision-making. \Gls{RL} provides a mathematical and algorithmic framework for creating agentic behavior. Not all agentic models use \gls{RL} (e.g., agentic behavior can be hand-coded), although \gls{RL} can still be used to enhance agentic qualities.

    \item \textbf{Big data versus small data}: The ``bigness'' of data is highly subjective and dependent on the field, application, and eye of the beholder. A reasonable working definition of big data is data that is big enough that you need to think about managing your data access, mandating thoughtful approaches to storage, analysis, visualization, and sharing. This can be any data set where the volume and complexity is an order of magnitude more than has been commonly used in existing applications. It is the rapid increase in the relative amount of data that drives the challenge in individual fields (even if the absolute volumes might be considered small by other fields).
    By contrast, some MPS applications involve ``small data,'' where there is significant cost to acquiring reliable experimental results or performing computational simulations.
    The small-data regime requires complementary strategies to ensure that data is used efficiently (\Sec{sec:data-efficient}). 
    
    \item \textbf{Symbolic AI and formal methods:} Logic-based symbolic methods have traditionally played a crucial role in computer science, particularly in specifying and verifying hardware, software, and communication protocols. These formal methods primarily utilize \glspl{SAT} and more generalized frameworks such as \gls{SMT}. Fundamental computational breakthroughs in these areas during the 1990s significantly improved the scalability and practicality of symbolic reasoning tools, according to a report by the National Academies \cite{NationalAcademies2023}. In contemporary mathematics, automated reasoning has become a central focus, leading to the development of first-order theorem provers and interactive proof assistants, enabling machine-assisted proofs for problems such as the Pythagorean triplet problem. Recently, integrating symbolic methods with AI has spurred innovations in mathematical reasoning systems, such as AlphaProof and AlphaGeometry, which demonstrate unprecedented capabilities in automated theorem proving and geometric reasoning.

    \item \textbf{Data science}: Data science is the discipline of extracting knowledge from data, typically using tools from computer science, applied mathematics, and statistics. Frequently, the problems involved are knowledge discovery from data at large scales (i.e., using big data). In this context, data science requires combining the expertise to synthesize domain knowledge with the skillful application of computer science and statistics. 
\end{itemize}

\subsubsection{How AI is Accelerating Scientific Discovery in MPS}
\label{sec:AIadvanceMPS}

There are many examples highlighting the impact that the rapid adoption of AI has had already in the MPS domains. A few examples that span across disciplines include:

\begin{itemize}
\item \textbf{Enabling \acrfull{SBI}:}
AI has opened new avenues for drawing scientific inferences from models so complex that their statistical likelihoods are intractable or unknown, though reliable simulations do exist.  This capability of \gls{SBI} can bring fresh insights across disciplines---such as identifying the parameters that characterize the cosmological model that best describes the universe or refining collider event modeling in particle physics. AI’s potential to model high-dimensional probability distributions also helps scientists quantify and understand the uncertainties inherent in their predictions. Opportunities for cultivating further advances in \gls{SBI} are described in \Sec{sec:sbi}.

\item \textbf{Accelerating simulations:}
\Gls{SciML} techniques \cite{Keith2025} accelerate simulations in various scientific and engineering domains by addressing the limitations of traditional computational methods, as discussed further in \Sec{sec:simulation}.
AI can be used to predict the time evolution of complex dynamical systems and construct efficient surrogate models that can make predictions at a fraction of the computational cost, which is especially critical in outer-loop and many-query problems in inference and design. The learned collective variables can also expose new fundamental understandings of the dynamical systems and the models can be profitably constructed within virtuous iterative loops of simulation and model training. Algorithms leveraging the function approximation properties of neural networks have also enabled simulation of high-dimensional dynamical systems beyond reach of traditional techniques.  

\item \textbf{Facilitating pattern recognition and anomaly detection:} AI techniques, such as \gls{RL} and generative models, offer complementary frameworks for exploring vast search spaces. \Gls{RL} can help identify ``needles in the haystack'' in applications ranging from drug discovery and materials design to the construction of counterexamples for mathematical conjectures. Generative models are effective for anomaly detection through statistical inference and likelihood evaluation. These approaches enable AI to uncover patterns in massive scientific datasets from space- and ground-based telescopes, particle colliders, and other research facilities, paving the way for discoveries. More detail on these techniques is provided in \Sec{sec:research-opportunities}.

\item \textbf{Leveraging representation learning:} AI enables scientists to augment incomplete, simplified first-principle models with data to simulate, understand, and control complex processes. Representation learning can be a tool for extracting key scientific features from data by, for example, identifying low-dimensional structure and relevant variables. These features then enable better data-efficient predictions, simulations/surrogate models of complex systems, digital twins, etc. Better representation learning can also facilitate  AI robustness (see \Sec{sec:robust-and-reproducible-AI}) if one can validate that the learned representations are aligned with domain knowledge.
Through representation learning, AI can help build bridges and be incorporated in solutions such as digital twins, self-driving labs, and human--machine interfaces (\Sec{sec:conducting-research}).

\item \textbf{Enhancing predictions:} Large AI systems, such as foundation models, that are pre-trained on extensive datasets from observations and simulations can drive breakthroughs in domains with limited physical models and small, noisy datasets. For instance, domain transfer techniques enable the use of abundant, simpler simulations to enhance predictions in more \textit{ab initio}, accurate simulations that are fewer in number 
(see also domain adaptation in \Sec{sec:research-opportunities}). Similar approaches could also be applied to areas such as studying chemical reactions at equilibrium, identifying construction defects in additive manufacturing, or analyzing climate and weather patterns.

\item \textbf{Streamlining the scientific workflow:} AI has begun to fundamentally impact the scientific workflow by streamlining background research, effectively collecting useful data, automating the design of experiments, and rapidly testing and validating hypotheses. For example, AI has the potential to change the way that complex experiments operate in the future, with an eye on developments beyond data analysis. 
Better data quality monitoring algorithms, for example, will result in larger datasets with fewer pathologies that need to be understood later, potentially making the entire scientific enterprise in such collaborations more efficient. Further suggestions for leveraging AI for conducting research are presented in \Sec{sec:conducting-research}.

\item \textbf{Enhancing communication:} AI has significantly influenced the ways in which MPS researchers communicate, collaborate, mentor researchers, and educate students. It holds immense potential to bridge communication gaps across diverse scientific disciplines by translating complex concepts and findings, offering advanced coding support and assisting researchers in making their writing more clear and impactful (see \Sec{sec:conducting-research}). 

\end{itemize}
Additional details on these AI examples and more are discussed in the context of scientific research  generally in \Secs{sec:techniques}{sec:conducting-research} and within specific MPS domains in \Sec{sec:domains}.

\subsubsection{How MPS Research is Driving AI Understanding}
\label{sec:MPSadvanceAI}

AI innovation, especially to meet the growing demands of scientific discovery, requires advances that go beyond traditional computer science, making MPS essential to this effort. MPS offers a unique advantage as a relatively low-risk, high-reward ``playground'' for testing and refining AI tools, given its history of sharing and using data in a way that is impractical for many other fields (due to, e.g., privacy concerns for health data, financial implications for proprietary corporate data, etc.). Pursuing research in key MPS areas not only drives breakthroughs in AI itself but also enhances AI tools in ways that are uniquely suited to solving complex scientific problems. This creates a virtuous cycle where \textbf{AI innovation and scientific progress reinforce one another, with MPS at the core of both.}

Some of the most pressing AI research priorities that MPS disciplines are well-poised to advance include:
\begin{itemize}

    \item \textbf{Incorporating fundamental physical laws into AI}: In using AI for scientific discovery, researchers are creating methods that systematically embed physical laws---such as ordinary and partial differential equations, symmetries, invariances, and equivariances---directly into AI models, as described in \Sec{sec:ai-innovations}. This approach can enhance the models' predictive accuracy and scientific validity and inform the construction of digital twins. In many cases, the AI models can also learn physical laws, such as \glspl{HNN} and \glspl{LNN} learning the Hamiltonian and Lagrangian of a system. 
    On the other hand, there are applications where purposefully modifying or augmenting the physical laws may be desirable in the service of, for example, faster searches, more efficient models, or generation of more creative hypotheses or solutions. This can also be a good test of model robustness. 
    
    \item \textbf{Improving interpretability and transparency}: MPS researchers have the expertise to develop new mathematical, statistical, and physical frameworks to understand AI models, transforming them from ``black-box'' systems into scientifically transparent tools that lead to reliable results. In some cases, the expertise may expose bias or unreliable results, motivating further research and testing toward achieving reliability and transparency and avoiding societal harm. Opportunities in this direction are discussed further in \Secs{sec:understand-ai-behavior}{sec:robust-and-reproducible-AI}.

    \item \textbf{Improving efficiency and reducing data dependency}: Leveraging expertise from across MPS disciplines using scarce or noisy scientific data has led to improved computational efficiency of AI techniques, as discussed in \Secs{sec:ai-innovations}{sec:cost-efficient}. 
    Unlike in many industry settings, where access to large datasets and compute resources are the norm, MPS research seeks to gain insight from expensive, complex, scarce data with modest computational means. This motivates basic research into data-efficient and physics-informed AI schemes that generalize well without excessive training and hyperparameter tuning.

    \item \textbf{Creating benchmark problems and datasets}: The immense progress in AI can often be linked to benchmark problems and datasets that have been used to track and motivate advances over long periods of time. By providing well-curated, thoughtfully designed benchmarks, MPS disciplines can contribute to progress within their fields, simplify interdisciplinary collaborations, and even enable advances in AI itself. However, one must be cautious of the tendency of benchmarks to focus effort on one problem at the expense of others, incentivize ``gamification,'' and generate overly narrow solutions that do not generalize well beyond the benchmark. Ideas for balancing these concerns are discussed in \Sec{sec:benchmark}. 
    
    \item \textbf{Quantifying uncertainty and ensuring reliability}: In an effort to ensure AI-generated scientific predictions are robust, reliable, and suitable for high-stakes scientific applications, MPS researchers are developing rigorous techniques for \gls{UQ}, error estimation, and out-of-distribution detection that have the potential to be extremely important for many other application domains, particularly those where it is much more difficult to recognize unreliable predictions \cite{Abdar2021}. \Sec{sec:uncertainty_quantification} discusses \gls{UQ} opportunites in more detail.

    \item \textbf{Enhancing \gls{RL} for sparse rewards}: By inventing novel \gls{RL} algorithms tailored to scientific scenarios where rewards (e.g., corresponding to new discoveries or detected anomalies) are extremely sparse, MPS researchers are making advances in \gls{RL} that are critical for breakthroughs in increasingly complex tasks, such as those used for experimental control as described in \Sec{sec:rl_for_experiment}.

    \item \textbf{Providing a safe ecosystem of development:} Many MPS fields offer an ideal sandbox for developing AI tools in an environment where the data and problems are less likely to lead directly to negative human impacts. Such a safe environment could be the key to developing \gls{AGI}, without being severely red-taped by policy. \Sec{sec:domains} provides insight into the domain-specific problems that can be used for developing AI and their impact on AI innovation. While safe, most of the MPS discoveries are high stakes and important for national prestige and competitiveness on the world stage.
\end{itemize}

\subsection{The Future of AI+MPS}
\label{sec:future}

There are many opportunities across domains to advance AI+MPS and in doing so, have a substantial impact on research outcomes in both AI and MPS. Though it is challenging to predict what would be the most impactful, we can extrapolate some possibilities based on existing efforts. Current AI advances are trending toward the development of more capable AI co-scientists and could even lead to fully autonomous AI scientists that can generate complex, scientifically meaningful hypotheses, design experiments, and interpret results.
Ideally, such AI co-scientists would operate in a way that preserves (and even enhances) existing standards for scientific rigor and integrity.
Perhaps the most ambitious goal is the Nobel Turing challenge \cite{Kitano2021}, wherein AI makes a discovery at the Nobel level. These are certainly interesting and potentially worthwhile pursuits, but they would require substantial domain-specific meta-analysis on the use of such tools. 

More specific to MPS, we anticipate that \textbf{combining AI with domain sciences will lead to significant, transformational outcomes}, including: 

\begin{itemize}
    \item \textbf{Establishing a new scientific domain:} By cultivating an AI+MPS community, training an AI+MPS workforce, and funding collaborative AI+MPS research, we can establish a new domain at the intersection of AI and Science, similar to interdisciplinary fields like biophysics  and quantum computing, and a coordinated effort would establish the U.S.\ as a leader in this new domain.
    \item \textbf{Breaking open the AI ``black box'':} Much of AI research, even in other disciplines such as neuroscience, tends toward identifying \textit{how} to build AI that will work as expected, which has led to the perception of AI as a ``black box.'' The inherent nature of the MPS domains demands to understand \textit{why} it works, and we have the tools to in these domains to make substantial progress. This ``Science of AI,'' if duly recognized and funded, has the potential to transform our understanding of AI and subsequently enhance our ability to interpret, develop, and control it.
    \item \textbf{Accelerating the pace of scientific discovery:} Innovative technology has exponentially impacted the amount and precision of data that MPS domains have to work with, but our ability to analyze, interpret, and make discoveries from that data has not kept up. By building rigorous, domain-specific AI tools, MPS researchers have the potential to accelerate the pace of discovery, potentially leading to major breakthroughs, such as understanding the nature of dark energy and dark matter, developing new drugs, designing new materials for extreme environments and/or sustainability, verifying complex mathematical results, or identifying new particles. Such discoveries become the foundations for far-reaching applications in fields such as medicine, environmental remediation, social science, and national defense. 
\end{itemize}

There is ample opportunity to take a strategic and intentional approach to building a dynamic and innovative AI+MPS community that harnesses the AI revolution to advance cross-disciplinary research and education, as well as domain-specific advances. This white paper is an effort to present a broad sampling of these opportunities and communicate the impact they could have on both AI and scientific research and innovation.

\section{Cross-Disciplinary AI+MPS Opportunities}
\label{sec:cross-disciplinary}

\glsresetall

Opportunities for advancing AI+MPS across the MPS domains range from logistical (e.g., funding, infrastructure) to scientific (e.g., developing techniques, conducting research) to educational (e.g., workforce training, curriculum). MPS will have the biggest impact in these areas if the \textbf{domains collaborate to develop strategies, transfer knowledge, and innovate.} The following sections lay out specific approaches and areas of impact that are applicable to all the MPS domains and, in many cases, require bringing the domains together. 

\subsection{Advocate for Diverse Funding Streams}
\label{sec:funding}

\glsresetall

With the growing use of AI tools has come a paradigm shift in how research is conducted. 
As plans for the future of AI funding in MPS move forward, nimbleness in the opportunities and structures made available will be increasingly important. 
\textbf{Diverse funding streams on multiple scales will help to harness a range of talent and expertise across the U.S.}, all to the benefit of AI+MPS. Funding needs include a range of approaches: 
\begin{itemize}
    \item \textbf{Exploratory work to application-based work}: Flexible funding for research that explores development of new AI, improvements to AI techniques, and applications of AI is as important as funding for research with a clearly established application. Ideally, this would also allow for successful exploratory projects to increase their funding with an equally quick turnaround to ensure the work does not go out of date.
    \item \textbf{Cross-domain collaborations to interdisciplinary experts:} In some cases, the most impactful approach to make advances will be to bring together cross-disciplinary teams of domain experts and AI experts; in other cases, it will be more effective to fund individuals or groups that have interdisciplinary expertise and can advance the field independently and then share with the broader community. Such polymaths, or ``centaur scientists,'' are crucial to advancing AI+Science and it is important to find ways to support their work.
    \item \textbf{Short-term to long-term projects:}
    Short-term projects may be sufficient for some advances, especially if they have clearly established goals or are led by individuals with interdisciplinary expertise, but long-term projects may be more effective in cases where researchers need to build collaboration and/or are exploring opportunities to advance the field.
    Small awards with rapid review times enable nimble research, especially due to the exceedingly fast pace of development in AI.
    For example, programs like NSF ACCESS \cite{ACCESS} have enabled quick turnaround times for small requests for computing resources (though funding for computing is typically separate from funding for people).
    At the other end of the spectrum, longer-term funding on the 5–10 year scale can better enable the development of large-scale AI projects and ensure their sustainability for the community.
\end{itemize}

For any of these AI+MPS funding streams, \textbf{proposal evaluation will need to take into account both cross-cutting and domain-specific issues}. 
For example, it would help support innovative research to ensure that the evaluation criteria for reviewers not only consider the expertise of the \glspl{PI} in each of the involved disciplines, but also their ability to assemble and lead an interdisciplinary team, which involves a complementary set of skills that are not always considered in evaluation metrics. For interdisciplinary proposals, it may be (and is often) the case that the variety of expertise required to evaluate the full scope of the proposed research cannot be covered by individual reviewers, and thus multiple individuals may be required to evaluate the work in full. This creates additional overhead to consolidate input, reach a consensus, and decide how to weigh strengths/weaknesses of AI aspects against MPS aspects. Assessment of individual interdisciplinary grants may benefit from having ad hoc review panels to supplement traditional panels that collectively review a full set of submissions.

The subsections below provide examples of funding streams that would help address the various scales and needs for conducting innovative, interdisciplinary research.

\subsubsection{Institute-Scale Activities}

Institute-scale activities involving dozens (or even hundreds) of investigators provide essential infrastructure where researchers can meet across disciplines and forge collaborative ties across institutions. These initiatives can also facilitate networks of institutes, uniting \glspl{PI} from interdisciplinary programs to share insights and best practices. Expertise is often dispersed across different universities and research organizations, making such networks particularly valuable for \textbf{driving cohesive, cross-disciplinary innovation.}

To support interdisciplinary work in AI, researchers will need funding to \textbf{support community hubs and research collaboration}. Currently, the \acrshort{NSF} and \acrshort{DOE} fund high-profile centers, such as the NSF National AI Research Institutes \cite{NSFAIInstitutes}, Centers for Chemical Innovation \cite{NSFChemicalInnovation} like the Center for Computer Assisted Synthesis that attracted substantial industrial funding, and big collaborations led by the national labs, which have shown to be successful in building cross-disciplinary teams, leading to impactful results. There are also examples of collaborations with funding partners; for example, the Simons Foundation has been equal partners with the NSF on both Center- and Institute-level AI-focused/related projects in MPS, including the \gls{MoDL} collaboratives (2020), the \gls{NITMB} (2023), and most recently the two new AST-themed National AI Research Institutes (2024). Broadening beyond this base is essential to grow a larger and more vibrant AI+Science community. One way to approach this is to leverage the existing institutes and build on the work they have already done, such as sharing educational and community resources, developing best practices for AI+Science, and supporting infrastructure. Combining efforts rather than launching new efforts every few years would allow the community to build on the momentum of existing institutes.

Center-type structures help support the formation of larger teams with \textbf{infrastructure to conduct large-scale research and build community}. In many cases, institute-scale activities are the primary venue for experts from different domains to meet and learn about each other's research, which may be complementary and therefore lead to further innovation through collaboration (see \Sec{sec:facilitating}). 

\subsubsection{Project-Scale Activities}

More traditional, project-based funding (led by 2–5 PIs) also has an important place in advancing interdisciplinary research, especially when there is an expected outcome to pursue. Importantly, the novelty requirements of proposals for AI methods applied to domain applications cannot be the same as for AI proposals generally, because many important breakthroughs come from applying existing AI methods in new ways. \textbf{Explicit inclusion of AI technology in project solicitations and proposals} can help ensure that criteria is clear for reviewers who may be less familiar with the potential of AI to push forward traditional research fields. 

\textbf{Incentivizing AI researchers to partner with domain scientists} to develop tools that are useful for MPS would help to advance AI+Science, in terms of both research and community. One way to do this is to provide funding for projects that encourage collaboration between AI developers, who have the general AI expertise, and domain experts, who understand their domain and what the tools need to do. Development of useful tools requires input from both sides, as well as sustained interactions. This may require that funding for such projects be contributed from multiple divisions within funding agencies. In many cases, it may also be helpful for these projects to include an interdisciplinary researcher who can help to build the bridge and open lines of communication between the domains.

An example of a program at the NSF that supports project-based funding for software development and scientific domain research is the \gls{CSSI} program \cite{NSFCSSI}.This could serve as a model for AI+MPS funding, with the software aspect of \gls{CSSI} replaced by AI. The model co-funds grants between the Directorate for \glslink{CISE}{Computer and Information Science and Engineering (CISE)}\glsunset{CISE} and a science domain, e.g.\ MPS/PHY. The solicitation includes software-specific questions and information, e.g.\ deployment plans, long-term maintenance, etc., but is open to all participating domains. The reviewers in this program are specifically chosen to be experts at software development, and come from various domains. Projects are only funded if the science domain is willing to co-fund, which ensures that the domain application is exciting, and the reviewers being experts on the technical side ensures that the right people are chosen to be funded to perform the project. Of course, in such a case, it would be important for the expectations that will be reviewed by both domain researchers and AI experts to be clearly stated in the initial solicitations.

\subsubsection{Individual Investigators}
\label{sec:individual}

A promising approach to advancing AI+MPS research is to \textbf{nurture interdisciplinary experts who integrate deep domain knowledge with advanced AI skills}. Fostering these professionals would create a positive pathway to rigorously benchmarking and refining AI tools for scientific applications (discussed further in \Secs{sec:science-of-ai}{sec:techniques}), and ensuring that computational methods are directly aligned with domain-specific challenges and opportunities.

Currently, researchers have few avenues to obtain funding that allows them to focus their time on bridging gaps between MPS domain science and AI. One promising approach is to \textbf{support a dedicated connector role---a ``centaur'' or polymath}---by offering incentives for universities to train or hire individuals in this capacity. For example, establishing a three-year Investigator program for PIs who have a proven track record in interdisciplinary initiatives could empower them to effectively link diverse teams and mentor emerging researchers. An existing example of this is the Schmidt Science Polymaths Program \cite{SchmidtPolymaths}, which funds recently-tenured professors working on interdisciplinary research broadly, including but not restricted to AI+Science. This could also lead to opportunities for expanding retraining and up-skilling programs (see more in \Sec{sec:faculty}), which would enable established scientists from both the domain and AI sides to immerse themselves in the complementary discipline, ensuring that more senior researchers remain agile and well-equipped to drive innovative, cross-disciplinary collaborations. Ideally, this type of funding would include a foreseeable pathway for individual researchers from smaller institutions (like R2 and smaller) to develop more comprehensive and theoretical AI research. For example, funding faculty to buy them out of their teaching for 6–12 months so that they can pursue opportunities for upskilling would complement the current 3-year Investigator program, helping to scale to a broader community of researchers.

A similar approach could be used for early-career researchers. \textbf{Funding postdoctoral and/or graduate student fellowships at the intersection of AI and domain science} would attract interdisciplinary experts from the next generation of talent who have already begun to gain experience connecting communities (see \Secs{sec:postdocs}{sec:graduate}). There have been examples of this as a successful model in programs like the NSF's MPS-ASCEND program \cite{NSFMPSAscend}, the NSF \acrshort{IAIFI} Fellowship \cite{IAIFIFellowship}, and the Schmidt AI in Science Fellowship \cite{SchmidtAIScience}. Importantly, researchers hired into these roles should have substantial independence in their research direction, allowing them to pursue innovative ideas and collaborate effectively across domains.

One substantial challenge to this approach is that it requires identifying researchers with the right combination of expertise or finding and funding domain scientists willing to invest the time to do a deep dive into specific AI methods that could solve their problem. Funding people (whether junior researchers directly or research groups led by an experienced PI) for multiple years at a time might be one way to \textbf{encourage people to invest the time necessary to integrate AI} in thoughtful and deliberate ways.

\subsubsection{Industry Collaborations}

Engaging industry in the application of basic research, data generation, and workforce development through funding, incentives, and recognition could substantially increase real-world impact of this work. There is significant expertise in industry that is potentially relevant. Interactions/networks across industry and academia also provide a natural career pipeline for young researchers. Therefore, it is increasingly important to \textbf{identify how corporate partners can be better engaged in collaborating on MPS problems.} A useful strategy is to find ways to leverage the tremendous investments being made in commercial AI (e.g., \glspl{LLM} and their applications) and to interest the for-profit community in MPS science problems and data.
This has already happened to some extent with AI benchmarks based on mathematical problem solving, which can test AI reasoning abilities in a provably verifiable domain, or the development of AI applications in academic/industrial collaborations.

Close ties with industry will also better \textbf{enable academia to have a deep impact on the practice of modern AI }and to truly appreciate what problems are important in practice. Partnerships with industry to leverage interdisciplinary expertise and industry resources should be encouraged, where appropriate. For example, MPS can drive specialized models that utilize symmetries within the data, train small models, develop novel architectures, and use physics-based compression to reduce the memory footprint (and model size). Such contributions (many of which are described in more detail in \Secs{sec:ai-innovations}{sec:techniques}) should be communicated to industry to forge relationships and demonstrate the value of curiosity-driven research to industry efforts.

Developing \textbf{pathways and incentives to create a flow of talent between industry and academia} could include supporting ``reverse sabbaticals'' to rotate industry expertise in and out of academia. It could also include establishing fellowships specifically geared toward moving between data and domain science, increasing career mobility for students and postdocs. Establishing joint grant opportunities that require co-PIs from academia and industry and/or funding for PhD students and postdocs jointly with industry labs could also help facilitate industry collaborations with MPS.

\subsection{Pursue the Science of AI}
\label{sec:science-of-ai}

\glsresetall

MPS scientists excel at distilling insights from complex systems---such as neural networks---by uncovering underlying principles and emergent behaviors. They also have had to come up with innovative techniques to use AI methods on datasets that are substantially more sparse or noisy than the datasets used to develop the methods. Research in this direction is not currently considered a core part of the MPS portfolio, though, leading to few opportunities for appropriate funding and support. We advocate that funding research towards developing a \textbf{``Science of AI'' would advance efforts toward powerful and robust AI models} even beyond scientific applications.

Across MPS domains, there are significant opportunities for collaboration and progress within the framework of the ``Science of AI.'' 
Researchers can work together to adapt existing AI tools, leverage scientific frameworks to better understand how and why AI works, and drive the development of new tools that integrate scientific rigor. These efforts would not only help to penetrate the ``black box'' of AI, but would also help to develop AI tools that are embedded from the beginning with scientific principles, making them more robust for scientific applications. By leveraging the deep expertise of the community in uncovering fundamental principles from complex systems,\textbf{ MPS is well-positioned to lead the way in developing AI techniques that are rigorous, interpretable, and firmly rooted in scientific understanding} \cite{Krenn2022}.

\subsubsection{AI Innovations from Science}
\label{sec:ai-innovations}

Because there are often requirements in scientific domains for models to adhere to fundamental laws, or to adhere to limitations in the data, MPS researchers have developed models that are infused with scientific principles, which have the potential to ultimately improve AI. Building off of the examples from \Sec{sec:MPSadvanceAI}, opportunities include:

\begin{itemize}
\item \textbf{New AI architectures:} By creating novel mathematical and statistical frameworks that make models transparent and robust, as well as infusing physical principles in AI methodologies, MPS can have a substantial impact on the development of AI. This includes systematically embedding physical laws, such as symmetries, or intentionally modifying physical laws to improve model speed and efficiency. In addition to improving upon existing AI models, MPS researchers are developing novel algorithms to solve scientific problems that can then be brought back to AI for broader applications, including models with learnable activation functions, custom pooling operations, and neurosymbolic elements, as well as optimization strategies based on physical behaviors. 
\item \textbf{Efficient computing:} AI models often require substantial computational resources and large datasets for training. This high demand can lead to inefficiencies, especially when models are applied to complex scientific problems where data may be scarce or expensive to obtain. Often, MPS scientists are using unique state-of-the-art instruments/observatories for which access is extremely limited and competitive.  Developing algorithms that are “data efficient” (not just wall-time efficient or energy efficient) enables AI to have a meaningful impact on many of the most pressing problems in MPS domains. In general, developing more efficient algorithms is a widely active research area and is critical to improving sustainability as AI is scaled up (see discussion of data efficient methods as a key research opportunity in \Sec{sec:data-efficient}).  MPS researchers are also contributing to alternative computing platforms (see \Sec{sec:cost-efficient}).
\item \textbf{Physics-based AI simulations:} Generative AI models are fast for simulations, but they do not inherently adhere to fundamental physics laws.
Building physics concepts and constraints into generative AI has the potential to improve these tools and increase the realism of their outputs.
Physics-based AI models have applications ranging from video generation to fusion reactor designs, and they are also relevant within MPS to enable fast inference (\Sec{sec:sbi}) and multi-scale simulations (\Sec{sec:simulation}).
\item \textbf{Raising the bar:} True scientific creativity requires out-of-the-box thinking. To improve AI's scientific taste requires developing techniques that force AI to make unlikely connections from the data and to recognize when a result is both new and impactful. Benchmark problems within the MPS domains can be used to train AI reasoning systems to help identify such connections in way that makes sense scientifically (see further discussions of AI co-pilots in \Sec{sec:ai-co-pilot}). 
\end{itemize}

\subsubsection{Understanding AI Behaviors}
\label{sec:understand-ai-behavior}

AI methods are particularly effective at handling large and complex data sets and nonlinear relationships, which can be difficult to interpret and therefore lead to general distrust in their conclusions. In the pursuit of verifiable insights, MPS researchers can \textbf{work across disciplines and with computer scientists to understand how AI methods make decisions.} In addition to infusing scientific insights into AI, as described above, we can use scientific principles to better understand the behavior of AI, improve trust and usability, and develop a theoretical understanding of learning.

Again building on \Sec{sec:MPSadvanceAI}, opportunities include:
\begin{itemize}
    \item \textbf{Apply scientific frameworks to answer AI questions.}  While empirical studies of AI behaviors are valuable, MPS researchers can leverage experience analyzing physical systems to develop novel ways to understand AI phenomena. For example, analytical physics techniques have been used to explicitly calculate learning curves, i.e., the typical training and generalization errors as a function of the number of training examples, for specific one- and two-layer neural networks; 
    these calculations demonstrated that learning can exhibit phase transitions from poor to good generalization.  Further, understanding the origins of such transitions can inspire strategies to avoid undesirable behaviors.
    \item \textbf{Extrapolate from behavioral patterns of agentic systems.} 
    Agentic AI systems---which autonomously observe, reason, and act to achieve specified goals---exhibit complex emergent behaviors that parallel self-organizing systems studied across physics and mathematics. These models integrate multiple capabilities---including planning, tool use, memory management, and adaptive decision-making---creating rich dynamics similar to phase transitions that can be analyzed through frameworks from physics.

    \item \textbf{Perform interpretability experiments on black-box models.}
    While we generally advocate for incorporating physical principles into AI systems, there is also value to understanding whether black-models without physical priors can nevertheless ``learn'' physical intuition, or even identify missing physics.
    Comparing the performance of models with and without priors can also yield insights into the tradeoff between domain knowledge and training dynamics.
\end{itemize} 

\subsubsection{Robust and Reproducible AI}
\label{sec:robust-and-reproducible-AI}

MPS researchers are concerned about AI’s heuristic nature and its potential lack of the scientific rigor required in domain science. 
Even in cases where AI appears to be successful, there is often a lack of transparent comparison with existing methodologies, making it unclear if an AI tool is the ``best'' way to solve a given problem, or if a more classical tool would have been more effective.  

Many current AI models lack robustness (e.g., \glspl{LLM} can hallucinate) and reproducibility (e.g., models do not necessarily work across multiple domains or applications). While smaller, domain-specific models---when paired with sufficient domain knowledge---have historically been considered especially valuable tools for advancing science, infusing science into general AI models can lead to robust and reproducible AI that could be used across domains.
Targets in this area include the following.  
\begin{itemize}
    \item \textbf{Develop systematic verification procedures.}
    Instead of performing \emph{post hoc} verification of AI outputs, 
    one can build AI predictions models that incorporate explainability by design.
    For example, one could first define concepts based on the explanations to be made, and then make predictions based on those concepts.
    This kind of hierarchical prediction approach may also lead to reusable network components if problems share the same underlying concepts.
    \item \textbf{Expand the scope of \gls{UQ}.} 
    To meet the standards of scientific rigor, it is essential to robustly quantify uncertainties in AI-based approaches and assess out-of-distribution behaviors. 
    This is especially important for applications that involve inference of physical model parameters (see \Sec{sec:sbi}).
    AI offers the potential to enhance the ways that systematic uncertainties are identified and quantified, for example, by training ensembles of models on simulations that scan over a range of settings and stress-testing models by training on one dataset and testing on another (see further discussion in \Sec{sec:uncertainty_quantification}).

    \item \textbf{Keep mathematical understanding and rigor central to the scientific process.}
    Many AI techniques are grounded in mathematical and statistical principles (see \Sec{sec:dms}), so there are opportunities to develop AI systems with provable guarantees of performance and robustness. These systems have the potential to enhance our ability to build and reason about scientific and mathematical models, but it is essential that researchers maintain rigor in the scientific process.
\end{itemize}
Reproducibility can also be addressed by curating AI benchmarks, as discussed in \Sec{sec:benchmark}.

\subsection{Establish Scalable AI Infrastructures}
\label{sec:infrastructure}

\glsresetall
With the exponentially increasing use of AI, \textbf{the need for scalable infrastructures is crucial}. AI infrastructure comprises the hardware, software, networking resources, and tools to develop, deploy, and manage AI tools and models. The components of AI infrastructure are highly multifaceted, including accessible resources, data management and life-cycle tools, platforms for data discoverability, protocols to make data AI-ready, distributed data systems for caching and connecting data to computing, and software, middleware, and management tools for models (i.e., models-as-infrastructure). Centralizing the infrastructure could increase access, provide support systems, and align expectations across disciplines for better reproducibility (as discussed in \Sec{sec:robust-and-reproducible-AI}). Existing resources like NSF ACCESS \cite{ACCESS}, NSF CloudBank \cite{CloudBank},\ and the \gls{NAIRR} pilot \cite{NAIRR} are providing a solid foundation to build upon in this regard, but are relatively nascent. Now is the ideal time to identify what the MPS community's needs are so that solutions can be implemented as part of existing plans.
This section highlights opportunities to build scalable resources for computing, data management, and curated AI+MPS benchmarks.

\subsubsection{Computing Resources}
\label{sec:computing-resources}

Increased access and sharing of high-performance computing resources, such as moderate-scale \gls{GPU} clusters, open-source and open-weight models, and tool repositories, will help offset the growing costs of AI tools. Practically speaking, \textbf{compute will increasingly become a utility} and it would benefit from being treated as such in the scheme of the broader nationwide infrastructure. Some particularly impactful opportunities for computing resources include:

\begin{itemize}
    \item \textbf{Accessible cloud computing:} For centuries, a reliable approach to scientific discovery has been to collect data and analyze this data with increasingly sophisticated methods. Correspondingly, the data and computing requirements of leading edge science have consistently increased.
    While public cloud AI computing resources can meet the requirements of leading edge research computing, they tend to be prohibitively expensive.
    The reason for the cost is that cloud services must meet the confidentiality, integrity, and availability requirements of e-commerce and enterprise computing, which are significantly more costly than those required for fundamental research.
    As a result, the scientific environment has separated into those institutions (often in partnership with large labs) that have made the commitment to provide tailored, more cost-effective research computing and those that have not. The latter will be increasingly unable to conduct leading edge AI research, unless a change is made to how cloud computing is offered. 
    At the same time, cloud computing presents an attractive and flexible alternative to on-premises compute for deployments requiring large compute resources for a short period of time, applications with evolving compute needs, or for users who value the absence of capital costs in favor of a pay-as-you go approach with outsourced maintenance and support. Particular applications popular for cloud compute include training or fine-tuning large machine learning models (e.g., protein language models for deep generative protein design, diffusion models for image generation) that require intensive compute resources for training but are relatively lightweight in deployment.
    \item \textbf{Software development and maintenance:} 
    Many essential software tools for AI, such as for automatic differentiation, numerical linear algebra, and numerical equation solvers, have been developed by the MPS community.
    However, the software and model outcomes of academic research often become inaccessible or inoperable as the personnel who developed them move on to new positions or projects or as funding for development and maintenance ceases.
    Therefore, there is a need for sustained support for researchers and professionals working on technical software development and code maintenance.
    This could include hiring \glspl{RSE} \cite{USRSE}, who work with researchers directly to help improve their software.
    Some labs already invest in \glspl{RSE}, but one could envision a centralized federal or national infrastructure to provide these services, perhaps as an expansion of \gls{NAIRR}. This may complement the opportunities and resources available at other federally supported facilities in providing long-term support to make products from federally supported research available to the community and public in the context of a ``models-as-infrastructure'' paradigm. Centralizing these products within a national resource would preserve indefinite access to these tools.
    \item \textbf{Funding for \gls{API} services:} Currently, grants are rarely structured to accommodate AI \gls{API} services, despite these being functionally equivalent to traditional computing resource expenditures. Equipment/computing allocations as part of grants would help ensure researchers have access to the compute they need. Some positive developments on this front so far include recent initiatives by the NSF and the DOE to establish partnerships with model service providers, both proprietary and \gls{MaaS}, allowing researchers to access \gls{API} services directly through grants.
    \item \textbf{Software/hardware interface development:} There is opportunity for development at the software/hardware interface for parallelization of ML frameworks on \glspl{GPU}. While there are many new algorithms for efficient training on small-scale systems, there is less work on practical implementation and inference on large-scale systems, where leadership-class computing is required. For example, massive parallel training on a national cluster is non-trivial. Though it is not very difficult, because we are not at the industry level, there are still barriers to performing mid- to high-tier training. The recent NVIDIA library for equivariant neural networks \cite{cuEquivariance} is a promising step forward, but more focus on this kind of optimization would make these tools more accessible and scalable.
    \item \textbf{Testing unique resources:} In terms of the resources themselves, AI processors based on existing or emerging computing paradigms such as neuromorphic computing, photonic computing, and quantum computing built on new materials, can lead to better/faster AI that could change the way we deploy AI for many scientific and commercial applications. MPS has unique science drivers that push this technology for somewhat orthogonal scientific (not commercial) purposes, allowing for the cultivation of this work and potential breakthroughs that can go downstream. Providing unique resources can allow for exploration of different AI hardware/design infrastructures that are not yet mainstream.
    \item \textbf{Computing investment:} Although existing NSF initiatives such as ACCESS have made positive strides toward addressing the gap in computing resources, they are not enough to support the rapidly growing computational demands of cutting-edge AI research that is \gls{GPU} intensive. To fully realize the potential of AI for scientific discovery, it would be highly beneficial to advocate for separate, dedicated government funding specifically aimed at significantly expanding academic computational infrastructure in the next several years. Investments in AI infrastructure should be driven by need and demand, not just fixed funding, which may require an assessment of the current and future needs of MPS in this space.  Additionally, the AI infrastructure that is supported needs to be agile  and support a range of users and use cases. Achieving such an increase in computational resources would empower researchers to tackle ambitious projects and enable breakthrough discoveries across various fields and institutions.
\end{itemize}

\subsubsection{Data Management and Access}
\label{sec:data-management}

In addition to cutting-edge computing resources, MPS researchers \textbf{require better data to build models that can perform reliably} enough for scientific applications. Though MPS researchers have the potential to make advances in data-efficient methods (\Sec{sec:data-efficient}), it is only possible to train a robust, generalizable model when there is a reasonably large quantity of high enough quality data. Data quality can be characterized by both universal qualities, like adherence to consistent data standards, well-documented collection protocols, inclusion of negative examples, etc., and task-specific qualities, such as the degree to which data being generated aligns with the task for which the AI model is being developed. As a result, there are several opportunities to improve data management and access for MPS: 

\begin{itemize}
    \item \textbf{Domain-specific data generation:} MPS researchers need support for the generation and free availability of high-quality, domain-specific, and structured data. In chemistry, for example, while significant strides have been made in AI-driven small molecule and protein design, there remain opportunities to further advance fields such as polymers, peptides, hard materials, and mixtures.  A central challenge to this is managing high-quality data---where manually labeled data can be cost-prohibitive, and simulated data, while more affordable, may not fully capture real-world observations. Similarly, the inherent complexity of astronomical data calls for benchmarks and diverse simulation scenarios to test for unknown systematics. On the other hand, self-driving labs have shown success by iteratively generating data through feedback between the model and the experiment in such a way that the data is maximally informative relative to the task. Overall, the large differences in data volume, quality, heterogeneity, and cultural expectations across MPS disciplines highlight the need for robust data management strategies and tailored benchmarking approaches, rather than a one-size-fits-all solution.
    \item \textbf{Centralized data facility:} Individual researchers and individual universities should not become experts in building and maintaining databases. Each discipline (and each subdomain within a discipline) has different expectations for data distribution and access.
    A centralized federal or national data infrastructure could facilitate the development of data-intensive AI systems; this could replace individual data management plans and would provide expertise of staff data scientists that interface with research groups to help bring the data into state-of-the-art formats and standards conforming with best practice and then host the data on dedicated hardware. Such facilities could host centralized data repositories and establish standardized metadata for both experiments and simulations when reporting/sharing results. These measures would require better attribution when people use datasets, and would ensure that data used for sponsored research is collected and maintained. National Labs may be good resources to inform data infrastructure and management, as well as data generation, given their expertise in \gls{HPC}.
    \item \textbf{Curated data archives:} At minimum, the MPS community should invest in archives of well curated data and where possible standardize technologies and access protocols. The contribution of identifying and curating specific data sets rather than just preserving all data (as the current data management plans require) would be a significant benefit to researchers. It is worth emphasizing that ``data'' goes beyond just the the raw data and includes metadata about how the data is organized and processed. In mathematics, for example, data can include the proof of a theorem written in a formal language and contributed to public libraries such as MathLib, which provides new opportunities for AI researchers to develop AI models to assist in theorem proving.
    Often, when researchers implement a novel AI tool, it can become stranded with dependencies on software that is no longer maintained. There is a need for curated datasets with well-defined, stable metadata/interfaces  (see further discussion in \Sec{sec:benchmark}). Otherwise, the cost and time required for data acquisition, especially in small experiments, will be an obstacle to the application of AI for accelerating fundamental discovery in these areas.
    \item \textbf{Public access to data:} Moving toward centralized data that is not held by individual universities would also help democratize AI tools. There are examples of large data activities that have been lost and the data no longer ``exists'' in the public domain. There is a pressing need to make sure that large datasets generated using government funding have long-term storage and support. These need to be curated and freely accessible databases of high quality, structured data (e.g., the Open Reaction Database \cite{OpenReactionDatabase}). Federal support of data structures could use a model similar to the current support of computing centers in ACCESS or \gls{NAIRR}: operated by partners (usually universities) but with long-term funding by federal sources. Sustainability is a key aspect here; compute resources can be quickly rebuilt when there is a funding lapse but data, once gone, cannot be recovered at scale. Greater access to data could play a role in developing better quality models and lower the cost of using AI. There has been a strong focus on FAIR principles (data should be findable, accessible, interoperable, and reusable) for data, but less so on models. Large amounts of time, resources, and expertise go into constructing models as part of MPS projects, but it is laborious, time consuming, and expensive to maintain and update these models to keep them current and operational. There are few incentives for researchers to do so after expiration of an award or departure of the personnel who created the models. A centralized infrastructure and repository with long-term stable funding and the human expertise to maintain these models and make them available for public access long after expiry of the awards that create them would be valuable in preserving access.  
    \item \textbf{AI-ready data and connecting to computing:} The fuel for AI models is AI-ready data and metadata. \Gls{DLM} tools present a means to manage data throughout its entire lifespan from collection/creation, storage, processing, archiving, and disposal. Particular datasets (e.g., human subjects, medical data) may also entail special privacy or security concerns or the preservation of key metadata attributes that can be addressed by effective \gls{DLM}. Indeed, \gls{NAIRR} has provisions for resources and data management, including handling of sensitive data through NAIRR Secure. A key component of \gls{DLM} specific to AI is the processing and archiving of data in machine-readable AI-ready formats and the connectivity of data to compute. This tight integration between \gls{DLM} and compute means that many compute providers and facilities also provide \gls{DLM} solutions in tandem with compute access and resources. The specifics of connecting data to compute is an important issue and solutions vary widely depending on the size, proximity, velocity, and variability of the data. Small data sets may be locally cached whereas large, multifaceted data in distributed storage may require distributed file systems and local transient storage and compute hardware. Infrastructure capable of handling these diverse AI-ready and compute connectivity needs would help accelerate and enable MPS research endeavors.
    
\end{itemize}

\subsubsection{Benchmarking and Reproducibility}
\label{sec:benchmark}

While many MPS domains have achieved state-of-the-art results using AI tools, these results do not typically generalize well to new questions. This is due in part to a scarcity of quality data and computing resources with which to train these models for broad applications. Domain experts who develop solutions often lack the time or expertise to benchmark them for wider use. 
Similarly, the lack of standardized evaluation metrics specific to solving domain problems through AI further complicates meaningful comparisons between traditional and AI-driven methods. While some of this can be addressed by developing standards of rigor within the community, it would also help to improve the quality of AI tools using \textbf{accessible, scalable training data that is augmented with domain knowledge.} This would require: 

\begin{itemize}
    \item \textbf{Documentation:} There is opportunity to develop data hygiene incentives and standards for documenting how AI techniques are used to support knowledge sharing across domains. There are existing tools/best practices that work well enough when the unit tests and benchmarks run on one machine in a few hours, but the large cost of computing and/or data storage required to replicate a training make it impractical to apply standard software engineering tools. To improve reproducibility, researchers could be encouraged to upload the files associated with computation (and associated version of the code), and/or work with \glspl{RSE} for benchmark development (see \Sec{sec:computing-resources}).
    
    \item \textbf{Community standards:} Future MPS work will benefit from community-wide agreement on data standards, policies, definitions, and benchmarks. Even though benchmarks are not perfect, they represent a crucial step toward building trust and credibility. This includes establishing best practices for training AI tools and reporting AI findings, which in turn will help standardize data practices and enhance reproducibility. For example, AlphaFold for protein folding and machine-learned \gls{DFT} potentials are successful use cases because of structural data standards such as \gls{CIF} \cite{CIF} and access to structural databases such as the \gls{ICSD} \cite{ICSD} and the \gls{PDB} \cite{RCSB}.
    \item \textbf{Evaluation frameworks:} For evaluating AI-enabled scientific research, we need frameworks that balance efficiency with the human elements that make scientific discovery meaningful, ensuring that AI-augmented productivity is clearly distinguished from genuine human insight and creativity. Part of this may involve expanding the metrics used for evaluating AI methods from accuracy to other metrics like interpretability, explainability, and minimizing bias out-of-sample, particularly for MPS research.
\end{itemize}
More discussion of community standards and reproducibility is included in \Sec{sec:scientific-integrity}.

\subsection{Facilitate Interdisciplinary Collaborations}
\label{sec:facilitating}

\glsresetall

There are many opportunities for joint AI research projects across MPS domains and with AI experts.
Part of what makes collaborations across MPS domains beneficial is that \textbf{there is often significant overlap in the uses and challenges that integrating AI tools can bring.}
An advancement in workflow or other solutions in one domain can then be applied to other domains to advance their work, such as the way advances in computer vision enhanced the analysis of scientific images. 
In addition, fostering collaborations across domains has the potential to eliminate duplicate efforts across subdomains.

In many ways, AI is a common language between scientific domains that can open lines of communication.
However, many MPS researchers face challenges in initiating interdisciplinary collaborations across domains and with AI researchers.
The subsections below present mechanisms to consider for \textbf{strengthening interdisciplinary collaboration}, while  \Secs{sec:techniques}{sec:conducting-research} present more information about potential directions for such collaborations.

\subsubsection{Knowledge Transfer}
\label{sec:knowledge-transfer}

Learning and communication barriers across scientific domains creates a challenge for facilitating interdisciplinary research.
While there are many resources for starting to learn or implement various AI techniques, \textbf{matching the right scientific questions with the right AI approach and learning the nuances of different approaches} is challenging and has a steep learning curve. 
For example, most paper/blog posts on AI are written by and for computer scientists, which is not approachable for most MPS researchers.
This can result in misunderstandings of what AI can/should do in the sciences.
Integrating AI tools into existing workflows demands significant expertise in both science and AI, creating a skills gap that can be a barrier for widespread implementation.

Therefore, providing targeted opportunities for domain experts interested in using AI for their research to learn about AI and its applications could significantly reduce barriers to entry.
Such opportunities should use a broad definition of AI (as advocated in \Sec{sec:ai-context}) to ensure that researchers have the flexibility to use the techniques most appropriate for their datasets.
For example, the full machinery of \gls{SBI} (see \Sec{sec:sbi}) would not be needed if a researcher is working with a problem where the likelihood is tractable.

There are a number of opportunities to facilitate knowledge exchange: 
\begin{itemize}
    \item \textbf{Recruit cross-disciplinary experts:}  By hiring ``centaur scientists'' (see \Sec{sec:individual}) with dual AI+MPS expertise, institutions can create a culture of cross-disciplinary dialogue;  
    \item \textbf{Conduct trainings:} Workshops and training programs can help familiarize faculty with AI tools (see \Sec{sec:faculty});
    \item \textbf{Establish an AI literacy framework:} The operational aspects of AI tools can be simplified by adopting a modular, stackable framework aligned with the AI literacy pyramid (see \Sec{sec:public});
    \item \textbf{Improve the quality of AI tools:} Community standards (see \Sec{sec:benchmark}) can help curate clean training data; for the models using this data, it will be important to make these tools more accessible, scalable, and to augment foundational models with domain knowledge; and 
    \item \textbf{Develop a shared scientific language:} State-of-the-art AI methods should be expressed as closely as possible to shared scientific language and jargon, so that the transfer of knowledge has minimal barriers. This can be partly accomplished through standards and benchmarking described in \Sec{sec:benchmark}.
\end{itemize}

To effect knowledge transfer, there are often very different outlooks and practices across fields that need to be overcome, such as disparate terminologies, communication styles, expectations, timelines, and methodological rigor. Thus, there is a need to develop frameworks that can systematically bridge between AI's heuristic insights and formal scientific verification---depending on the circumstance, this may mean identifying scenarios in which the science can accept non-interpretable functional relationships, but in many cases it will require developing models that are interpretable. Also, even within MPS, AI applications can be siloed, even though various techniques would be advantageous across a broad range of problems. AI and data science hold promise as potential common frameworks to overcome these barriers, offering a \textbf{unified language and approach that can streamline interactions and facilitate meaningful cross-disciplinary research.} Utilizing AI systems to facilitate cross-disciplinary research and knowledge exchange across traditionally separate domains could go a long way in bridging the gap.

\subsubsection{Workshops and Conferences}
\label{sec:workshops}
    
Community events such as workshops and conferences offer valuable opportunities for AI+MPS researchers to \textbf{collaborate, share resources, exchange insights, and discuss challenges. }It would benefit the AI+MPS community to consider opportunities such as the following: 

\begin{itemize}
    \item \textbf{AI+Science workshops:}
    Rather than only hosting AI+[Domain] events (which can also be very effective), there is a clear need for more interdisciplinary AI+Science workshops.
    For example, meetings that cut across MPS domain boundaries and bring together researchers who are interested in common AI techniques could be organized, either as few-day to week-long events targeting knowledge sharing or as multi-week workshops aimed at fostering collaborations.
    These events should occur on a regular cadence---ideally annually---and be designed to dig into the detailed nuances of AI tools and methodologies regardless of application. By providing focused sessions that explore both successes and failures, these workshops can facilitate a deeper understanding of AI integration in science, encourage innovative solutions, and strengthen interdisciplinary collaboration. 
    \item \textbf{Science at AI conferences:} AI conferences could benefit from broadening their scope to include more science applications, as defined by scientists themselves, rather than narrower computer science perspectives. Co-located AI+Science Workshops would also help to raise awareness about science efforts to the AI community. An established example of this is the Machine Learning and the Physical Sciences Workshop \cite{ML4PhysicalSciences2024}, held at the \acrfull{NeurIPS} for the past several years.  
    Ideally, such workshops would be held regularly, to build a community of researchers who target those workshops to present results. 
    \item \textbf{Coding camps/hackathons}: Organizing coding camps or hackathons to engage the computer science community on problems that are important to MPS is another way to engage across disciplines. There are already examples of such events that have led to successful collaborations, such as the LHC Olympics anomaly detection data challenge \cite{LHCOlympics} and the Photometric LSST Astronomical Time-Series Classification Challenge (PLAsTiCC) challenge \cite{PLAsTiCC}. Improved communication between AI and science researchers would enrich both communities and accelerate progress in AI+Science.
    \item \textbf{National joint working group:} Building community for interdisciplinary researchers and educating the broader scientific community on interdisciplinary research is crucial. One way to facilitate this could be to create a national joint working group of AI researchers and MPS scientists interested in using AI with mutual interests in advancing AI+Science to organize such events and establish communication across disciplines.
\end{itemize}

\subsubsection{Collaborating Beyond MPS}
\label{sec:beyond-mps}

While this white paper is focused on the AI+MPS community, there are also opportunities for collaboration with domains outside of MPS.
These could be related to the topics described in \Secs{sec:techniques}{sec:conducting-research} below, but also beyond, with the goals of:
\begin{itemize}
    \item \textbf{Creating a new language/science to understand AI;} 
    \item \textbf{Generating radically new models of AI;} and
    \item \textbf{Understanding the broad implications of using AI.}
\end{itemize}
For example, researchers are using AI tools in health care and biological engineering, as well as in finance and logistics, to inform strategies for integrating simulation data with experimental measurements.
In addition to scientific research, interactions with jurists and ethicists may be valuable to help define regulations and best practices for the ethical use of AI technologies, and with philosophers to enrich discussions and broaden our understanding of AI's impact (see more discussion of this in \Sec{sec:scientific-integrity}).
While specific ideas for collaborating beyond MPS are outside the scope of this paper, it is important to encourage and pursue such opportunities as AI becomes ever more pervasive and continues to blur disciplinary lines.

\subsection{Cultivate Key AI Techniques for Science}
\label{sec:techniques}

\glsresetall

AI serves as a common language across various subfields within MPS, providing a unifying framework that fosters holistic understanding. 
The advancement of AI can benefit from embracing the varying complexities and practical demands of different fields, encouraging innovation. To that end, \textbf{organizing collaborative efforts around common AI techniques} can be a catalyst for progress in both MPS domains and AI. This section highlights some key techniques that have already proven valuable in multiple scientific domains, making them ideal candidates for bringing interdisciplinary researchers together.
Domain-specific applications of these techniques are described in \Sec{sec:domains}.

\subsubsection{Simulation-Based Inference}
\label{sec:sbi}

Statistical inference is a common task in the MPS domains, where the goal is to determine the underlying parameters of a theoretical model from measured experimental data.
To perform classical statistical inference, one needs to know the \emph{likelihood}, which is the probability of seeing a measured result given the model parameters.
Often in MPS, the likelihood is not known exactly, but we can nevertheless use AI to estimate the likelihood from trusted forward simulations that generate synthetic data from a given model.
This is the basis for \gls{SBI}, which has been used extensively in the sciences \cite{Cranmer2020}.

While \gls{SBI} itself is a statistical framework that predates modern AI, \textbf{contemporary SBI approaches leverage various AI techniques to make the inference process more efficient and scalable}. These AI-powered \gls{SBI} methods include neural density estimation and amortized inference techniques. Deep generative models such as \glspl{NF}, diffusion models, and conditional \glspl{VAE} enable the approximation of complex posterior distributions, while neural networks serve as surrogate models for likelihood-free inference.

In this way, AI is opening new avenues for drawing inference of physical parameters in models so complex that likelihoods are intractable or unknown.
For example, in both nuclear/particle physics and cosmology, \gls{SBI} can play an important role in processing large amounts of data from large experiments such as the \gls{LHC} and galaxy surveys from new telescopes. At the \gls{LHC}, it can be used for triggering, anomaly detection, and surrogate modeling.
In cosmology, it can be used to enable cosmological parameter inference.
In chemistry and materials research, SBI can be used to infer molecular interaction parameters.

The basic framework of \gls{SBI} is well established, but there are ample research opportunities to improve the robustness of \gls{SBI} techniques \cite{SimulationBasedInference}.
For example, the mathematical sciences are developing algorithms for training models to produce simulations for \gls{SBI}, including ones based on optimal transport.
Collaborations between MPS researchers could ensure that these algorithms produced reliable and generalizable results across disciplines.

\subsubsection{Multi-Scale Simulations}
\label{sec:simulation}

AI approaches are revolutionizing multi-scale modeling across MPS disciplines by bridging traditionally separated scales of physical phenomena. In astrophysics, neural networks enable the seamless integration of sub-grid physics models with large-scale cosmological simulations, capturing complex feedback processes that span orders of magnitude in spatial and temporal scales. In materials science and chemistry, AI creates hierarchical models that connect quantum-mechanical calculations at the atomic scale to mesoscale properties and macroscopic behaviors. \Glspl{PINN} and deep operator networks enable the development of surrogate models that preserve physical constraints while drastically reducing computational costs compared to traditional numerical methods \cite{Azizzadenesheli2024}.

Multi-scale AI models offer unique advantages over conventional approaches by: 
\begin{itemize}
    \item learning effective coarse-graining procedures directly from high-resolution simulation data; 
    \item maintaining consistency across scales by incorporating physical conservation laws and symmetries; 
    \item enabling real-time prediction and analysis of complex systems that would otherwise require supercomputing resources; and 
    \item quantifying uncertainties that arise from scale transitions.
\end{itemize} 

These capabilities are particularly valuable for modeling systems where dynamics at different scales are tightly coupled and cannot be easily separated---a common challenge across MPS domains from quantum phenomena to cosmological structure formation. By collaborating on multi-scale modeling techniques, MPS researchers can \textbf{develop shared frameworks for scale bridging, uncertainty propagation, and physics-informed architecture design} that benefit all disciplines dealing with complex hierarchical systems.

\subsubsection{Uncertainty Quantification}
\label{sec:uncertainty_quantification}

Robust scientific conclusions require identifying various sources of uncertainty and quantifying their size.
This includes statistical uncertainties associated with sampled data as well as systematic uncertainties from imperfect modeling. Unlike classical statistical methods that provide well-defined error bounds, off-the-shelf deep learning models often yield overconfident predictions without robust uncertainty estimates.
This limitation is especially critical to bear in mind for inference problems where propagating measurement uncertainties through complex models is essential for drawing valid scientific conclusions.
Approaches like Bayesian neural networks and ensemble methods are promising ways to address these issues, but they still face computational challenges that hinder full uncertainty propagation for large-scale models.
Advancing the frontiers of \gls{UQ} can therefore have a broad impact, for example by providing opportunities to address uncertainty in AI models, enabling ``decisions that are directly informed by \gls{UQ},'' and pushing \gls{UQ} toward near real-time applicability. 

AI’s potential to model high-dimensional, complex probability distributions can help scientists quantify and understand uncertainties of predictions. 
Specifically, \textbf{AI can handle many more nuisance parameters than traditional methods, enabling more robust \gls{UQ}}.
In addition, AI ensemble methods can compare results from multiple models to see how different assumptions affect the resulting scientific conclusions.
Collaborating on \gls{UQ} can lead to advances that are important across MPS domains, such as validating models, assessing errors in experiments and simulations, identifying systematic errors to ensure consistency across equipment and software, and ensuring reliable predictions.

The design of successful scientific experiments also depends on accurate \gls{UQ}. For example, Hertz’s experiments demonstrating the correctness of Maxwell’s equations, Lavoisier’s seminal discoveries in chemistry, the Michelson-Morley experiment, and many others all used extremely careful experimental design, which required accurate \gls{UQ}. Looking to the future, a key theoretical challenge in designing and building practical quantum computers at scale is accurate \gls{UQ}, which could be improved using AI. 

\subsubsection{Foundation Models}
\label{sec:foundation_models}

Generative AI presents a common opportunity across scientific domains to enable high-fidelity, fast simulations, serving as a replacement for compute-intensive methods where feasible, and supporting scalable synthetic dataset generation. Given these opportunities, it is important for MPS researchers to \textbf{develop generalizable foundation models that are domain-informed.}

Foundation models, with domain-informed training, can make significant advances in data analysis.
This has begun to be explored for big data applications in astronomy and physics, though infrastructures needs to be developed to handle increasing amounts of data (see \Sec{sec:infrastructure})---such as in preparation for a new generation of telescopes in astrophysics.
Fine-tuned foundation models have been tested for mathematical proof generation, and there are relatively recent applications of molecular and biological foundation models.
The MPS fields may benefit from better integration and collaboration with each other to build on the advances they have made so far (see \Sec{sec:facilitating}).
Importantly, we need to build models that can robustly analyze multimodal data, which are ubiquitous in the sciences.

Foundation models pre-trained on large datasets could have major downstream applications that cross disciplines. Creating such models would be most successful as a community effort that embeds knowledge from the sciences and is hosted on dedicated hardware. Such large AI systems could enable breakthroughs in applications with limited physical models and small, noisy data sets (e.g., studying chemical reactions at equilibrium or construction defects in additive manufacturing).
Such scientific foundation models might also form the backbone for agentic/reasoning approaches (see \Sec{sec:ai-co-pilot} for further discussions of  AI co-pilots).

\subsubsection{AI for Experimental Control}
\label{sec:rl_for_experiment}

AI-enabled workflows that incorporate human-in-the-loop decision-making offer \textbf{transformative potential for experimental design, optimization, and control} across the MPS portfolio.
These workflows can enable experiments to adapt dynamically to incoming data, efficiently explore complex parameter spaces, and operate with increasing autonomy under expert guidance.
For example, in AST, \gls{RL} is being used to improve the adaptive optics to directly image exoplanets.
By supporting tasks such as experimental shifts and adaptive planning, they lay the foundation for AI-assisted systems that accelerate discovery and maximize the impact of limited resources  (see \Sec{sec:self-driving-labs} for further discussions of self-driving labs).

Adaptive decision making is particularly relevant in the small data regime.
Small-scale experiments in universities, government labs, and industry are typically built with emphasis on flexibility of design but at the cost of requiring the optimization of an enormous number of control parameters.
Historically, this optimization was performed solely by skilled researchers, often with years of accumulated intuition.
AI-enabled experimental control offers potential gains in efficiency for achieving research targets in learning rate and in solution efficacy.

Beyond experimental control, adaptive AI approaches can also be used to find patterns in massive scientific datasets from space telescopes, the \gls{LHC}, and other big data facilities.
Bayesian optimization, active learning, and \Gls{RL} are flexible frameworks for exploring search spaces too large for humans, which can be used to ``find the needle in the haystack'' and support discovery.
Such AI techniques also have the potential to discover new drugs and materials and even construct counter examples for mathematical conjectures.

\subsubsection{Data-Efficient Methods}
\label{sec:data-efficient}

Many MPS applications involve small data sets and bespoke experiments, particularly in chemistry and materials research.
Thus, there is a need to \textbf{develop the mathematics and data science to support AI methods for small datasets}.
This includes experimental selection, \gls{UQ}, and management of systematic errors, with a natural extension to autonomous experimentation (see \Secs{sec:rl_for_experiment}{sec:self-driving-labs}).

It is important to cultivate AI techniques that focus on making the most of the limited datasets available, such as pretraining/transfer learning, multimodal learning, data cleaning, and prediction models for decision making.
One possibility for developing data-efficient methods is to take AI toolsets being built for the big data regime and then adapt them to work well on smaller datasets, applying a lens more similar to classical \gls{ML} and statistics.
Another possibility is to develop methods designed to generalize beyond where good data exists, for example by building in known physical laws and symmetries, so that AI models are constrained to a subspace of plausible models.
It would also be beneficial to develop computationally efficient methods for Bayesian deep learning, which promise optimal accuracy and optimal robustness for any amount of training data, however small. \Sec{sec:robust-and-reproducible-AI} also discusses some opportunities in this direction.

Given a fixed computing budget, model performance can scale more favorably if training data is collected in a task-specific way.
Thus, there are interesting tradeoffs to explore between the cost of computing and the cost of data acquisition.
For tasks where data acquisition is relatively rapid and cheap (e.g., high-throughput screening), it may be more beneficial to invest in data generation rather than compute.
For problems with very small data sets, further investment in data collection may be required before it is wise to start investing significant compute time into analysis.
Correctly identifying the required tasks and characterizing these tradespaces could provide increased flexibility in their solutions.

\subsubsection{Additional Research Opportunities}
\label{sec:research-opportunities}

In addition to the six key AI techniques described above, other examples of cross-cutting AI+MPS research opportunities are as follows.
\begin{itemize}
     \item \textbf{Inverse problems:} 
     \gls{SBI} for parameter estimation (\Sec{sec:sbi}) is an example of an inverse problem, where one knows how to go from ``cause'' to ``effect,'' but going from effect to cause is challenging (and possibly ill-posed).  Solving inverse problems is a generic challenge that AI could help address, for example to model open systems with irretrievable information loss.

    \item \textbf{Anomaly detection:}  In cases where the forward model is known, \gls{SBI} can be used for statistical inference.
    If there anomalous features in the data, though, alternative approaches are needed.
    AI can be used to detect (and interpret) anomalies in data, though further research is needed to ensure that these approaches are sufficiently robust for scientific applications.

    \item \textbf{Heterogeneous data:}
    Many MPS problems involve heterogeneous datasets, which could differ in type (e.g.\ images versus time series), fidelity (e.g.\ low versus high resolution), or origin (e.g.\ broad survey versus targeted follow-up).
     AI could assist in handling heterogeneous data and making reliable predictions from them.

    \item \textbf{Balancing physical insights:}
    By incorporating physical priors, one can compensate for missing or noisy data, which is beneficial for data-efficient methods (\Sec{sec:data-efficient}).
    Physical insights such as symmetry equivariance are also helpful when building multi-scale simulations (\Sec{sec:simulation}).
    However, there is a balance between strict adherence to physics principles and the need for flexible modeling.
    Further AI research is needed to find the optimal balance between model accuracy and physical insight.

    \item \textbf{Improving simulations:} 
    Simulations are essential for statistical inference (\Sec{sec:sbi}) but they can be computationally expensive and not always responsive to experimental information.
    AI could help build surrogate models to speed up and scale up complex simulations.
    It would also be interesting to develop methods to fuse experimental data with simulation data in a way that is more robust than ad hoc corrections.

    \item \textbf{Accounting for bias:}
    There are many ways for bias to affect AI models, including beneficial cases like inductive bias where domain knowledge is used to make AI predictions more reliable.
    As an extension of \gls{UQ}, it would be helpful to find ways to account for potential sources of bias in AI-based approaches, in order to ensure more robust scientific discoveries and increase trust in AI techniques (as discussed in \Sec{sec:science-of-ai}).

    \item \textbf{Transfer learning and domain adaptation:}  Transfer learning allows knowledge gained through one task or dataset to enhance model predictions on another related task and/or different dataset.
    Typically, transfer learning requires some amount of domain adaptation; for example, adapting a model trained on simulated data to perform well on real data.
    Making transfer learning more reliable would benefit AI+MPS applications with a variety of data types.

    \item \textbf{High-throughput experiments:} 
    One strategy to move beyond the small data regime is to collect data through high-throughput experiments, possibly in the context of a self-driving lab (\Sec{sec:self-driving-labs}).
    AI could be used to help design high-throughput protocols.
\end{itemize}
As the MPS domains gain more experience with AI methods, we anticipate many further cross-cutting research opportunities.

\subsection{Leverage AI for Conducting Research}
\label{sec:conducting-research}

\glsresetall

Beyond AI+MPS as a means to advance discovery and understanding, AI can also be used as a tool for streamlining the way research is conducted.
AI can support \textbf{automation, synthesis, and even at times act as a thought partner in hypothesis generation and scientific workflows.} Using AI for this purpose still requires MPS domain expertise, but the tasks themselves could be developed generically across disciplines as a community effort. 

At a high level, AI can enhance the scientific workflow by streamlining background research, efficiently collecting useful data, automating design of experiments, and rapidly testing and validating or refuting hypotheses. AI can help translate research literature, reduce language barriers, and streamline education. In these ways, AI has the potential to accelerate scientific research.

\subsubsection{AI Co-Pilot}
\label{sec:ai-co-pilot}

The National Academies identified literature search and complex hypothesis generation as two grand AI challenges for Science \cite{NationalAcademies2024}. These are two of the use cases for developing ``AI co-pilots,'' or an interactive AI that assists with performing complex tasks. A key opportunity to support research is to \textbf{develop AI-based literature search and visualization} to instantly return data in tabular, graphical, and/or textual form. This could include AI-based searches that assemble novel databases from across the literature for input into AI-generated code. In addition, \textbf{synthesis/visualization tools that apply causal AI methods} for retrieving structured/unstructured data to create a logically consistent ``worldview'' from the literature corpus can aid scientific discovery. The latter has the potential not just to verify or exclude hypotheses but also to generate ideas for new ones.   

Months or years of painstaking work are currently required to create an original and reliable plot from the scientific literature, to track the progression of key measurements and concepts within or across disciplines, and to tie disparate findings into a coherent synthesis that generates new hypotheses. \textbf{A reliable search, synthesis, and visualization system could enable automated hypothesis generation}, identifying causal claims from structured data or phrases such as ``is driven by'' and applying causal AI methods to discover new causal relationships between variables. Agentic AI is making such goals feasible. By constructing \glspl{KG} linking, for example, molecular properties and behaviors, AI can identify overlooked connections to expose blind spots in the literature with humans in the loop. 

The ability of AI to generate plots and synthesize text could be valuable in \textbf{supporting scientific communication} as well. As essentially a ``neutral observer,'' agentic AI can help researchers translate technical content for a broad audience. It can also help to save time with tasks where content is often being repeated, such as reports (like this one).
Finally, generative reasoning models now allow scientists to create detailed prompts to return code and implementation strategies. As such, AI can \textbf{speed up hypothesis testing and validation}. Using AI in these ways can help researchers to more quickly move on to conducting the research. There is immense interest in applying AI for reasoning tasks in the MPS domains, and this is the first step that could be expanded for additional uses.

\subsubsection{Self-Driving Labs}
\label{sec:self-driving-labs}

The integration of AI tools with automated robotics has driven a surge of interest in automated robotic science or ``self-driving labs.'' In self-driving labs, the AI ingests data and determines what experiments to conduct to advance a scientific objective. The AI then controls the robotics to conduct the experiment and return the data. In the CHE and DMR fields, which often operate in the small-data regime, the ability to collect additional data, refine the models, and advance inquiry within a virtuous design-build-test-learn cycle has long been recognized, and \textbf{self-driving labs present a means to accelerate this process through automation}. 

The success of self-driving labs typically rests on very tight coupling between the domain science, AI/data science, and robotics, and having \textbf{a team with interdisciplinary expertise is vital}. Cultivating this cross-talk with AI as a common language (as discussed in \Sec{sec:facilitating}) is an important bridge. It is also important to consider the transferability of a self-driving platform across different scientific questions and applications. Platforms can be brittle and overly tuned for one particular application. The development of improved modularity and flexibility from the ground up is of strong interest to the domains, as is engineering a level of modularity and robustness to permit their integration into industrial workflows. Of course, a future with more self-driving labs or otherwise automated experimentation demands hands-on lab experience for the users even more as they need to establish intuition for how experiments work and go wrong in order to effectively use these systems.  

\subsubsection{Digital Twins}

Digital twins extend the idea of a computational simulation (see \Sec{sec:simulation}) by incorporating feedback loops to synchronize the simulation with its real-world counterpart and even perform real-time actions in response to model predictions.
Digital twins could be particularly useful for MPS research applications where there is a significant cost to performing an experiment. 
For example, digital twins are being explored to improve the operation of particle accelerators, where the they can help maintain performance as operating conditions drift and new parameter regimes are explored.

To the extent that the underlying simulations are reliable, digital twins offer a valuable sandbox for hypothesis testing and optimization, where one can explore many more perturbations (e.g., changing boundary conditions, materials compositions, or control parameters) than possible with physical devices.
If the digital twin incorporates \gls{UQ}, then the model's uncertainty estimates can be used to prioritize more fine‑grained simulation or targeted measurements, thereby concentrating resources where they would be most informative.
Incorporating digital twins into MPS workflows will likely require rethinking the way experimental devices are designed, though, to prioritize additional sensors and interactive controls over robust steady-state operations.

\subsection{Educate and Train an AI+MPS Workforce}
\label{sec:education}

\glsresetall

Building an AI-literate workforce is critical to the nation's success, particularly in MPS fields that have historically driven major technical and industrial advances across society. It is essential to \textbf{cultivate opportunities for leveraging and developing AI across all career stages}, such that it becomes a robust tool that researchers can effectively deploy for solving complex problems. The economic implications for integrating AI and education are profound. Job postings increasingly require AI competencies within engineering and scientific fields, and it is expected that the majority of companies worldwide will adopt AI-driven solutions in the coming decade. This widespread adoption creates significant demand for professionals skilled in applied AI across multiple domains. 

In the MPS fields, AI literacy includes both training in the use and development of AI methods for scientific research and leveraging AI to support teaching and learning. In the subsections below, we outline opportunities for MPS researchers at various career stages to move toward these AI literacy goals. Updating undergraduate and graduate curricula, providing compute and data resources to students, and investing in continuing education for established researchers are all necessary steps. However, the uneven distribution of resources for AI training could exacerbate existing gaps across institutions and between socioeconomic classes if not carefully managed. This underscores the importance of ensuring broad access to AI education, allowing us to bootstrap to a future where AI is thoughtfully integrated within MPS education. As such, we also discuss the importance of K-12 and public education in helping to create a pipeline of AI-literate students who can substantially impact the future of the MPS fields.

\subsubsection{Faculty and Leadership Training}
\label{sec:faculty}

Integrating AI into academic systems will require engaging with faculty and  senior researchers, particularly with those who have not already incorporated AI into their research and teaching.
Effective engagement with academic leadership can help overcome existing hesitations about new methodologies, create pathways for broader adoption, and encourage further AI innovations. Key considerations for faculty and leadership training are presented below.
\begin{itemize}
    \item \textbf{Supporting interdisciplinary faculty:} Institutions have an opportunity to evolve their support for interdisciplinary faculty, particularly regarding tenure and promotion evaluations. Traditional assessment frameworks can be augmented to better recognize interdisciplinary research, especially work that integrates and advances AI methodologies. Developing clear criteria and institutional policies that adequately value and reward interdisciplinary scholarship---within both tenure evaluations and funding panel assessments---would motivate more cross-disciplinary initiatives. 
    \item \textbf{University hiring:} Universities can strengthen AI+MPS development by recruiting innovative AI pioneers in the MPS domains who can advance genuine interdisciplinary work.
    This would involve recruiting ``centaur scientists'' (see \Sec{sec:individual}) who are committed to conducting state-of-the-art interdisciplinary research in AI+MPS.
    While perspectives vary on the ideal balance between AI expertise and domain knowledge that institutions should adopt, there is agreement that there should be a shift away from only conducting faculty searches for experts deeply embedded in their domains.
    By hiring AI+MPS faculty who can train the next generation of centaurs, universities can help cultivate an AI-enabled workforce in \gls{STEM}.
    \item \textbf{Faculty upskilling:} Innovation in AI presents an exciting opportunity for upskilling senior researchers to advance AI integration. MPS researchers often face time constraints when seeking to effectively integrate AI tools into existing workflows. One way to address this challenge is to develop modular, stackable training programs that can accommodate limited bandwidth while delivering focused skill development. Professional development workshops, sabbaticals in AI/ML groups, personnel exchanges, and incentives for AI/ML certifications also offer promising pathways for engaging and upskilling faculty and other senior researchers.
    Strategic incentives, including grants or financial support to enhance courses with AI components or enable faculty to expand their expertise, can accelerate this progress. For example, the MPS fields could work with professional societies to develop standardized guidelines for crediting and encouraging AI-enabled work.

    \item \textbf{Directly address AI concerns:} 
    Some faculty resist the use of AI due to legitimate concerns that AI tools might not adequately respect fundamental scientific laws, properly calibrate uncertainties, or function effectively outside narrow regions of applicability; see \Sec{sec:scientific-integrity} for further issues related to scientific integrity.
    These AI concerns should be addressed directly through AI+MPS events like colloquia and workshops (\Sec{sec:workshops}), and training programs.
\end{itemize}

\subsubsection{Postdoctoral Training and Research Scientists}
\label{sec:postdocs}

As part of the career pipeline being built for students with interdisciplinary credentials, \textbf{postdoctoral programs and research scientist positions that specifically leverage interdisciplinary expertise will be crucial} for maintaining global leadership in the academic workforce. 

\begin{itemize}
    \item \textbf{Postdoctoral fellowships:} As discussed in \Sec{sec:individual}, funding postdoctoral fellowships at the intersection of AI and domain science will attract interdisciplinary experts from the next generation of talent who likely already have experience connecting communities. Within such a framework, it is important to provide infrastructure and support to help ensure the success of these early-career researchers and make sure they are attractive for their career prospects. Existing fellowship programs, such as the \acrshort{IAIFI} Fellowship, have established a dual mentoring system, providing one AI mentor and one domain mentor. This way, the Fellow has guidance and an advocate in both directions.
    \item \textbf{Industry support:} Postdoctoral positions co-sponsored with industry represent another valuable opportunity. These joint positions would increase connections between academic and industrial research communities, enhance career options for early-career researchers, and facilitate knowledge transfer between sectors. Such programs could be structured to allow researchers to spend time in both environments, developing skills and networks that span traditional boundaries.
    \item \textbf{Research scientists:} Research scientists across academic campuses face similar challenges to faculty---limited time availability coupled with the need for continuous upskilling as AI technology rapidly evolves. These professionals, who are past the classroom experience but integral to research productivity, require specialized training opportunities that accommodate their constraints while delivering practical skills. Developing focused workshops and online modules specifically for research scientists would strengthen the research ecosystem's capacity to integrate AI methods effectively.
\end{itemize}

\subsubsection{Graduate Education}
\label{sec:graduate}

The traditional model for graduate education in the MPS domains may require structural changes, as the ability to collaborate across disciplines becomes increasingly important in the AI era.
In addition, AI can enable more flexible, modular training approaches that can be assembled into an education that is tailored to the needs of individual students.
Graduate training must \textbf{go beyond teaching students how to use AI tools}; it must develop their ability to identify solutions using AI tools and recognize opportunities for AI development. This expanded focus supports more diverse career pathways and prepares students for both academic and industrial environments.

\begin{itemize}
    \item \textbf{Interdisciplinary PhD programs:} More PhD programs designed specifically for AI+Science would substantially improve training and certification opportunities for graduate students. Truly integrated and cross-disciplinary programs can help students gain a broader perspective. AI courses have traditionally been taught by \gls{CS} departments, so they often lack domain-relevant examples and face significant capacity limitations that make them difficult for non-CS students to access. Domain scientists could develop and teach specialized AI courses tailored to their specific scientific disciplines, which would ensure relevance and appropriate application within their fields and allow for emphasis on core competencies---understanding domain knowledge in MPS should remain the priority. These core competencies are crucial to help students develop the capacity to recognize characteristics that make a given problem more suitable for AI solutions rather than ``classical'' approaches, and to review AI results critically for issues like hallucinations and computational costs. This problem identification skill is essential for research innovation and the effective application of AI tools.
    
    \item \textbf{Graduate certificates:} Graduate training would benefit from implementing stackable credentials with modular training programs and establishing a well-defined AI literacy framework with clear competency benchmarks (described in more detail in \Sec{sec:public}). Currently, most stackable programs are implemented at the master's level, where curriculum flexibility makes it easier to create customized educational pathways than it is in traditional PhD programs. Expanding these options to the PhD level would provide students with recognized credentials while allowing them to build expertise aligned with their specific research interests and career goals.

    \item \textbf{Graduate admissions:} Student evaluation and admission criteria would benefit from implementing a strategy to identify future interdisciplinary leaders. Publication metrics at the undergraduate level may no longer adequately reflect students' potential capabilities in interdisciplinary AI+MPS research. Instead, students who demonstrate mastery across multiple domains through rigorous coursework may prove particularly adaptable and successful in this evolving research landscape, regardless of their publication records.

    \item \textbf{Ad hoc training:} Beyond PhD programs, the demand for AI literacy is high across multiple professional levels. One opportunity to address this would be to create online master's programs in AI+Science, with credits that could be transferred to a university. In addition, summer schools and other less formal training opportunities are important to help keep students up to date on the latest tools and research. Funding and opportunities for network exchanges between institutions could also be beneficial, as expertise in AI+Science remains dispersed. These exchanges would enable students to gain exposure to diverse applications and methodologies while building collaborative networks that span traditional disciplinary boundaries.
    
    \item \textbf{Industry experience:} 
    Developing formal partnerships with industry for internship programs would provide students with practical experience in applying AI within commercial settings while building valuable connections between academic and industrial research communities. Such programs could be based on opportunities that already exist at many universities, structured as summer experiences, cooperative education semesters, or year-long research sabbaticals. Universities and funding agencies could incentivize these connections through matching funds or dedicated fellowship programs for students pursuing cross-sector training in AI+Science. 
\end{itemize}

\subsubsection{Undergraduate Education}
\label{sec:undergrad}

Undergraduate education faces both challenges and opportunities in the AI era. Students who effectively leverage AI tools will have significant advantages over those who do not. This raises fundamental questions about which skills and knowledge should be emphasized in modern curricula, and which traditional elements might be reconsidered. For instance, as \glspl{LLM} and programming co-pilots become increasingly sophisticated, the approach to teaching programming and domain-specific knowledge may need reevaluation. A valuable investment for the MPS community would be to \textbf{study how best to use AI in teaching}, including how to incorporate AI skill development while still fostering MPS core competencies.

\begin{itemize}
    \item \textbf{Undergraduate curriculum:} Students would benefit from developing proficiency in coding, mathematics, and \gls{ML} within their engineering and science coursework; as such, AI and computational skills should be integrated directly into the undergraduate curriculum. This approach ensures wide access to essential skills while recognizing the practical constraints many students face. 
    It is increasingly important for undergraduate curricula in MPS domains to incorporate probability, statistics, data science, and computational science as complements to traditional calculus-based education. It is also important to integrate AI in the context of domain-specific problems already at the undergraduate level. Technical skills should be accompanied by training in responsible use and ethical considerations surrounding AI applications. Developing critical thinking skills is essential, as students, especially undergraduate students, tend to place excessive trust in AI outputs and must learn to maintain appropriate skepticism. As routine coding tasks become increasingly automated, it is imperative to emphasize in classes a combination of conceptual understanding of the mathematics underlying ML algorithms and a practical ability to interpret computer output. This requires expanding the amount of rigorous mathematics/statistics that students interested in AI need to learn and developing new courses that mix theory and practice. 

    \item \textbf{Teaching approaches:} Undergraduate (and graduate) education must balance two complementary approaches: traditional teaching of fundamental concepts through simplified examples that build deep understanding of underlying principles and core competencies; and AI-assisted exploration of complex problems that develop generalization skills. These approaches support different but equally important skill sets for the future workforce. AI technologies themselves present opportunities for enhancing the educational experience, from automating assessment through \gls{LLM}-proctored examinations to personalizing learning through interactive engagement with recorded content. These applications can free faculty from routine tasks and allow greater focus on high-value educational interactions with students, such as flipped classroom models that emphasize in-classroom work over take-home assignments. Research into pedagogical approaches that incorporate AI tools while maintaining educational integrity will be essential for developing best practices and evidence-based methodologies. It is also important for institutions to recognize the challenges that AI can bring to teaching and classroom evaluation and provide guidance to faculty for addressing these challenges.

    \item \textbf{Student use of AI:} In order to expand their skill sets, undergraduates will need access to public domain AI tools, computational resources including \glspl{GPU} and \gls{API} services, and relevant datasets. Students should learn to critically analyze and implement these tools through hands-on experience. Currently, there is a lack of established pedagogical practices for effectively using AI in education. This represents a critical research opportunity---we must systematically evaluate how AI tools can enhance rather than replace critical thinking in education, developing evidence-based approaches that maintain rigor while leveraging AI's capabilities for tailored education and augmented instruction.

    \item \textbf{Educational infrastructure:} Faculty require support structures for sharing resources, assessment techniques, and strategies to maintain academic integrity in an AI-enabled environment. The MPS community has an opportunity to combine their expertise in developing introductory courses to collaborate on the development of common curricula and teaching materials for AI education. Some materials will be broadly applicable across domains (similar to calculus), while others will be discipline-specific. This community-driven approach would accelerate adoption while maintaining flexibility for institutions and departments to adapt materials to their specific contexts and student needs. More modular degree structures that allow flexible pathways between undergraduate and graduate education warrant consideration. While models such as European 3+2 or 2+3 BA+MA/MS programs offer potential templates,  implementation of such a structure within the U.S.\ academic system would face significant institutional challenges.

\end{itemize}

\subsubsection{K-12 and Public Education}
\label{sec:public}

It is important to cultivate a strong pipeline of researchers to join the MPS community in the coming years, especially as AI continues to develop.
This presents a compelling opportunity for the AI+MPS community to engage K-12 students and the broader public through thoughtfully designed workshops and activities.
There is also opportunity to leverage AI tools for education, both of which can be extended beyond university education. 

\begin{itemize}
    \item \textbf{Educating about AI:} Introducing AI literacy at an early age can demonstrate the vital role of \gls{STEM} in powering AI-driven innovation and boosting economic competitiveness in the U.S. Outreach initiatives play a key role in this effort. For example, press releases can effectively highlight how federal funding is driving advances in science and engineering. Likewise, developing user-friendly web applications that enable the public and K-12 students to interact with AI models and MPS-generated products can demystify the technology and inspire future innovators. Science fairs, open houses, and community events featuring hands-on computational and AI activities further reinforce public trust and showcase the tangible benefits of our research investments. It is important to adopt an agile approach to AI education that mirrors the rapid evolution of the field while maintaining scientific rigor. To build an AI literacy pyramid, as introduced in \Sec{sec:knowledge-transfer}, we propose: 
\begin{itemize}
    \item \textbf{Establishing a system of stackable modules} that each represent a clear set of competencies. These modules can then be assembled into individualized learning pathways and serve as credentials for the modern workforce.
    \item \textbf{Defining a progressive set of skills on each level} of the AI pyramid, ranging from foundational concepts to advanced applications, tailored to a range of academic and professional trajectories in the sciences. 
    \item \textbf{Establishing standardized definitions} for AI literacy and proficiency, ensuring that key terminology and concepts are consistently understood across disciplines. This common curriculum would not only streamline transitions between educational levels and into industry but also validate continuous, modular learning through recognized credentials.
\end{itemize}
Such a framework could foster a more adaptable and approachable educational ecosystem and help learners---from K-12 students to seasoned researchers---develop both the deep understanding and the practical skills necessary to thrive in an AI-enabled future.

    \item \textbf{Community colleges and technical schools:} There is a strong need to broaden our focus and invest in community colleges and high school technical schools. These institutions serve as critical bridges between pre-college education and the evolving demands of the AI-driven workforce. By dedicating resources to these settings, we can ensure that high-quality AI education and training extend beyond traditional university programs. Expanding our support in this way ensures a more robust talent pipeline, preparing a range of students and professionals to contribute to---and thrive in---an AI-enabled future.

    \item \textbf{Using AI tools in education:} As with undergraduate education, AI tools have the potential to enhance K-12 learning itself. AI-powered educational technologies can provide immediate feedback and adaptive content that addresses individual student needs and learning styles. These tools can help teachers identify knowledge gaps, suggest targeted interventions, and make abstract concepts more accessible through interactive visualizations and simulations, particularly in \gls{STEM} subjects where many students struggle.
\end{itemize}

\subsection{Empower AI Innovation}
\label{sec:empower-ai}

\glsresetall

While the use of AI in MPS research is rapidly growing, there are still significant barriers to its broader adoption, both logistically and scientifically.
By removing these barriers, we can help \textbf{empower more researchers to integrate AI into their work and contribute their domain expertise to innovation.}
Below, we discuss ideas related to lowering the cost of computing for AI, preserving standards for scientific integrity in the AI era, and engaging with the public about AI+Science.

\subsubsection{Cost-Efficient Computing}
\label{sec:cost-efficient}

As emphasized in \Sec{sec:infrastructure}, AI requires robust infrastructures, but the energy costs of the compute and data storage resources required to support modern AI-enabled science and technology are non-trivial.
As an extreme example, development of GPT-4 is estimated to have  
cost \$100M to train \cite{Buchholz2024}.
Of course, most applications are not nearly so resource demanding, but as a community \textbf{we must reckon with the cost-benefit tradeoff in our endeavors}. It is important to facilitate access to cutting-edge tools, such as free computing resources for all students and postdocs, but one must also consider the high power cost associated with large-scale computation.  How can we capture and price these externalities into AI-enabled science since these efforts have a largely invisible impact on energy? Some potential opportunities to address this include: 

\begin{itemize}
    \item \textbf{Leveraging scientific innovation to develop more efficient computing solutions}, such as by developing algorithms to improve the data and computational efficiency of AI (\Sec{sec:data-efficient}).  This will help enable broader access to AI tools and battle the immense energy costs. 
    \item \textbf{Increasing the adoption of smaller, more structured models} could help address challenges associated with the high resource requirements of the currently used models, as can the judicious use of fine-tuning strategies and computation-efficient adapters (e.g., \gls{LoRA}). 
    \item \textbf{Utilizing model distillation as a technique,} whereby much smaller models are trained by leveraging responses from a larger one. This has the advantage of less expensive training, inference, and fine tuning of the distilled model while maintaining comparable performance. \item \textbf{Exploring emerging computing paradigms such as neuromorphic, in-memory, photonic, and quantum computing} for cost-efficient computing for AI. Just as \glspl{GPU} are particularly well suited to training \glspl{DNN}, alternative computing paradigms might inspire new AI models.
    \item \textbf{Prioritizing well-designed code and professional approaches to software development}, which can have a large impact on cost efficient computing. This requires either training in software development or for professional software engineers to develop core libraries for the MPS community.  There may also be opportunities for AI co-pilots to assist in code maintenance and development (\Sec{sec:ai-co-pilot}).

\end{itemize}

As discussed in \Sec{sec:ai-innovations}, developing more efficient algorithms is an active research area in the MPS domains. These MPS efforts can play an important role in addressing the above needs and contribute to energy security and grid modernization in the U.S.\ by reducing the computational energy costs of AI systems.

\subsubsection{Scientific Integrity}
\label{sec:scientific-integrity}

Researchers’ identities are shaped by their approach to questions, methodological choices, and interpretive frameworks. When AI begins to influence or execute these aspects, it is necessary to reconsider what defines our distinct contributions to scientific knowledge and what gives meaning to scientific work. This leads to many questions, including: 
    \begin{itemize}
        \item How do we establish domain assurance, validate results, and maintain scientific rigor when using AI for scientific discovery? 
        \item What does ``rigor'' look like in the AI era? Who (or what) makes this judgment? 
        \item What constitutes the essence of scientific discovery when AI systems can handle increasing portions of the research process?
        \item What does it mean to ``do science'' collaboratively with AI? How do we preserve the uniqueness of human inquiry and creativity while leveraging AI capabilities?
        \item At what point does extensive AI assistance transform the nature of authorship and scientific contribution?
        \item How can we responsibly integrate generative AI into teaching, doing, and reporting science rather than avoiding it?
        \item What will we do when more people tend to use \glspl{LLM}/reasoning agents rather than papers as a main source of knowledge? How will this affect how we choose problems and evaluate the work of others? 
        \item What are the safety risks of the application of AI in MPS, especially when they cross over to the physical world (e.g.\ in autonomous labs) either by malicious use or accidental harm?
    \end{itemize}

Given these important ethical and conceptual questions, \textbf{the philosophy of science should be included in discussions of scientific integrity} as much as possible. AI+MPS researchers could collaborate with social scientists/humanities to help build trust and consider the impact beyond science (see \Sec{sec:beyond-mps}). 
Important considerations related to scientific integrity that the MPS community should take into account include: 
\begin{itemize}
    \item \textbf{Academic considerations:} Given the fast pace of AI, academic integrity is harder to enforce for scientific research using AI; for example, it is challenging to identify enough experts to provide quality scientific reviews (e.g., for manuscripts or grant applications) in this space. This may require updated guidelines or workflows for reviewing research that considers the review cultures in both AI and MPS communities. The scientific community should also remain vigilant about the potential for AI to disproportionately attract researchers and students toward subfields more readily compatible with AI methodologies, potentially impeding progress in other important areas within MPS disciplines that are less amenable to current AI approaches.
    \item \textbf{Addressing AI skepticism:} Distrust of AI seems to take two forms: 
\begin{enumerate}
    \item \textit{Skepticism about AI's usefulness in making discoveries.} There needs to be care from the community and federal agencies to ensure that ``good science'' is being followed in order to mitigate potential falls off the ``hype curve.'' 
    \item \textit{Distrust of AI results given the history of hallucinations and lack of explainability.} A hidden cost of AI is that, when used well, it creates substantial extra work for the researchers to convince themselves and others that the algorithm will provide accurate results for the particular application at hand. 
\end{enumerate}
    To enhance adoption of AI in domain sciences, success stories need to be highlighted through real-world case studies. In doing so, it is important not to exaggerate the performance of AI methods. It also helps to assuage distrust when researchers provide the details needed to replicate their results. Another way to combat distrust could be to implement data challenges that demonstrate the reliability of AI for scientific applications. An undertrained scientific community can produce superficial work labeled as AI research, which further reinforces the skepticism among domain specialists. Increasing AI literacy in domain scientists would reduce misinformation, improve AI/statistical practices, and underscore the value of AI/ML backgrounds within the domains. 
    \item \textbf{AI interpretability:} While AI models may produce accurate predictions, they often function as ``black boxes'' that provide limited scientific insight into the underlying physical processes. The MPS community increasingly demands explainable AI methods that not only make predictions but also reveal the physical relationships and features driving those predictions. This opacity can undermine trust in results and limit the perceived scientific value of AI applications. The need for interpretability across MPS is domain and problem dependent, with more practical research areas somewhat willing to sacrifice model interpretability if one can verify that the required tasks were performed correctly. That said, developing stronger \gls{UQ} capabilities that are interpretable by both experts and non-experts alike could increase trust in AI-enabled simulations (\Sec{sec:uncertainty_quantification}). The challenge of balancing model complexity with interpretability remains an important research opportunity for many scientific applications (see \Sec{sec:robust-and-reproducible-AI}). 

    \item \textbf{Data guidelines:} Foundation models are a key AI technique for science (see \Sec{sec:foundation_models}), but the extremely high cost of training them means that academic researchers lacking the resources to do so from scratch may instead fine tune models developed by industrial researchers. This can raise issues associated with transparency and reproducibility if models have been trained on non-public data sets that are not available to the users. More generally, advocating for the disclosure (if not availability) of sources of training data, particularly for large foundation models is crucial. There is an opportunity to define and promote community ethical standards (likely field-specific) for the accessibility and use of data in AI training (also discussed in \Sec{sec:data-management}), which will be critical to encourage best practices and promote trust in AI applications and results. 

    \item \textbf{Data provenance:} The notion of data provenance as necessary for robust AI is also important. Techniques like \gls{RAG} can aid in ensuring \glspl{LLM} are leveraging known data. In addition to techniques like \gls{RAG}, it is also important to consider quantitative evaluation frameworks such as \gls{RAGA}.
    \Glspl{RAGA} provide structured metrics for assessing key aspects of \gls{RAG}-based systems, including the relevance of retrieved contexts and the faithfulness of generated responses, enabling more efficient and objective evaluations. 

    \item \textbf{Access and ownership:} The resources to use AI also need to be broadly accessible to avoid the ability to use AI going to only a privileged few. Continued support for the development and dissemination of resources through the NAIRR Pilot program will be important to continue (see \Sec{sec:computing-resources}). AI tools need to be made accessible across the board, with a pathway for software coming out of AI research to be documented and maintained so that it is usable and able to be adapted to the needs of domain scientists. On the other hand, it is important to consider ownership of data used to train the models. There have been many cases where content was used for training purposes without agreement by the owner of the data. Establishing guidelines for the use of data and models  that are not clearly open source or open weight should be a priority for the MPS community (see also \Sec{sec:data-management}).

\end{itemize}

\subsubsection{Public Engagement on AI+Science}

\textbf{Proactive public engagement can help foster public trust in scientific research}, and this is especially relevant for AI+Science.
 The societal conversation around AI involves topics like how to align AI with human values and goals, how to ensure that AI is safe and trustworthy, and how to mitigate potential ethical issues in using AI.
 The MPS community must be cognizant of ways to connect our work to these concerns, especially in cases where scientific principles can be used to improve our understanding of AI systems (see \Sec{sec:understand-ai-behavior}).

At the same time, AI could be considered the most positive and impactful development in a lifetime, so the MPS community as a whole has the opportunity play a substantial role in communicating its potential.
Approaches to this could include the following. 
\begin{itemize}
    \item \textbf{Establish programs to increase AI literacy} in the community, including beyond academia and industry.
    This includes communicating general knowledge about how AI works (see \Sec{sec:public}) but also highlighting specific MPS advances enabled by AI (see \Sec{sec:AIadvanceMPS}).
    
    \item \textbf{Emphasize the connection between AI and national interests/economic competitiveness}.
    One could imagine presenting the AI revolution as being similar to the public-participating ``race for space'' that helped drive science and engineering innovations in the 1950s/60s.
    \item \textbf{Address ethical considerations important to the general public}.  This includes topics like data privacy (especially in medical research) and efficient use of natural resources when using AI tools (as in \Sec{sec:cost-efficient}). 
\end{itemize}
These public engagement efforts are also an opportunity to emphasize the importance of foundational research to advancing both AI and Science.

\section{Domain-Specific Opportunities}
\label{sec:domains}

\glsresetall

Innovation comes not from the mere use of AI tools, but from the scientific work that AI enables, which could not have been done before, or at least not done as easily.
As discussed in \Sec{sec:facilitating}, a key benefit of facilitating connections and collaborations between AI+MPS researchers is the opportunity for cross-pollination and transferable learning across disciplines.
At the same time, it is important to \textbf{recognize and engage the specific methods, culture, and drivers for each MPS domain}, which affects the potential integration and impact of AI in the research process.
In this section, we summarize how AI is advancing (and is advanced by) each of the MPS domains and highlight future research opportunities in each area.
While the discussion below takes a primarily optimistic perspective, we anticipate that substantial original research will be needed to overcome the challenges of integrating AI into each MPS domain.

\subsection{Astronomical Sciences (AST)}
\label{sec:ast}

\glsresetall

\subsubsection{AI in the Context of AST}

Astronomical Sciences has emerged as a domain where AI plays an increasingly important role. With investments from the \gls{NSF}, \gls{DOE}, and \gls{NASA} amounting to billions of dollars in recent years, the field stands at a critical juncture for maximizing scientific returns from major observational initiatives such as the Vera C. Rubin Observatory and the Nancy Grace Roman Space Telescope. These facilities will generate unprecedented data volumes---petabytes to exabytes annually---that far exceed traditional analysis capacities. \textbf{These heterogeneous, massive datasets present substantial computational and analytical challenges, making them ideal candidates for AI applications.} AI methodologies have advanced data classification, identification, and discovery while offering cost savings by processing astronomical data at unprecedented scales. Generative models have improved the identification of interesting outliers, a critical task in astronomy's search for rare phenomena. In observatory operations, AI has made inroads in astro-instrumentation through workflow optimization using \gls{RL} techniques. Beyond these operational improvements, AI has enabled new approaches including \gls{SBI} for parameter estimation and neural emulators that accelerate computationally expensive simulations.

\subsubsection{How AI is Advancing AST}

Across astronomical subfields, AI adoption has created valuable research opportunities with varying impact:

\begin{itemize}    
    \item \textbf{Cosmology} research has developed deep learning approaches that have advanced parameter inference from large-scale structure and cosmic microwave background data, with field-level inference methods going beyond traditional summary statistics to utilize more complete information in complex stochastic fields. Neural networks have been used to advance and speed up numerical solvers of expensive N-body simulations, accelerating computational cosmology.  Specialized architectures like \glspl{GNN} and topology-based approaches have been designed to handle the unique characteristics of cosmological data.

    \item \textbf{Time-domain astronomy} has benefited from neural networks that support real-time classification of transient events and early detection of phenomena like supernovae, allowing for timely follow-up observations. Deep learning methods effectively handle the irregular sampling and heterogeneous noise characteristic of astronomical time series. Unsupervised techniques and clustering algorithms enable identification of anomalies and discovery of new classes of transient phenomena without theoretical biases, streamlining observational follow up. Moving object detection algorithms distinguish asteroid and comet motion from artifacts in image sequences, expanding our census of solar system objects.

    \item \textbf{Exoplanet research} has progressed using machine learning techniques that better distinguish planetary signals from stellar activity and systematic noise in transit photometry and microlensing data. Techniques from AI enable efficient exploration of highly multimodal posterior distributions such as those encountered in characterizing microlensing events or sparsely sampled spectroscopic timeseries containing an unknown number of planetary signals. Modern deep learning frameworks integrate multiple observation modes for comprehensive parameter estimation, characterizing planetary systems from orbital parameters to atmospheric properties. \Gls{RL} enhances adaptive optics performance by learning optimal control strategies for atmospheric correction, improving the sensitivity of ground-based exoplanet detection.

    \item \textbf{Stellar astrophysics} has applied AI to automate spectral analysis across surveys like the Sloan Digital Sky Survey, extracting physical parameters from low-resolution and low signal-to-noise spectra efficiently. Transfer learning bridges the gap between synthetic models and real observations, with networks pre-trained on theoretical spectra adapting to analyze observed data. AI has advanced the study of eclipsing binaries through automated pipelines that map light curves directly to physical parameters, enabling rapid processing of thousands of light curves and helping eliminate false positives in searches for stellar-mass black holes. Neural networks extract rotation periods from photometric time series significantly faster than traditional methods. In asteroseismology, neural networks analyze stellar oscillation data to determine stellar interior properties and evolutionary phases, capturing complex multi-scale phenomena. \Glspl{PINN} have been applied to accelerate stellar atmosphere modeling.

    \item \textbf{Galaxy evolution} studies employ deep learning for both morphological representation extraction and for simulation emulation, reducing computational costs while maintaining physical accuracy. Generative models provide new insights into galaxy morphologies by learning their underlying distributions and enabling conditional generation based on observable parameters. These models help understand morphological evolution through latent space representations and support of synthetic populations for training data augmentation. \Glspl{GNN} model how galaxy properties depend on their cosmic environment, capturing how proximity to clusters, filaments, and neighboring galaxies influences star formation and morphology. Deep learning enables automated identification of rare populations such as low surface brightness dwarf galaxies across large survey volumes.

    \item \textbf{Solar and heliophysics} utilize AI for forecasting space weather events through ensemble methods that analyze solar magnetic field data, with applications to satellite operations and Earth-based technologies. Neural networks now predict flare magnitudes and timing, enabling graduated operational responses. Multi-modal architectures combine magnetograms with chromospheric and coronal observations to capture full magnetic connectivity determining eruption likelihood. Neural networks excel at automatically identifying and classifying dynamic features like sunspot groups and analyzing photospheric vector magnetic fields. Developments in neural fields have enabled improved reconstruction of solar surface features, providing a more complete picture of solar dynamics.

    \item \textbf{Gravitational wave astronomy} has integrated deep learning for signal processing, noise mitigation, and parameter estimation, supporting multi-messenger follow-up observations. Neural networks effectively model complex physics like spin precession and higher-order multipoles in gravitational waveforms. Surrogate models trained on numerical relativity simulations accelerate waveform generation, enabling extensive parameter space exploration. Deep learning approaches offer comparable sensitivity to matched filtering but with dramatically reduced computational costs, enabling rapid detection and characterization. Early warning networks analyze pre-merger signals to predict coalescence times for gravitational signals, enabling electromagnetic facilities to capture merger and post-merger evolution.

    \item \textbf{Astronomical instrumentation} has integrated AI to optimize operations of costly ground-based and space-based telescopes, significantly reducing operational expenses. Deep learning approaches demonstrate substantial efficacy in image reconstruction and turbulence prediction for adaptive optics systems, critical for high-resolution imaging applications. Neural networks enhance telescope target acquisition, tracking precision, and alignment calibration in multi-mirror telescopes and interferometric arrays. AI transforms telescope scheduling—essentially a multi-dimensional traveling salesman problem—by optimizing paths through thousands of targets while adapting to weather conditions and time-critical follow-up requirements. Deep learning excels at data contamination removal, including radio frequency interference identification and cosmic ray rejection, preserving astronomical signals while eliminating artifacts.

\end{itemize}

These advances illustrate how AI is transforming astronomical research across the entire scientific pipeline, from instrumentation and observation planning through data reduction and analysis to theoretical model building. Progress at the AI+AST intersection includes the following themes.

\begin{itemize}
    \item \textbf{Parameter inference through \gls{SBI}} has transformed how astronomers extract maximum information by solving the ``inverse problem.'' These approaches make previously intractable analyses more feasible and improve computational efficiency via fast emulators for generative models and representational learning. By learning directly from simulations rather than assuming analytical likelihoods, \gls{SBI} can extract information from complex observables that traditional summary statistics would miss. These methods show promise for tighter cosmological constraints from weak lensing fields and more complete characterization of exoplanet architectures. These models synthesize and analyze complex signals and derive new physical insights from summarization of simulations.
    
    \item \textbf{Data processing scalability} has been dramatically enhanced, addressing the challenge of working with large, complex datasets. Techniques like cosmic ray rejection, efficient sub-sampling, and amortized neural posteriors have made AI essential for processing astronomical data at unprecedented scales. Neural networks enable real-time processing of millions of transient alerts nightly, automated feature detection across imaging and spectroscopic surveys, and rapid data quality assessment. Beyond traditional automation, \gls{RL} transforms telescope operations by learning optimal scheduling policies and discovering non-obvious observational strategies. For adaptive optics, neural networks learn temporal patterns in atmospheric turbulence, enabling predictive correction. This combination of scalable processing and intelligent control systems represents a fundamental shift in how astronomical facilities operate, with AI both managing unprecedented data rates and optimizing scientific return through learned operational policies.

    \item \textbf{Anomaly detection and novel discovery} capabilities address the ``needle in the haystack'' problem through unsupervised and generative models. Unlike supervised approaches that find outliers from known categories, these methods discover ``unknown unknowns''—entirely new phenomena without predefined labels. Generative models learn the underlying distribution of astronomical observations, identifying low-probability objects as candidates for detailed investigation. These data-efficient approaches have facilitated the discovery of unusual transient events (electromagnetic or gravitational wave), rare stellar types, and peculiar galaxy morphologies that do not fit established classification schemes. Multi-messenger and multi-wavelength data integration combines heterogeneous information types (imaging, spectroscopy, time series, polarimetry) to construct more complete physical models, with foundation models providing unified representations across different observational modes.

    \item \textbf{Simulation acceleration} through \gls{GPU}-powered numerical solvers and emulator models has dramatically reduced computational requirements for astrophysical modeling. A persistent challenge in astronomy involves multi-scale simulations that must simultaneously capture both large-scale cosmic structures and small-scale physical processes. Neural network emulators have been developed to replace computationally expensive components such as N-body gravitational simulations, radiative cooling calculations, and sub-resolution physics prescriptions that model processes occurring below the simulation's spatial resolution. Neural \glspl{ODE} learn effective dynamics from high-resolution simulations, enabling efficient coarse-grained evolution while preserving essential physics. Physics-informed approaches ensure these accelerated models maintain conservation laws and respect fundamental constraints, enabling more extensive parameter space exploration and statistical analysis of theoretical models.
\end{itemize}

\subsubsection{How AST is Advancing AI}

Astronomy offers AI researchers valuable opportunities through its large and complex open datasets, physics-based models, broad public interest, and well-defined literature. Additionally, extrasolar astronomy provides a uniquely safe, low-risk ecosystem for experimentation due to its open-source, non-proprietary data culture and because most errors or failures have minimal direct consequences for human life or well-being. This combination allows researchers to freely test novel AI approaches without ethical concerns or commercial constraints. Further, astronomy presents case studies where low-dimensional underlying physical models may be discoverable through AI techniques, potentially leading to fundamental advances in how machine learning systems can extract scientific knowledge from complex observations. Specific contributions astronomy is making to advance AI include:

\begin{itemize}
    \item \textbf{Rich, open datasets} with physical relationships spanning large dynamic ranges. Astronomy provides access to complex, non-proprietary data that contains interesting domain shifts between training and test sets. These datasets enable AI researchers to study generalization under realistic constraints, addressing one of machine learning's fundamental challenges.

        \item \textbf{Well-curated scientific literature} with exceptional accessibility spanning decades. Nearly all astronomy papers are available through arXiv and cataloged by NASA's \gls{ADS}, with journals archiving data tables and figures directly. Complementary systems like the \gls{CDS} provide standardized datasets from papers predating the modern open data movement. Critically, observatories provide public access to both published and unpublished observations, creating opportunities for AI to conduct genuine research without acquiring new data. This comprehensive ecosystem—combining literature, data, and analysis codes—provides an ideal proving ground for developing AI systems capable of autonomous scientific discovery.

        \item \textbf{Physics-informed machine learning frameworks} that integrate fundamental principles into model architectures. Astronomical research is deeply grounded in physics, providing an ideal testing ground for AI approaches that incorporate physical laws. Astronomy drives innovation in encoding symmetries, conservation laws, and other physical constraints directly into neural networks. This includes developing flexible approaches that balance physical constraints with discovery potential---models must be physically sound yet adaptable enough to discover new phenomena, such as through automatic symmetry detection.

        \item \textbf{Multi-modal data integration}, which presents a fascinating AI challenge, especially with non-proprietary data across different observation types, wavelengths, and methods. Astronomical research necessitates combining imaging, spectroscopy, time series, and theoretical models, pushing AI to develop more sophisticated techniques for integrating heterogeneous information sources. This domain features unique modalities not typically handled in mainstream AI research, particularly spectroscopic data with its high dimensionality and complex noise characteristics, presenting distinctive challenges compared to standard image, text, or audio processing tasks.
    
        \item \textbf{Experimental design optimization} for new instruments, target selection, and data discovery workflows. Telescope scheduling exemplifies complex constrained optimization—balancing weather conditions, target visibility, instrument changes, and scientific priorities across thousands of objects. The exploration-exploitation trade-off inherent in astronomical observations creates novel challenges for \gls{RL}. These applications push \gls{RL} beyond typical domains by requiring long-term planning horizons, handling partial observability of atmospheric conditions, and optimizing for diverse scientific objectives simultaneously, advancing decision-making AI under realistic operational constraints.

        \item \textbf{Multi-agent AI systems} that combine specialized tools for different astronomical tasks. Such systems face a particularly interesting challenge in astronomy, where knowledge is highly intersectional and requires both intuitive reasoning and mathematical derivation across a vast domain. Astronomy uniquely demands physical reasoning that combines approximations, limiting cases, and order-of-magnitude estimates with mathematical rigor, presenting a distinct frontier beyond pure mathematical theorem proving. These systems show promise for creating more comprehensive research workflows that integrate literature knowledge, observational planning, data reduction, and theoretical interpretation, serving as testbeds for advanced agent architectures.
\end{itemize}

The astronomy community is uniquely positioned to contribute to AI advancement through its interdisciplinary institutes and collaborative research traditions. The field has undergone a community-wide effort to build connections between AI and domain scientists, creating momentum that should be maintained and expanded. Incentivizing interdisciplinary collaborations, particularly those connecting larger institutes with smaller research groups and individual interdisciplinary researchers, will further accelerate progress at this intersection.

\subsubsection{Opportunities in AI+AST}
\label{sec:ast-opportunities}

In addition to the existing advances, other opportunities remain. Opportunities to leverage and develop key AI techniques identified in \Sec{sec:techniques} include:

\begin{itemize}
    \item \textbf{\glsreset{SBI}\Gls{SBI}:} While \gls{SBI} has shown promise for parameter estimation, opportunities remain to extend these methods to more complex astronomical scenarios. Future developments could enable inference from incomplete or heterogeneous multi-messenger observations and handle systematic uncertainties in simulations themselves. Real-time \gls{SBI} for transient follow-up decisions presents a key frontier for the Rubin Observatory, where millions of alerts nightly will require rapid probabilistic classification to prioritize follow-up resources. These methods will be equally critical for the Roman Space Telescope, where extracting cosmological information from weak lensing fields beyond traditional statistics could significantly tighten constraints on cosmological parameters. Particularly valuable extensions of current \gls{SBI} would be methods that reveal which observational features drive constraints, enabling understanding of what aspects of data provide the most information. Rigorous validation on benchmark datasets will be essential to establish trust in these powerful but complex methods.
    
    \item \textbf{Multi-scale simulations:} Current neural emulators primarily target specific physical processes, but astronomy needs methods that seamlessly bridge the many orders of magnitude in spatial scale—where small-scale physics (e.g., stellar feedback) determines large-scale evolution (e.g., circumgalactic medium enrichment). For simulation acceleration, while neural operators show promise, current approaches struggle with non-periodic boundary conditions typical of astronomical systems. Critical applications include modeling the circumgalactic medium where parsec-scale supernova bubbles and stellar winds drive turbulence and metal enrichment in the galactic gaseous halos that regulate galaxy growth. Similarly, \gls{GRMHD} simulations of black hole accretion require bridging from horizon scales to jet propagation across millions of light years. Opportunities include developing hierarchical emulators that preserve causal connections across scale separations and robust probabilistic approaches that capture the stochastic nature of sub-grid physics. \Glspl{PINN} should be more universally incorporated across these applications to ensure conservation laws and physical consistency are maintained throughout the modeling hierarchy.

    \item \textbf{\glsreset{UQ}\Gls{UQ}:} While deep learning excels at point predictions, astronomy requires rigorous error propagation for scientific inference. Opportunities include developing methods that distinguish aleatoric uncertainty (inherent noise) from epistemic uncertainty (model limitations), particularly crucial when extrapolating beyond training domains. Networks that can identify and flag when they encounter out-of-distribution data—such as unusual transients or extreme astrophysical conditions—would prevent overconfident predictions. Despite promise, Bayesian neural networks remain computationally challenging for astronomical data volumes, and existing approximations often produce poorly calibrated uncertainties. Critical needs include computationally tractable methods that provide well-calibrated uncertainties throughout analysis pipelines.
    
    \item \textbf{Foundation models:} The development of large-scale pre-trained models for astronomy could transform how we analyze diverse observations. True foundation models would move beyond current domain-specific architectures to achieve generality across astronomical applications. A critical frontier is multi-messenger astronomy, where foundation models could unify analysis across electromagnetic observations for maximizing science from Rubin Observatory's rapid transient alerts that trigger follow-up across these diverse channels. While existing explorations with smaller models have shown promise, robust few-shot learning—adapting to new instruments or phenomena with minimal examples—and zero-shot transfer across wavelength regimes remain limited in scope and scale. Opportunities include multi-modal pre-training that jointly learns from imaging, spectroscopy, and time-domain data in shared representation spaces, and models that can handle extreme dynamic ranges in observations.
    
    \item \textbf{AI for experimental control:} \Gls{RL} applications in astronomy remain largely experimental, with most implementations still in proof-of-concept phases despite showing significant promise. The opportunity exists to move from demonstrations to operational deployment across observatory networks. Key applications include adaptive optics systems where \gls{RL} can learn predictive control strategies for atmospheric turbulence correction—essential for ground-based telescopes to achieve space-like resolution and the extreme contrast needed for direct exoplanet imaging—and intelligent telescope scheduling that optimizes target selection across thousands of objects while adapting to weather, seeing conditions, and time-critical alerts. Future systems could discover entirely new observational strategies—innovative coordination patterns between facilities, predictive responses to atmospheric changes, and multi-year survey optimizations that no human would design. Key challenges include bridging the gap from simulated environments to real telescope operations with all their complexities and creating safe exploration methods that learn without risking valuable observing time.
    
    \item \textbf{Data-efficient methods:} The key to learning from limited astronomical data lies in encoding appropriate physical priors through architectural design, training strategies, and data augmentation approaches. Critical applications arise when training data requires prohibitively expensive simulations—3D stellar atmosphere models requiring months of supercomputer time per star, or \gls{GRMHD} simulations of black hole accretion. For these cases, networks must extract maximum information from perhaps dozens of examples rather than millions. Opportunities include developing networks with built-in symmetries—rotation equivariance for galaxy morphology, scale invariance for self-similar structures, and Lorentz invariance for relativistic phenomena—that learn universal patterns rather than memorizing orientations. Physics-informed architectures that encode conservation laws and governing equations could enable robust extrapolation from sparse training sets, particularly for extreme astrophysical conditions rarely observed.

\end{itemize}
Additional high-priority opportunities for AST include: 
\begin{itemize}

    \item \textbf{Validation and benchmarking:} The accuracy of AI-based inference is fundamentally limited by simulation fidelity. As we extract more information from data, we expose unknown systematics in our models. This is particularly critical in cosmology, where \gls{SBI} promises to extract information beyond traditional two-point statistics, but only if we can trust that neural networks learn cosmological signals rather than simulation artifacts. Establishing trust requires comprehensive benchmark datasets that test AI methods against traditional approaches using both simulated and real observational data. The community needs standardized evaluation protocols specifically designed to uncover whether AI methods are learning from baryonic physics approximations, numerical artifacts, or true cosmological signals—a challenge that intensifies as we push toward percent-level precision for cosmological constraints from Roman observations.
    
    \item \textbf{Interpretable AI for science:} The black-box nature of many AI methods challenges astronomical applications where understanding systematic biases is crucial for distinguishing artifacts from astrophysics. Opportunities include developing sparse representations that isolate physically meaningful features, robust models that maintain performance when key assumptions are violated, and attention mechanisms that reveal which input features drive predictions. Causal inference frameworks and graph-based approaches could reveal the underlying physical dependencies between observables, moving beyond correlation to understand what drives astrophysical phenomena. These approaches would enable scientists to discover new physical processes and relationships that may be hidden in complex observational data, potentially revealing unknown mechanisms governing stellar evolution, galaxy formation, or cosmological structure growth.
    
    \item \textbf{AI-driven scientific discovery:} While AI for Astronomy has matured with many statistical and operational methods gradually permeating all subfields, there remains a notable gap in truly surprising discoveries directly attributable to AI. The field has yet to see breakthrough insights or novel theoretical models proposed by AI systems themselves, rather than through human-guided analysis of AI results. A critical near-term application lies in intelligent triage of Rubin Observatory's alert stream—moving beyond simple classification to systems that identify scientifically interesting outliers, synthesize contextual information from archives and literature, and autonomously flag candidates worthy of immediate follow-up based on their potential for discovery. This frontier may open up with advances in multi-agent systems and reasoning capabilities that go beyond routine workflow optimization and advanced statistical inference. The next phase of AI in astronomy may progress from tools that assist human-led science to systems that actively participate in the scientific process through hypothesis generation and experimental design.
\end{itemize}

\subsection{Chemistry (CHE)}
\label{sec:che}

\glsresetall

\subsubsection{AI in the Context of CHE}
Chemistry and AI have long been working together, creating an exchange that is reshaping both fields. On the one hand, AI is helping chemists predict reactions, design molecules, and analyze complex datasets with speed and precision---accelerating discovery across chemical synthesis, drug development, materials science, and catalysis. On the other hand, chemistry is pushing the frontiers of AI. Theoretical chemistry has inspired some of today’s most powerful generative models, including diffusion models which were directly influenced by non-equilibrium thermodynamics. In fact, chemistry has provided some of AI’s most celebrated successes; notably, AlphaFold earned the 2024 Nobel Prize in Chemistry \cite{NobelPrize2024Chemistry} for its breakthrough in protein structure prediction and protein design. These examples highlight a growing, bidirectional relationship:\textbf{ AI is transforming how we do chemistry, and chemistry continues to shape and challenge the future of AI itself.} 

\subsubsection{How AI is Advancing CHE}

AI research is revolutionizing the field of chemistry, especially (so far) in areas like theoretical and computational chemistry, catalysis, chemical synthesis, biomolecular structure prediction, and protein design. Naturally, there is some overlap between \acrshort{CHE} and \acrshort{DMR} (\Sec{sec:dmr}) research on some of the topics described below, especially related to molecular and polymer discovery and synthesis.

\begin{itemize}
    \item \textbf{Computational chemistry} increasingly uses AI tools to predict molecular behavior and accelerate simulations, particularly for complex systems such as protein dynamics or excited-state reactions in a complex medium \cite{Haghighatlari2019}. These often involve developing surrogate models to replace expensive simulation methods based on quantum mechanics. Beyond speed-ups, AI is also aiding in automated design of coarse-grained models, identifying low-dimensional collective variables that capture essential system dynamics, and enhancing sampling of rare events that are otherwise inaccessible to conventional simulations. These different sub-disciplines of computational chemistry are now rapidly being transformed with the thoughtful use of AI, often deeply integrated with statistical mechanics.

    \item \textbf{Catalysis}  leverages AI to pinpoint efficient pathways, optimize materials for sustainable reactions, and identify optimal catalyst features for asymmetric synthesis. Successfully applying generative AI to asymmetric catalysis could have a major impact on chemical synthesis and drug discovery. Another key opportunity lies in combining AI with traditional rare-event sampling methods, which are essential for accurately capturing slow but critical molecular changes in processes like protein conformations or catalysis.

    \item \textbf{Materials synthesis and discovery} is becoming faster and more accessible with AI-enabled automation, making it possible to plan and execute intricate synthetic routes that were previously too complex or time-intensive. However, one significant limitation is the inability of AI to generalize well on data beyond its training set. Better generalization could open doors to more reliable predictions in protein folding, reaction mechanisms, molecular simulations, asymmetric catalyst identification, and molecular function. Automated synthesis, while advancing, still requires more sophisticated decision-making algorithms for multi-step reactions and dealing with changes in substrate characteristics, which could enable more reliable and generalizable optimizations in chemical synthesis and drug and material discovery.

    \item \textbf{Interatomic potentials and atomistic simulation} studies have been transformed by machine learning, enabling first-principles accuracy with far greater computational efficiency. While traditional first-principles approaches like \gls{DFT} are limited to small systems (hundreds of atoms) and short timescales (picoseconds), \glspl{MLIP} make it possible to simulate tens of thousands of atoms over nanoseconds or longer \cite{Anstine2023}. These advances provide efficient and accurate ways to predict reaction pathways and transition states, rapidly screen molecular libraries, and design novel hypothetical materials with targeted properties for experimental synthesis.
    \item \textbf{Retrosynthesis analysis and automatic synthesis} can leverage AI to help plan retrosynthetic routes. Some tools use deep learning to find effective ways to make molecules without using fixed templates. These tools are also useful for new or strange molecules. When used in combination with robotics, these labs can test and improve the plans automatically. This moves toward fully automatic chemistry labs.

\end{itemize}    
Beyond specific domain applications, current progress and benefits of AI for chemistry in general are as follows.
\begin{itemize}
    \item \textbf{Simulation and prediction:} The use of AI tools in chemistry have enabled better simulations, allowed for automatic prediction and discovery, and accelerated subdomains like quantum chemistry. AI tools are being used to predict existing functions or design new functions for proteins, small molecules, catalysts, and materials. The 2024 Chemistry Nobel Prize winning AlphaFold has been a stand-out success in this domain. The use of AI in these areas has helped accelerate molecule and material simulations, and high-throughput data analysis, enabling rapid, low-cost predictions of properties, reactivity, and selectivity that were previously intractable.  
    \item \textbf{Discovery and engineering:} AI-driven models allow us to predict how specific amino acid changes will affect protein activity, stability, or binding properties, even in highly complex systems. These tools have not only accelerated the pace of discovery but have also expanded the range of possibilities, allowing us to tackle challenges in therapeutics, bioenergy, enzymatic catalysis, and environmental sustainability that were out of reach just a few years ago. AI has truly revolutionized the way we approach protein engineering, turning what once felt like a guessing game into a science-driven, rational design process.
    \item \textbf{Accuracy:} By leveraging architectures such as transformers and embeddings, originally developed for human language, we can predict protein structure, function, and evolutionary relationships with unprecedented accuracy. The experimental accuracy of condensed phase simulations with ML potentials has also improved.
    \item \textbf{Optimization:} AI has also played a role in the optimization of catalysts for synthetic procedures, synthetic route planning, and electronic structure calculations, as well as property, reactivity, and selectivity predictions.
\end{itemize}

\subsubsection{How CHE is Advancing AI}

Chemistry is also pushing the frontiers of AI, especially in the field of diffusion models. Below are examples of CHE research's impact on AI innovation.

\begin{itemize}
    \item \textbf{Diffusion models:} Theoretical chemistry has inspired the development of diffusion models, now a cornerstone of modern generative AI. These models borrow directly from stochastic processes and physical intuitions used in simulating molecular motion.
    
    \item \textbf{Sparse ML:} Chemical modeling challenges, such as conformational sampling and transition state discovery, as well as labor intensive experimental road blocks to generating large data sets, have motivated the design of new algorithms for learning from sparse and high-dimensional data, driving innovation in machine learning methodologies.
    
    \item \textbf{AI+Science integration:} The success of AlphaFold in protein structure prediction, awarded the 2024 Nobel Prize in Chemistry, highlighted chemistry as a proving ground for AI systems that integrate carefully curated chemical/biological datasets with large-scale neural architectures.
    
    \item \textbf{Physics-informed architectures:} Quantum chemistry and electronic structure calculations have created demand for AI models that can learn complex, many-body interactions with high fidelity, pushing the development of \glspl{PINN} and equivariant architectures. Chemistry's emphasis on interpretability and physical constraints has also led to a growing interest in building AI systems that are more explainable, robust, and generalizable—especially for safety-critical applications.
    
    \item \textbf{Structured data:} Datasets from chemical simulations and experiments are among the most structured and information-rich in science, providing ideal testbeds for developing, benchmarking, and stress-testing new AI models and training paradigms. In computational molecular sciences, many AI methods are built on statistical physics principles, enabling models to make extrapolative predictions beyond the training data. These physics-grounded approaches offer a controlled environment for experimentation, but it remains an open question how well such capabilities transfer to real-world experimental settings beyond computer simulations.

    \item \textbf{Adaptive AI for faster experiments and simulations:} AI can help run lab experiments and computer simulations in real time. These agents can change their plans based on new results, always keeping a given main target. In the lab, for instance, they can test different conditions to find the optimal ones. In simulations, they can conduct automatic decisions in a given theoretical/computational context or protocol. This can save time and reduce the need for manual work. These smart systems could speed up both experimental discovery and theoretical studies

    \item \textbf{Synthetic planning}: The synthon approach pioneered by Nobel laureate E.~J.~Corey has been recently implemented by the NSF \gls{CCAS}. This is a classic example of structured reasoning in chemistry that lends itself naturally to AI integration. By breaking down complex molecules into simpler synthons and identifying viable synthetic equivalents, \gls{CCAS} mimics how a human chemist might plan a synthesis. AI models can build on this logic to rapidly explore large chemical spaces, propose novel synthetic routes, and even adapt plans in response to constraints like cost, availability, or environmental impact. 

\end{itemize}

\subsubsection{Opportunities in AI+CHE}
\label{sec:che-opportunities}

In terms of the cross-cutting techniques described in \Sec{sec:techniques}, high-priority opportunities for CHE include: 

\begin{itemize}
    \item \textbf{\glsreset{SBI}\Gls{SBI}:} \Gls{SBI} combines detailed simulations with AI to extract hidden parameters or mechanisms directly from experimental data. While it has been used in things like force field tuning or reaction kinetics, its real promise lies ahead---helping make simulations adaptive and predictive as new data comes in. If developed well, \gls{SBI} could make solving inverse problems, like figuring out mechanisms or structures from messy, often statistically unreliable measurements, far more routine.
    
    \item \textbf{Multi-scale simulations:} Multi-scale in chemistry means more than just bridging length scales---it also means connecting different timescales, from fast bond vibrations to slow conformational transitions. While there has been real progress, like using machine-learned potentials to extend quantum accuracy into longer molecular dynamics simulations, major challenges remain. AI has shown promise in areas like enhanced sampling, where it helps explore rare events more efficiently, but more automation is needed to make sampling strategies adaptive and transferable across systems. Looking ahead, combining coarse-grained modeling for spatial resolution with AI-driven enhanced sampling for temporal acceleration could finally make full-system simulations---from electronic structure to macroscopic function---feasible and predictive.
    
    \item \textbf{\glsreset{UQ}\Gls{UQ}:} Uncertainty quantification is becoming essential as AI takes on more predictive roles in chemistry. We need to know not just what a model predicts, but how confident it is especially in domains far from training data set. In the near term, tools like model ensembles and Bayesian networks can help identify where predictions are shaky and guide experiments accordingly. Long term, the goal is to build research pipelines that adaptively reduce uncertainty, making AI more useful and trustworthy in real-world discovery.
    
    \item \textbf{Foundation models:} Foundation models have shown promise in handling a range of chemistry tasks, from property prediction to retrosynthesis and even literature mining. Fine-tuning them for synthesis, materials, or protein design could streamline workflows, but questions remain about how well they generalize to out-of-distribution chemistry or rare reactions. The longer-term opportunity is to build models that can reason across experimental data, simulations, and text—but we are still far from that level of reliability or interpretability.
    
    \item \textbf{AI for experimental control:} AI-driven experimental control is enabling autonomous, adaptive laboratories where algorithms plan and adjust experiments in real time, as shown by recent self-driving lab demonstrations (from AI-guided organic synthesis to autonomous materials screening in both academic and industrial settings) that optimized reaction conditions and achieved successful syntheses with minimal human intervention. In the near term, integrating AI controllers with robotics and advanced sensors will allow experiments such as multi-step chemical syntheses or high-throughput materials discovery to self-optimize based on interim results, dramatically accelerating research cycles. Ultimately, fully autonomous laboratories could execute complex research programs---formulating hypotheses, running iterative experiments, and interpreting results---vastly speeding discovery in fields from drug development to energy storage, beyond the pace of human-guided work. It remains to be seen how much the role of human-in-the-loop can be minimized or even eliminated, and if that is something worthy of pursuing due to possible hazards with synthesizing chemicals that could be dangerous.   In particular, it will be important to bear in mind the requirement for nucleic acid synthesis screening and ``safety-by-design'' protocols in federally-funded automated laboratories.
    
    \item \textbf{Data-efficient methods:} Data-efficient AI methods in chemistry aim to extract insight from limited experimental or simulation data by using prior knowledge, simulations, and learning strategies such as transfer learning, active learning, and few-shot modeling. While these approaches have enabled promising results in areas like reaction yield prediction or small-molecule property estimation, questions remain about their ability to generalize to fundamentally novel chemistry—especially when encountering mechanisms or structural motifs absent from training data and in environmental conditions different from those in the training data set. In the near term, integrating physics-based priors, better uncertainty quantification and statistical physics frameworks may improve reliability, but realizing true data efficiency for out-of-distribution discovery remains an open and critical challenge for the field.
\end{itemize}

For this domain, additional priorities for future use of AI tools include increasing the accessibility of these tools, creating more high-quality training data, and creating new, domain-informed models. Possible examples are as follows. 
\begin{itemize}
    \item \textbf{Develop models for new areas}, such as new \glspl{LLM} for chemistry and biology that can design proteins, catalysts, small molecules, large molecules, and materials with desired functions that can be tuned for different environments, revolutionizing applications in medicine, bioenergy, sustainability, and chemical synthesis \cite{Tiwary2025}. These models should go beyond static structure paradigm to embed thermodynamic and kinetic considerations directly into the generation process. Doing so could allow for more targeted design in out-of-equilibrium and complex environments.
    \item \textbf{Accelerate computational chemistry simulations} by using AI to improve the experimental accuracy of condensed phase simulations, advance enzyme designs, and aid with self-driving labs. One key desiderata is robustness in poorly sampled regions of phase space not seen in training data sets, where rare events and long timescale dynamics often dominate the phenomena of interest.
    \item \textbf{Utilize \glspl{PINN} to tie to molecular scale mechanisms} in collaboration with continuum and fluid dynamics fields. PINNs could be especially impactful in non-equilibrium or spatially heterogeneous environments.
    \item \textbf{Develop ``chemistry-aware ML''} by training foundation models on higher-level ``languages,'' such as molecular structure in higher dimensions than a 1D text string and incorporating chemical and physical knowledge into the models itself. Such models should be grounded enough to capture reactivity and mechanism as a function of environment, not just pattern-matching based on atom connectivity. Embedding prior knowledge should reduce the reliance on massive datasets and also makes the models more interpretable in terms of representations that make sense to chemists.
    \item \textbf{Develop a foundation model} that integrates universal interatomic potentials, generative sampling, and property predictions. The aim would be to unify representation, sampling, and evaluation in a single framework—something akin to a ``molecular copilot'' capable of guiding discovery in unfamiliar chemical space.
    \item \textbf{Accelerate the rate of discovery} by incorporating methods to ensure the synthesizability of candidates into discovery workflows and/or identify synthesizable analogs. 
    Models should not just predict exciting molecules but be able to assess chemical novelty, while also suggesting how to make them under realistic lab conditions. Incorporating synthesis constraints early could prevent costly dead ends downstream.
    \item \textbf{Develop models to predict asymmetric catalysts for small molecule synthesis} that lie outside of the training set. This will likely require models that are sensitive to subtle stereochemistry and electronic effects and capable of reasoning over transition state-like features and ensembles of structures, not just limited snapshots of reactant or product structures.
    \item  \textbf{Develop improved computer-assisted synthetic planning tools} that take mechanism and transition state considerations into account when predicting new routes.
\end{itemize}

In the next 5--10 years, if the computational chemistry community can work together to build an infrastructure for obtaining large and high-quality simulation data from wave function theory and for standardizing the publication of AI methods, the long-term impact can be significant.

\subsection{Materials Research (DMR)}
\label{sec:dmr}

\glsresetall

\subsubsection{AI in the Context of DMR}

The DMR fields benefit greatly from integrating AI tools with materials domain science to \textbf{accelerate the pace of molecular and materials discovery and enable the development of new design, discovery, and engineering modalities.} There is also the opportunity, perhaps under-exploited thus far, for feedback from the domain science to the data science to inspire refinements of these AI tools, or the development of new tools and approaches.

In general, DMR research, like CHE (\Sec{sec:che}), operates in the small-data regime. The reason for this is that materials research is exceedingly heterogeneous, with research spanning from perovskites to transporter proteins, and the interesting materials are---almost by definition---rare, meaning that there is sparse data in their vicinity. Furthermore, the details of composition, microstructure, and (in many cases) processing history are all important for understanding and predicting materials properties. Soft materials must also contend with entropic considerations, wherein the properties of the material are not dictated by a single structure but an ensemble of configurations. Furthermore, interesting materials with desirable properties may not be at equilibrium, meaning that it is often important to account for kinetic considerations and dynamical processing in AI/ML models. This can also mean incorporating factors from actual growth, synthesis, and fabrication processes---such as temperature, pressure, and fabrication resolution constraints---to better reflect real-world scenarios.

Materials data is also often uncertain. There are many ways to measure the property of a material, and they may give inconsistent results. For example, the predicted bandgap of a semiconductor can vary wildly depending on the level of theory employed in its calculation, and the glass transition temperature of a material can depend sensitively on its processing history and measurement technique employed. It presents a challenge and an opportunity to \textbf{build robust AI/ML models in the face of uncertain and sparse data.}

An additional theme is the line between materials discovery and materials optimization. In the small data regime, models tend to be rather restricted in their range of applicability and generalizing to fundamentally new materials behaviors is challenging. In AI-enabled protein design, for example, there are large successes in engineering ``super-natural'' function by, for example, substantially increasing the melting temperature or activity of a functional enzyme, but it is a much greater challenge to ask these models to discover fundamentally new ``non-natural'' function.

\subsubsection{How AI is Advancing DMR}

In materials research, AI offers powerful tools and concepts to enable new paradigms in materials and molecular modeling, provide deeper understanding of structure-processing-property relations, and accelerate molecular and materials discovery. AI is used across DMR in a number of applications, including in the advancement of prediction, characterization, and modeling of molecule interactions, as well as in the acceleration of existing workflows. AI also offers opportunities for hypothesis generation, advanced characterization, multi-modal data analysis, more efficient explorations over high dimensional spaces using surrogate models, and uncertainty modeling. Specific achievements in materials research subdomains include the following. 
\begin{itemize}

    \item \textbf{Small molecules} have seen huge strides in using AI to predict properties (given dataset prerequisites), generate novel/valid molecules, and identify synthetic pathways.
    \item \textbf{Organic molecules and materials design} has included the development of AI techniques with a focus on energy harvesting and storing and electronics applications. AI models have been developed to predict electronic, redox, and optical properties of pi-conjugated molecules and parameters related to charge-carrier transport in organic molecular crystals. Likewise, software has been developed to collect experimental data, with the goal of enhancing reproducibility, that can also control a robotic platform developed for automated electrochemistry experiments.
    \item \textbf{Photonic materials} are using AI to revolutionize the design, analysis, and optimization of complex optical systems. This has helped advance the development of engineered photonic materials, such as metasurfaces, photonic crystals, and novel metamaterials.
    \item \textbf{Quantum materials} such as superconductors, spin liquids, and topological insulators are benefiting from AI tools in both their understanding and discovery. AI techniques are predicting new materials properties, guiding materials discovery, extracting hypotheses from databases, and automating experimental campaigns.
    \item \textbf{Polymers} have been the focus of intense study due to the challenges in adequately describing not only their composition, but also the processing variables that strongly affect their performance. To this end, there have been complementary efforts in polymer informatics to define feature spaces that allow for the rapid exploration of polymers and the development of self-driving labs that enable exploration of these processing-property relationships. Examples include battery materials, electrochromic materials, and organic electronics.
    \item \textbf{Metals and alloys} are particularly amenable to high-throughput experimentation and have been the focus of efforts to combine combinatorial synthesis and AI-driven characterization. These efforts have allowed for the rapid mapping of phases in addition to determining the role of processing such as laser annealing. Further, additive manufacturing systems have been transformed into tools for high-throughput property-process exploration using autonomous experimentation with integrated characterization.
    \item \textbf{Heterogeneous catalyst} design is benefiting from large-scale databases like the Open Catalyst Project combined with AI-enabled tools to perform high-throughput virtual screening and optimization to predict novel metals, alloys, oxides, etc.\ capable of catalyzing particular reactions of interest.
    \item \textbf{Digital twins} are virtual representations of materials systems that can be updated through real-time data to represent the current system state and computationally predict functional performance or potential failure modes. AI tools are offering powerful advances in the development of digital twins, including real-time updating of the twin, and in efficient scenario generation to predict and engineer processing protocols to achieve desired states, properties, and behaviors.
    \item \textbf{\Glspl{MLIP}} have contributed to significant advancements in molecular and materials modeling including enhanced accuracy, robustness, and transferability. Universal \glspl{MLIP} may also offer a better alternative to traditional quantum, classical, or mixed quantum/classical methods for large-scale screening and simulations, elucidation of multiscale non-equilibrium phenomena relevant to synthesis and processing science such as the mechanisms and kinetics for phase transformations. Clearly there are strong intersections with CHE in this area. 
\end{itemize} 

Specific examples of AI applications spanning the DMR portfolio include, but are by no means limited to: 
\begin{itemize}
    \item \textbf{Data curation:} AI’s ability to predict and identify novel synthetic pathways have led to improved automation of molecule and materials discovery. These discoveries have also been enabled by multi-fidelity paradigms integrating computational predictions and experimental measurements. Some stand-out algorithms include inverse-designs for materials properties design and inverse folding models for proteins. 
    \item \textbf{Sampling techniques:} Development of new enhanced sampling techniques have accelerated molecular simulation to more efficiently sample configurational space to improve mechanistic understanding and molecular engineering and design.
    \item \textbf{Molecular and materials discovery and design:} Generative, data-driven design of functional molecules and novel materials has been impactful. Specific applications spanning the DMR portfolio include novel catalytic enzymes, high-affinity probes for environmental sensing and remediation, novel high-entropy alloys with high strength and hardness, and new compositions and processing protocols for polymeric materials.
    \item \textbf{Inference:} The learning of low-dimensional latent variables and latent embeddings has been used to expose emergent simplicity in complex molecular and materials systems, and has informed the development of computationally efficient and interpretable surrogate models.
\end{itemize} 

\subsubsection{How DMR is Advancing AI}

The influence of AI in advancing DMR priorities is clear, but the impact of DMR in advancing AI is comparatively underdeveloped and an area that is ripe with opportunity. In particular, there are a large class of materials problems that possess particular properties, symmetries, or idiosyncracies that cannot be well engaged by existing AI models and tools. Some examples of how DMR has influenced AI include:

\begin{itemize}
    \item \textbf{Symmetry-aware networks:} Materials research necessitates the development of invariant/equivariant neural network architectures to respect the fundamental symmetries underlying materials. In particular, in the development of \glspl{MLIP} it has been shown to be extremely beneficial in stabilizing and improving these models in the low-data regime to incorporate invariance of energies and forces to translation, equivariance of forces to rotation, and invariance of energy to permutation of identical atoms. 
    \item \textbf{Self-driving labs:} The rise of self-driving labs has spurred the integration of AI techniques and robotic control software. Frequently, this has necessitated a strong focus on robustness and incorporation of prior and constraints to assure stable unattended operation or the development of human-in-the-loop strategies for AI-guided materials discovery.
    \item \textbf{Bespoke networks:} Specific domain knowledge and understanding of molecular and materials behaviors have inspired the development of bespoke network architectures reflecting this understanding, such as the critical nature of multiple sequence alignments and pair representations in protein structure prediction in AlphaFold2 and the fundamental behaviors of electronic wavefunctions in FermiNet  and PauliNet.
    \item \textbf{Databases:} Like the Materials Project database for inorganic crystalline materials and its associated benchmarking test suite MatBench, materials databases have provided benchmarks for AI models, serving as a testbed and driver for state-of-the-art ML architectures for materials systems.
    
    \item \textbf{\Glspl{PINN}:} \Glspl{PINN} enable accurate prediction of material behavior, such as stress fields, heat transfer, diffusivity, microstructure evolution, and phase separation, from sparse or indirect data, with applications across binary alloys, ceramics, battery materials, membranes, and quantum materials.
\end{itemize}

Specific opportunities, thus-far underexplored, for DMR to influence AI/ML development to the mutual benefit of both communities include:

\begin{itemize}
    \item \textbf{Uncertainty quantification:} Methods to better deal with intrinsic aleatoric and epistemic uncertainties that plague materials data.
    \item \textbf{Symmetries and constraints in \glspl{LLM}:} Incorporation of symmetries and constraints (e.g., conservation of mass and energy) into domain-aware \glspl{LLM} used for scientific hypothesis generation
    \item \textbf{Multi-scale modeling:} Multi-scale AI techniques to bridge the physics at various materials scales. 
    \item \textbf{Multi-modality:} Multi-modal generative models capable of engaging the multifaceted property optimization (e.g., protein affinity, specificity, immunogenicity, expressibility).
    \item \textbf{Data fusion and heterogeneous integration:} Developing data fusion frameworks to integrate diverse materials datasets, from simulations and experiments to literature, which enhances data quality, reduces sparsity, and improves ML model performance; DMR can drive progress by supporting fusion algorithms tailored to materials-specific challenges such as scale differences, metadata structure, and uneven sampling.
    \item \textbf{\Gls{KG} integration:} Linking diverse datasets through \glspl{KG} enhances data fusion by capturing relationships among materials, properties, and processes; supporting standardized, open \glspl{KG} enables interoperable data ecosystems that strengthen AI models with richer context and reasoning capabilities.
    \item \textbf{Ontologies and semantic frameworks:} Formal knowledge representations that define relationships between materials concepts, enabling standardized data interpretation, interoperability, and machine reasoning, which are essential for building robust AI systems that integrate diverse materials datasets and support autonomous research tools.

\end{itemize}

\subsubsection{Opportunities in AI+DMR}
\label{sec:dmr-opportunities}

A coordinated approach to AI+Materials could have the dual impact of enabling novel scientific discoveries, and infusing AI tools with physical priors that enhance the performance of the tools themselves. 

In terms of the cross-cutting techniques described in \Sec{sec:techniques}, high-priority opportunities for DMR include: 

\begin{itemize}
    \item \textbf{\glsreset{SBI}\Gls{SBI}:} \Gls{SBI} presents a means to identify interaction parameters for molecules and materials by matching to higher-resolution (e.g., quantum mechanical) or experimental data (e.g., binding constants, rates). This can be achieved by black box optimization, introduction of regularizing biases, or through differentiable molecular simulations.
    \item \textbf{Multi-scale simulations:}  Multi-scale models are required to engage many materials problems due to inherent scale-bridging that yields the emergent properties of many materials. Good simulation tools exist at each scale, but it is not always easy to bridge between scales. Similarly, AI tools must also be capable of this scale bridging and there are key gaps and opportunities here.
    \item \textbf{\glsreset{UQ}\Gls{UQ}:} Quantification and control of uncertainties is of vital importance, particularly when operating in relatively small-data regimes. A number of mature tools to perform sensitivity analysis, Bayesian inference, and error estimation are available and can be profitably incorporated into materials engineering workflows for both mechanistic understanding, prediction, and engineering campaigns, but there is also a need and opportunity to refine and sophisticate these techniques to keep pace with new artificial intelligence and machine learning architectures.
    \item \textbf{Foundation models:} Foundation models for materials systems---models trained over large and diverse data sets that can be applied to a number of specific tasks with limited additional fine tuning---have emerged in a number of areas in recent years, including \glspl{MLIP}, molecular generation, materials synthesis, and molecular retrosynthesis. These models frequently take the form of large neural networks, frequently based on transformer-based architectures, and \glspl{LLM} are a highly visible subclass of foundational models. These models are very valuable in enabling applications to downstream prediction or generation tasks (e.g., protein design, synthesis condition optimization, molecular discovery) with relatively limited finetuning, alignment, or coupling to lightweight downstream architectures. 
    \item \textbf{AI for experimental control:} Self-driving labs integrate artificial intelligence tools with automated robotics to accelerate molecular and materials discovery and optimization. There is currently a great deal of interest in developing automated platforms that, in principle, are capable of scalable, automated, and unattended materials engineering campaigns. In practice, many workflows are more robust and effective when incorporating human-in-the-loop components, and it remains a challenge to construct generic platforms as opposed to those highly specialized to specific materials systems.
    \item \textbf{Data-efficient methods:} Materials discovery and engineering frequently operates in a small data regime. Techniques appropriate to, and making the most of, small data sets are critical to success in these applications. In particular, techniques to handle imbalanced data sets, control against overfitting, and incorporate informative priors via physics-aware modeling, transfer learning, or upstream use of foundational models.
\end{itemize}
The following domain-specific priorities would also help advance the impact of this growing field. 
\begin{itemize}
    \item \textbf{Databases:} Develop better community-supported databases for soft materials, like those that exist for hard materials, including negative examples (``dark data'') that are critical for model training. This could include developing community or institutionally-supported repositories to store and professionally maintain codes and models in a similar manner to how we treat data.
    \item \textbf{Benchmarks:} Define a suite of standard and challenging benchmarks to compare new algorithms in enhanced sampling and molecular and materials design applications. Care should be taken to ensure that the benchmarks are meaningful and representative of typical use cases and do not incentivize optimizations for the benchmark that do not translate to real-world advances. 
    \item \textbf{Domain-informed models:} Build research capacity and develop domain-specific interpretable AI models, \glspl{PINN}, and \gls{UQ} techniques that will enable researchers to answer fundamental science questions that advance our understanding of how to manipulate chemical and material systems.
    \item \textbf{Data analysis:} Combine molecular simulation methods with deep learning to develop neural network-based potentials, enhance sampling techniques, and facilitate lab design and optimization---advancing these methods of data analysis also creates the capability to identify new insights from large data-generating experiments.
    \item \textbf{Optimization:} Use AI techniques such as generative AI and \gls{RL} to efficiently navigate high-dimensional design spaces and achieve unprecedented performance metrics for broad materials classes.
    \item \textbf{Processing:} Develop robust techniques to encode processing history in addition to compositional and structural descriptions, such as physics-informed Bayesian co-optimization of materials composition and synthesis protocols.
    \item \textbf{Materials discovery and design:} Leveraging AI for the discovery and design of novel quantum materials, such as topological insulators, quantum defects, quantum spin liquids, and unconventional superconductors can also accelerate the identification of novel compounds with unique electronic, optical, and magnetic properties. 
\end{itemize}

\subsection{Mathematical Sciences (DMS)}
\label{sec:dms}

\glsresetall

\subsubsection{AI in the Context of DMS}
Mathematics and statistics form the foundational backbone of AI, supplying essential techniques, tools, and theoretical frameworks that make AI possible. Beyond serving as critical methodologies, \textbf{mathematics and statistics offer a precise language and rigorous framework for communicating, describing, analyzing, and understanding AI systems} and their diverse applications. This often makes DMS researchers the ``glue'' between fields that facilitates collaboration with other sciences and develops cross-cutting collaboration programs. Mathematics and statistics are also considered by some AI labs to be the ultimate Turing test toward \gls{AGI}.
 
The interactions between DMS and AI are a two-way street and both directions have become increasingly relevant with scaling of datasets and models. In one direction, mathematical and statistical techniques are needed to gain understanding of AI, improve current techniques, and inform the foundation and design of new AI models and paradigms. One can draw parallels to how geometry reshaped physics and Maxwell's equations transformed electromagnetic theory. In the other direction, AI allows for tackling previously intractable problems such as high-dimensional statistics and differential equations, data-driven simulation, surrogate modeling, theorem proving, and discovery of mathematical objects. Addressing mathematical challenges from the lens of understanding and improving AI can also  reshape DMS, echoing historical paradigm shifts such as those prompted by quantum mechanics a century ago and computational science half a century ago.

Numerous opportunities and challenges at this intersection underscore DMS's critical role in advancing AI and the transformative potential AI holds for reshaping DMS and all its subdisciplines.

\subsubsection{How AI is Advancing DMS}
\label{sec:AI-advancing-MPS}

AI has advanced several areas of DMS. A recurrent theme in AI for DMS research is that DMS researchers do not just use AI to discover and verify mathematical results, but also to develop tailored and often novel AI techniques that exploit problem structure and support mathematical research, which in turn can advance AI. There are  major opportunities for future breakthroughs in each of these fields as DMS researchers build and use more capable AI models and gain theoretical understanding. This can create a virtuous cycle between analyzing existing and generating new mathematical datasets. 

\begin{itemize}

    \item \textbf{AI in scientific computing:}
    Complementing theory and experiment, scientific computing has become the third pillar of science. Integrating AI has led to the development of the vibrant field of \gls{SciML} that supports emerging technologies such as digital twins by leveraging techniques from  AI,  applied mathematics, computational mathematics, statistics, and probability \cite{Keith2025}. Some key themes in this area are:

    \begin{itemize}
        \item \textbf{Discover, refine, and optimize numerical algorithms:}  AI helps with searching for novel numerical schemes, optimizing solver configurations, and learning adaptive discretizations and iterative methods. These efforts suggest a future in which AI is not merely a tool for modeling but aids algorithmic design, informed by mathematical insights from approximation theory, optimization, and numerical analysis. For example, \gls{DARPA}'s ``Mathematics for the Discovery of Algorithms and Architectures'' (DIAL) program \cite{DARPADIALD} is developing methods to discover new generalized numerical algorithms in a principled way. Similarly, their ``The Right Space'' (TRS) program \cite{DARPATRS} is seeking to create and field new tools for discovering useful mathematical transformations and establish the domain of ``algorithms for numerics for singularities.''

        \item \textbf{Data-driven modeling:} Leveraging data has become crucial when the underlying physical models are incomplete, unknown, computationally intractable, or involve sparse and noisy observations. Hybrid modeling approaches integrate fundamental physical laws with computational data and experimental measurements. These approaches increasingly leverage traditional AI tools such as \glspl{DNN}: when robustness and interpretability are key, imposing polynomial or tensor structure or using symbolic regression can yield parsimonious models of physical phenomena from observed data. Also, AI-based surrogate models approximate computationally expensive physical simulations, enabling rapid predictions and significantly reducing computational cost.

        \item \textbf{Outer-loop and many-query problems:} Accelerating simulations with AI has become valuable in decision-making contexts, including optimization, \gls{UQ}, inverse problems, data assimilation, and optimal control. Decision making often requires repeated evaluations of forward simulations of complex systems, which quickly become computationally prohibitive. Here, operator learning has emerged as an important technique, enabling efficient approximations of complex solution operators.

        \item \textbf{Characterizing and resolving singularities: } 
        \Gls{SciML} provides new avenues for characterizing and resolving singularities and other intricate solution behaviors in \glspl{PDE} that pose severe challenges for traditional numerical methods. Singularities such as shocks, discontinuities, boundary layers, and topological defects often cause instability, poor convergence, or high discretization costs in conventional solvers. In contrast, deep learning-based solvers, such as \glspl{PINN}, offer new ways to localize and capture these singular behaviors without requiring mesh refinement or prior knowledge of the singularity structure.

        \item \textbf{Tackling the curse of dimensionality:} Traditional numerical methods for \glspl{PDE}, \glspl{SDE}, and control problems generally scale poorly with problem dimension, limiting their applicability. \Glspl{DNN} have shown promise in overcoming this limitation due to their excellent approximation properties. Prominent examples include solving semilinear parabolic \glspl{PDE} using forward-backward \gls{SDE} formulations, developing generative models for \gls{UQ} and inverse problems, and \gls{DNN} approximations to Hamilton-Jacobi-Bellman equations for optimal control, mean-field games, and optimal transport.

    \end{itemize}

    \item \textbf{AI for mathematical discovery:} Formulating complex mathematical challenges as structured search tasks, optimization problems, or interactive games and using AI tools such as \gls{RL}, transformer-based models, and classification algorithms to efficiently navigate the resulting search spaces has led to new mathematical discoveries.

    \begin{itemize}
        \item \textbf{Counterexamples in combinatorics:} Identifying counterexamples to mathematical conjectures involves searching through vast combinatorial spaces. Formulating these discovery processes as games, defining suitable reward functions, and addressing them via \gls{RL} has become a compelling research direction. Applications in graph theory have uncovered counterexamples with fewer vertices than previously known manual constructions and have generated ``almost-examples,'' whose insights subsequently guided mathematicians in completing rigorous proofs. Additionally, integrating transformer-based models with classical local-search methods, such as PatternBoost, has led to novel mathematical constructions and solutions to open problems in graph theory and extremal combinatorics. These use cases underscore the substantial potential of framing mathematical challenges as games or structured search tasks. Future breakthroughs are likely, particularly as current limitations concerning interpretability, parsimony of the produced objects, and computational efficiency are addressed.

        \item \textbf{AI in knot theory:} Studying embeddings of closed loops in three-dimensional space presents significant opportunities for AI, especially in knot simplification and uncovering relationships between knot invariants. Traditional approaches often rely on exhaustive enumeration or heuristic searches, which quickly become intractable. 
        Recent advances in AI, including natural language processing, classification algorithms, and \gls{RL}, have introduced novel strategies for addressing these problems. \gls{RL}, in particular, has shown promise in generating efficient simplification paths and discovering connections between previously unrelated invariants. These techniques have also been applied to the UNKNOT problem, which asks whether a given knot is equivalent to the trivial (unknotted) loop, a decision problem known to be in NP and notoriously difficult in practice. These developments demonstrate the potential of AI to address hard classification and decision tasks in topology by guiding the search through structured, high-dimensional solution spaces with learned representations and strategic exploration. 
    \end{itemize}

    \item \textbf{AI-assisted theorem proving:} Recent advances toward using AI tools for formal reasoning and as ``theorem prover copilots'' helps mathematicians formalize proofs and enables interactive proof verification, such as in co-pilots for Lean and advances in SAT solvers \cite{Yang2024}. The use of proof assistants can reveal deep mathematical connections and has, for example, led to new logical insight about the notion of equality and unexpected connections between equality and the field of homotopy theory. 
    Homotopy type theory is considered a relevant complement to existing axiomatic foundations of proof assistants, especially for proofs involving diagrams or higher dimensional structures.

    \item \textbf{AI in mathematical biology:} AI has revolutionized  protein structure prediction and offers powerful tools for modeling complex biological systems. While traditional computational methods such as homology modeling, protein threading, and \textit{ab initio} modeling have long been used to predict protein structures from amino acid sequences, they faced challenges in accuracy, scalability, and generalization to novel proteins. AI, especially deep learning,  enable efficient search of high-dimensional parameter spaces and extract patterns from massive sequence databases, which dramatically improves predictive power and robustness. State-of-the-art models such as AlphaFold2 \cite{AlphaFold2}, RoseTTAFold \cite{RoseTTAFold}, and ProteinBERT \cite{ProteinBERT} have set new standards for protein structure prediction. Notably, AlphaFold2 achieved near-experimental accuracy in the CASP14 competition and demonstrated utility across the life sciences. These models enable a deeper understanding of protein function, molecular interactions, and therapeutic target identification by reliably inferring 3D structures, and are critical to interpreting the vast amounts of sequence data generated by large-scale genomics and proteomics initiatives.

\end{itemize}

\subsubsection{How DMS is Advancing AI}

Mathematics and statistics have been fundamental for the development of AI. The main techniques behind modern AI tools can be explicitly traced back to ideas developed by mathematicians and statisticians throughout history. 
Moreover, mathematics and statistics are the key to understanding the properties and behavior of existing AI models and to improve them. Current funding support mechanisms include the NSF-Simons \gls{MoDL} program and the NSF \gls{MFAI} program. 
There are several fundamental techniques and principles from DMS that have been enabling and advancing AI:

\begin{itemize}
    \item \textbf{Training dynamics and optimization:} Almost all learning tasks are formulated as optimization problems. Consequently, modern mathematical tools  resulting from DMS research have become essential for understanding and enhancing the training dynamics of \glspl{DNN}.
First-order optimization methods, including stochastic gradient descent, momentum-based variants, and adaptive algorithms such as Adam, build upon classical optimization research \cite{Bottou2018}.

Despite these advancements, significant practical challenges persist, such as tailoring optimization algorithms to specific neural architectures or tasks, ensuring convergence in non-convex and high-dimensional landscapes, and utilizing problem-specific structures to accelerate learning. DMS research plays a critical role by synthesizing numerical optimization, high-dimensional probability, and algorithmic theory to address these issues.

Training dynamics in large-scale neural network models often exhibit behavior similar to gradient flows in high-dimensional parameter spaces. Particularly in infinite-width or large-data regimes, these dynamics can be rigorously modeled using \glspl{PDE} and variational principles. For example, the training evolution of weights in single-layer neural networks has been precisely characterized as Wasserstein gradient flows in the space of probability measures.

DMS researchers have been extending these rules to more structured neural network architectures by leveraging tools from mean-field theory, interacting particle systems, and optimal transport. These theoretical frameworks offer systematic ways to analyze convergence properties, generalization capabilities, and the intricate geometry of loss landscapes. This research not only improves theoretical understanding, but also informs the development of novel training algorithms, refined scaling laws, and improved hyperparameter tuning methodologies.

The intersection of \gls{PDE} theory, optimization, and training dynamics exemplifies how rigorous theoretical insights can lead to impactful algorithmic innovations.
    
       \item \textbf{Mathematical foundations of AI models:}
Key challenges in training \glspl{DNN} include designing a network architecture,  randomly initializing its weights, and selecting hyperparameters such as learning rates, batch sizes, momentum coefficients, and regularization parameters.
These choices influence the numerical stability of learning and the convergence speed of optimization techniques, and are critical for generalization.

Building this foundation requires sophisticated tools from mathematics and statistics, and research in many DMS areas focuses on developing theoretical guidelines and model selection techniques that can guide practical choices. 
For example, viewing deep networks, especially ResNets and diffusion models, as discretizations of \glspl{ODE} and \glspl{SDE} helps analyze and ensure the architecture's stability. 
In AI+Science, it is often desirable to design architectures that reflect known properties of the data, such as symmetries, conservation laws, multi-scale behavior, or sparsity. For example, some studies in physics-informed machine learning use equivariant neural networks that impose symmetries and \glspl{GNN} that use graph structures.
Also, Gaussian processes and kernel methods (e.g., \gls{NTK} and \gls{RKHS}) have provided powerful frameworks for understanding training dynamics in the infinite-width regime, creating insights into benign overfitting and double descent.
Furthermore, insights from \gls{RMT}, free probability, and multiplicative ergodic theory have helped understand the spectral properties of weight matrices and Jacobians and led to better initialization schemes. 
Moreover, understanding the implicit regularization properties of \glspl{DNN} and optimization algorithms using tools from \gls{RMT}, interacting particle systems, mean-field dynamics, \glspl{PDE}, and \glspl{SDE}, can help select hyperparameters that favor generalization.
In these cases, mathematics has directly influenced the design of modern deep learning libraries and remain central to ongoing progress in AI.

    \item \textbf{Statistical foundations of AI}: Statistics  develops methods for collecting, analyzing, interpreting, and presenting empirical data. Data science combines statistical and computational tools to address domain problems. These fields provide methods that are fundamental for scientific discovery and engineering, especially in the context of AI. These include sampling techniques, statistical inference, hypothesis testing, parameter estimation, causal inference, and \gls{UQ}. They also provide techniques to understand the fundamental limits of existing AI techniques in terms of privacy, robustness, uncertainty, and generalization.
    
    Moreover, statistics and data science provide data collection, data processing, experimental design frameworks, and other tools to understand the processes for obtaining scientific results. This includes identifying the sources of uncertainty and quantifying them throughout the data's life cycle. AI provides novel avenues for scientific discovery and statistics delivers a framework to establish the validity of the process to obtain these new scientific outcomes. 
 
    \item \textbf{Generative models}: 
     DMS research, especially in statistics, probability, and optimization, helps define, analyze, and improve generative models.

One of the earliest breakthroughs, \glspl{GAN}, was explicitly framed as a two-player minimax game, making game theory a natural mathematical language for studying their dynamics and stability. While \gls{GAN} objectives relied on divergences early on, later improvements, such as the Wasserstein \gls{GAN}, introduced optimal transport theory and the Wasserstein distance to stabilize training and provide meaningful gradients. This linked generative learning to deep results in convex analysis and geometry.

\Glspl{VAE} use tools from variational inference and Bayesian statistics to define a probabilistic generative process with latent variables. The analysis and improvement of \glspl{VAE} draw on information theory, variational calculus, and latent-variable graphical models, making them a key intersection point between statistical theory and deep learning.

Normalizing flows rely on tools from differential geometry, especially the theory of diffeomorphic maps, to construct exact likelihood models. These flows are trained using change-of-variable formulas, emphasizing the importance of continuous, invertible transformations in probabilistic modeling. Optimal transport techniques have been introduced to regularize flows and reported to improve numerical efficiency during training and sampling. 

Recent generative AI models are based on \glspl{SDE} such as diffusion processes and Langevin dynamics. These models construct a forward stochastic process (usually a diffusion process that gradually adds noise) and learn a reverse-time process (often governed by an \gls{SDE}) that maps noise back to structured data. The mathematical analysis of these models draws on stochastic calculus, Fokker-Planck equations, and measure transport theory. They are closely related to Schr\"odinger bridges, score-based modeling, and variational inference, creating a fruitful intersection of ideas from probability, \glspl{PDE}, and information theory.
\end{itemize}

\subsubsection{Opportunities in AI+DMS}
\label{sec:dms-opportunities}

AI offers tremendous opportunities to DMS researchers from all areas to solve open  mathematical and statistical problems with AI, while also contributing foundational insights into the science of AI itself. The DMS community is uniquely positioned to develop the next generation of interpretable, robust, and reliable AI tools needed for rigorous scientific simulations and discoveries. 

Crucially, the history of mathematics shows that abstract or foundational advances often find unexpected applications. Broad investments in DMS, including areas not originally motivated by AI, are likely to yield powerful insights that shape the structure and behavior of future AI systems.

Since mathematics and statistics form the language to develop, enhance, and deploy AI technologies, DMS researchers can be essential connectors in interdisciplinary teams, bridging domain expertise and computational techniques. 
There are also significant opportunities for DMS researchers to develop the cross-cutting techniques needed to advance scientific applications of AI (see \Sec{sec:techniques}):
\begin{itemize}
    \item \textbf{\glsreset{SBI}\Gls{SBI}:} DMS research plays a central role in advancing \gls{SBI}, providing critical insights into approximate Bayesian computation, likelihood-free inference, and surrogate modeling. Leveraging tools from probability, statistics, and scientific computing, these approaches are particularly valuable in scientific domains within MPS, where simulation is tractable but likelihood computations are challenging or infeasible. 
    \item \textbf{Multi-scale simulations:} DMS researchers bring substantial expertise with several types of systems exhibiting multi-scale characteristics across different dimensions, including temporal, spatial, modeling, and data. 
    For example, addressing temporal scale separation requires specialized time integrators, while spatial scale variation is often addressed by homogeneization theory or multigrid techniques. Bridging modeling scales requires effective surrogate models or upscaling, and rigorously addressing gaps in the data scale in learning is a prime challenge for statistical and probabilistic modeling. 
    
    \item \textbf{\glsreset{UQ}\Gls{UQ}:} \Gls{UQ} is a core area in statistics and applied mathematics, and DMS researchers actively advance its theoretical and computational foundations. Ongoing efforts develop rigorous frameworks for characterizing and quantifying uncertainty, designing efficient and statistically principled estimation algorithms, and modeling uncertainties throughout the data's life cycle. Since most \gls{UQ} tasks involve expensive outer-loops, they demand efficient inner-loop solvers, surrogate models, and reduced-order representations. Hence, \gls{UQ} benefits from advances in scientific machine learning. There are also growing connections between \gls{UQ} and generative modeling. Advances in \gls{UQ} also drive progress in DMS-relevant applications, including inverse problems, mathematical biology, and data assimilation. 
    \item \textbf{Foundation models:} There are opportunities for DMS researchers to use, fine-tune, and develop foundation models. In formal mathematics, \glspl{LLM} such as Minerva, Codex and OpenAI's GPT-4 have been fine-tuned for mathematical reasoning and proof generation, and systems such as DeepMind's FunSearch have shown that \glspl{LLM} can help discover novel mathematical structures. Chain-of-thought prompting and in-context learning have emerged as effective strategies to enable these models to solve increasingly complex symbolic and quantitative problems. In applied mathematics, emerging work treats families of \glspl{PDE} as a unifying domain and develops multioperator learning techniques that seek to learn from solution data and symbolic representations of the \glspl{PDE}.
    \item \textbf{AI for experimental control:}
    The growing use of \gls{RL} for experimental control is accompanied by its emerging role in mathematical discovery, theorem proving, and optimization of numerical algorithms. This leads to opportunities to advance \gls{RL} algorithms in settings characterized by sparse rewards, non-stationary environments, partial observability, and complex physical constraints. DMS research is aimed at sample-efficient algorithms, strategies for trading off exploration and exploitation, and approaches that incorporate prior knowledge. Reducing the computational cost of agentic workflows and integrating symbolic, probabilistic, and numerical reasoning are other opportunities for DMS researchers.
    
    \item \textbf{Data-efficient methods:} DMS research can provide the principles needed to design AI methods that operate effectively in applications with small, structured, yet high-dimensional and complex datasets. A related opportunity is developing improved AI techniques that learn from extremely sparse rewards, which is common in scientific discovery. Manifold learning, sparse recovery, nonlinear approximation, and concentration inequalities offer tools to understand and improve AI performance where data is limited, noisy, or expensive.

\end{itemize}

Other key opportunities for AI+DMS can be organized in the following thematic directions.

\begin{itemize}
    
\item \textbf{AI for mathematical discovery:} 
    AI can help discover and formalize mathematics. 

    \begin{itemize}
        \item \textbf{Agents for theorem proving: } 
        AI tools already help to digitize and formalize mathematics, with promising developments in discovering proof techniques and symbolic reasoning. Domain-specific AI agents could support the creation of formally verifiable solutions to major open problems, such as the Millennium Prize problems, and spark broader mathematical insights.

        \item \textbf{Discovering for mathematical objects:}  AI can also be used to perform neural search and discover  symbolic objects such as ribbons and Lyapunov functions. AI also has potential to identify patterns in mathematical data.

        \item \textbf{Rethinking statistics: } AI also enables us to rethink the classical statistical foundations. There is a major opportunity to develop scalable versions of core statistical procedures, including hypothesis testing, confidence set construction, and causal inference—that apply to modern data modalities, with theoretical guarantees and minimal modeling assumptions. These methods can extend statistical reasoning to domains where classical assumptions break down, with potential impact across science and engineering. Another major opportunity is the use of synthetic data, generated by AI, to boost statistical inference in data scarce applications and to preserve privacy.

        \item \textbf{Reasoning benchmarks: } 
         Mathematics itself is seen as a milestone toward \gls{AGI} due to the centrality of reasoning in the mathematical process.
        Industry labs such as DeepMind have incorporated such tools into frameworks, such as AlphaProof and AlphaGeometry, that have potential for mathematical discovery and their capabilities improve rapidly.
    Future AI systems will likely solve the large majority of exercises in standard graduate-level mathematics texts and even propose new ones.
    \end{itemize}

\item \textbf{Mathematical innovations for advancing AI:} 
    Mathematics and statistics are not merely support disciplines for AI; they are foundational engines of its advancement. From core algorithm design to emerging frontiers like interpretability, generalization, and abstraction, the future of AI depends critically on sustained mathematical innovation. As AI systems scale and enter scientific applications, new mathematical theories are urgently needed to ensure these systems are stable, efficient, and reliable.
This presents opportunities for researchers across the mathematical sciences to help define the next era of AI.

\begin{itemize}
    \item \textbf{Scaling laws and hyperparameter selection:} Modern \glspl{DNN}
    are often too large and expensive to train repeatedly, making correct choices for initialization, parameterization, and optimization critical. Mathematical theory, drawing from random matrix theory, dynamical systems, and mean-field analysis,  guides the design of stable initialization schemes and shaped understanding of infinite-width neural networks. As data and model size scale and training runs become increasingly expensive, establishing general rules for designing architectures and selecting hyperparameters become ever more crucial, especially for language, image, and video generation tools.
    \item \textbf{Understanding learning in overparameterized models:} \Glspl{DNN} generalize well despite being overparameterized and trained with non-convex optimization, phenomena that defy classical learning theory. There is an urgent need for new frameworks to describe what models learn, why they generalize, which distributions are learnable, and how these properties depend on specific architecture, data structure, and optimization dynamics. Questions about implicit regularization, benign overfitting, double descent also fall into this scope. Answers will support the development of AI systems that are more robust, interpretable, and predictable, which is especially crucial in scientific settings.

    \item \textbf{Probabilistic modeling and stochastic processes:} Neural networks are intrinsically probabilistic due to stochastic initialization, data sampling, and optimization. This opens new directions in probability theory, linking AI to \gls{RMT}, \glspl{SDE}, and statistical mechanics. Understanding the stochastic nature of learning systems can improve robustness, enable principled \gls{UQ}, and foster the design of more stable and calibrated models.

    \item \textbf{Functional analysis and representation learning:} A major strength of neural networks is their ability to learn nonlinear, data-dependent feature representations. This can be formalized using tools from functional analysis, where networks are viewed as performing regression or classification in data-dependent Hilbert spaces. Advancing this perspective could guide the development of new architectures, explain empirical success in structured domains (e.g., images, graphs), and lead to generalization guarantees that account for the geometry of learned features.

    \item \textbf{Weight space learning and symmetries:} 
    Neural network models are publicly available in platforms like Hugging Face through calls in the form of lists of weights. This new data modality brings interesting mathematical challenges that are related to problems in algebraic geometry and invariant theory. These include characterizing the weight space, their symmetries, scaling laws. Inferring model properties, such as test performance or generalization error from the weights. Generating model weights for specific datasets and tasks through weight manipulation (e.g.\ finetuning).
    Using data from neural network populations to identify and exploit structure in trained models.

    \item \textbf{Abstraction, compositionality, and mathematical structure in learning: }  Mathematics excels at reasoning about abstraction and compositionality, properties that remain elusive in current AI systems. Concepts from category theory, measure theory, and algebraic topology could help formalize compositional architectures and modular reasoning in machine learning. Developing these capabilities is essential for AI systems that aim to perform tasks like scientific reasoning, theorem proving, or programming.

    \item \textbf{\glsreset{TDA}\Gls{TDA}:}  
    Built upon advanced mathematical concepts from algebraic topology, \gls{TDA} reveals robust geometric and structural insights in complex, high-dimensional data \cite{Hensel2021}. Using tools such as persistent homology, simplicial complexes, and topological invariants, \gls{TDA} provides AI systems with meaningful shape-based features that remain stable under noise and deformation. These methods complement statistical and optimization-based approaches and have proven particularly effective in improving the interpretability and robustness of AI models. Recent advances further integrate \gls{TDA} directly into neural network architectures through differentiable persistence modules, guiding model training toward capturing intrinsic topological features and leading to more reliable, interpretable AI solutions.

    \item \textbf{Efficient AI algorithms:}    Improving the efficiency of AI systems by developing massive-scale scientific computing algorithms presents unique challenges for AI. DMS researchers can contribute to the design of distributed and federated learning algorithms that are both statistically and computationally optimal. Key issues include communication constraints, heterogeneity in computational resources, and siloed or non-identically and independently distributed (non-i.i.d.)\ data, all of which require new theoretical and algorithmic insights.

    \item \textbf{Digital twins: } Mathematical abstraction is also fundamental in the design of digital twins \cite{NationalAcademies2024DT}. DMS research on inverse problems, model reduction, \gls{UQ}, structural identifiability, and optimal control enriches AI tools and  supports the creation and refinement of digital twins.

    \item \textbf{Physics-informed models: } Scientific discovery is increasingly based on hybrid modeling approaches that combine ML methods with mechanistic understanding. DMS researchers are uniquely positioned to innovate AI methodologies that integrate neural networks with symbolic reasoning, numerical simulations, and domain-specific constraints. Effective hybrid models require careful balancing: embedding physical laws and enforcing symmetry properties, ensuring consistency with established theoretical knowledge, and maintaining sufficient flexibility to learn from data.
\end{itemize}

\end{itemize}

\subsection{Physics (PHY)}
\label{sec:phy}

\glsresetall

\subsubsection{AI in the Context of PHY}

To facilitate groundbreaking discoveries \textbf{in Physics, AI is being used to accelerate theoretical calculations, improve the operations of frontier experiments, and analyze and interpret rich and unique datasets.}
Physics is a broad discipline that spans a range of scales, both in the actual physical length- and time-scales of scientific interest, and in the size of collaborative efforts.
As a result, physics subfields are interfacing with AI in different ways.
Some fields have used traditional \gls{ML} for decades, while others are using AI for the first time.
Many subfields are developing custom AI tools matched to specific physics problems of interest. For example, physicists have developed new methods for anomaly detection, fast simulation using generative AI as surrogate models, and new classification and regression methods for more accurate and sensitive inference. These examples involve heavy adaptation and extension of ``off-the-shelf'' techniques for the specific physics problem at hand and creative uses of existing AI techniques informed by the physics application (e.g.\ leveraging normalizing flows for learning likelihoods used in \gls{SBI} or anomaly detection).

Here are some examples of different considerations that come into play when incorporating AI into various physics research efforts:
\begin{itemize}
    \item \textbf{Large-scale experiments}:
    Many flagship physics experiments, like the \gls{LIGO}, IceCube neutrino observatory, and the \gls{LHC}, involve multi-decade investments in technology and hundreds to thousands of collaborators.
    Many of the designs for these experiments predate the rise of modern AI techniques.
    Thus, a key challenge to incorporating AI into these experiments is achieving end-to-end deployment, where AI can coexist with legacy pipelines for operations and analyses.
    \item \textbf{Small-scale experiments}:
    Advances in physics also happen at table-top and mid-scale experiments, where there is in principle more flexibility to incorporate AI.
    Nevertheless, these experiments often involve highly specialized and bespoke devices and data types, which in turn require customized AI methods and tools.
    \item \textbf{Theoretical investigations}:
    Physics has a long tradition of using symbolic mathematical frameworks to gain insights into physical systems.
    AI techniques are often based on statistical methods, so one must find ways to translate AI insights into the kinds of symbolic expressions needed for theoretical calculations.
    \item \textbf{Computational modeling}:
    There are many situations in physics where the underlying physical laws are known, but it is computationally daunting to calculate their implications.
    AI can enhance these computational models of complex physical systems, but this needs to be done in a way that ensures that fundamental physical principles are respected.
    \item \textbf{Data analysis}:
    AI is blurring the traditional boundary between ``theorist'' and ``experimentalist,'' creating a role more accurately described as a ``data physicist.''
    Data analysis often involves inferring underlying theoretical structures from observed experimental data, and both ends of this inference chain have to be made AI ready. To do so, we are developing new data analysis tools and methods, powered by AI, shifting domain scientists from tool users to tool makers. 
    
\end{itemize}

Physics involves various unique data types and data requirements that push the boundaries of AI.
For example, some experimental applications have very low latency requirements, where AI decisions have to be made at the sub-microsecond level.
On the computational side, simulations often involve multi-scale structures with different physical effects dominant at different resolutions. 
Flagship physics experiments generate a deluge of data, which come in a myriad of formats, including time series, image-like representations, and graph-like structures, sometimes even from the same experiment.
Many physics analyses need high-quality simulations that are based on first principles and well-calibrated to match the data, which is especially challenging for small-scale experiments.
Importantly, quantum data/experiments are fundamentally different from classical data/experiments, and therefore need different AI approaches.

There is also a spectrum of interpretability requirements across physics subfields.
In general, physics thrives on understanding solutions, and physicists use various strategies to distill complex physical phenomena into tractable descriptions, including  latent structures, coarse-grained solutions, and effective descriptions.
Could AI produce new latent representations that model the data better?
If so, how could those representations be converted into more familiar mathematical models or equations with interpretable meanings? 
Even in cases where interpretability \emph{per se} is not needed, it is still necessary to validate that AI solutions are robust and assign appropriate uncertainties.

\subsubsection{How AI is Advancing PHY}

The broad spectrum of approaches and needs within the physics community have already led to innovative research in AI+Physics upon which to build \cite{Carleo2019}.
 AI has been used to advance the analysis of large datasets, enhance pattern recognition, improve the detection and removal of noise artifacts in data, and revolutionize simulations.
Below are some specific examples from each of the physics subdomains that demonstrate success interfacing with AI.
We emphasize that this is not an exhaustive list of examples, but rather case studies drawn from the workshop participants.

\begin{itemize}
    \item \textbf{Atomic, molecular, and optical physics} has applied \gls{RL} to experimental matter-wave interferometry using quantum degenerate atoms.
    In these experiments, \gls{RL} is used to control quantum gases composed of Bose-Einstein condensates confined in 3D optical lattices, enabling the implementation of atom-optic elements such as beam splitters and mirrors directly in the laboratory.
    This approach supports precision measurements using accelerometers, gyroscopes, gravimeters, and gravity gradiometers.
    It opens a new frontier for remote sensing of gravitational fields and inertial sensing, with potential applications in \gls{PNT}, as well as Earth observation from space-based platforms.

	\item \textbf{Elementary particle physics} has been a voracious user of AI for collider experiments.
    Existing modules in the collider analysis pipeline have been replaced with AI-powered alternatives, which have been delivering broad performance gains.
    Deep-learning-based AI has enabled high-dimensional statistical analyses, allowing physicists to leverage far more information than was previously possible, without relying solely on summary statistics.
    More recently, anomaly detection methods have been deployed to support the discovery of physics beyond the Standard Model.
    Anomaly detection represents a fundamental change in the way data is collected, processed, and interpreted, which would not be possible without AI.
    These efforts have fostered interdisciplinary collaboration among theorists, experimentalists, \gls{ML} researchers, and hardware specialists.
    One notable outcome is the development of real-time AI systems for large-scale experiments, where AI can now be deployed at the \gls{LHC} to process data with just 50 nanosecond latency.
    Additional uses of AI in high energy physics include event reconstruction, object identification, \gls{SBI}, neural importance sampling, surrogate modeling, full phase-space unfolding, and more \cite{HEPMLReview}.

    \item 	\textbf{Gravitational physics} is increasingly leveraging near real-time AI to detect gravitational wave signals from compact binary mergers in \gls{LIGO} data.
    Traditional simulation tools and analysis techniques often fall short of the precision and scalability needed to handle the growing data volumes from gravitational wave detectors. 
    AI-driven methods enable rapid alerts to the astronomical community for electromagnetic follow-up, facilitating the emerging field of multi-messenger astrophysics.
    This approach supports key scientific goals, including measuring the Hubble constant without relying on the cosmic distance ladder, probing the neutron star equation of state, and investigating the $r$-process responsible for forming the universe's heaviest nuclei.
    Similarly, AI and citizen science have been used to classify and characterize transient glitches in \gls{LIGO} data, which will help improve the rate and accuracy of gravitational-wave observations.

    \item \textbf{Particle astrophysics} is applying AI to data analysis and event reconstruction across a range of observatories.
    At the IceCube neutrino observatory, AI has enhanced the detection of neutrinos originating from the Milky Way's galactic plane by distinguishing them from the overwhelming background of atmospheric muons and by improving the energy resolution of reconstructed neutrino events.
    The \textit{Gaia} satellite has measured the positions and motions of approximately two billion stars, and AI has been used to identify stellar streams---structures that offer a new way to probe dark matter in the Milky Way.
    Meanwhile, the \textit{Fermi} satellite has observed a long-standing excess of gamma rays from the galactic center; \gls{SBI} techniques are being used to determine whether this excess is caused by dark matter annihilation or a population of unresolved pulsars.

    \item \textbf{Nuclear physics} is planning for the \gls{EIC}, the first major nuclear physics experiment being designed in the AI era, with AI-assisted detectors and workflows in mind.
    AI is being integrated into the detector design phase and will be applied across all stages of the experimental lifecycle.
    This approach will enable scalable, distributed, multi-objective optimization to maximize the science goals of the \gls{EIC}.
    At the same time, the nascent \gls{EIC} collaborations are using generative AI tools for summarization and decision support, where the technique of \gls{RAG} is used to ensure that the results are based on well-vetted scientific and technical documentation.

    \item \textbf{Plasma physics} is leveraging a wide range of AI techniques---including \gls{RL}, Bayesian optimization, statistical modeling, random forests, and neural networks---to enhance performance, control, and understanding of complex plasma systems.
    These applications span magnetic and inertial confinement fusion experiments, plasma-based accelerators and light sources, and space weather forecasting.
    AI is also driving the development of theoretical models by uncovering governing equations from high-fidelity simulation and experimental data.
    In addition, AI is enabling new analysis techniques for large, high-dimensional datasets, from spacecraft observations of plasmas to high-repetition-rate high-energy-density experiments.
    Differentiable simulation methods are further advancing the ability to model, interpret, and probe the complex dynamics of plasma systems.

    \item \textbf{Quantum information science} is using AI both to predict the behavior of quantum systems and improve the operations of quantum devices.
    On the theoretical side, neural networks are being used as a variational ansatz to represent many-body quantum wavefunctions.
    This has enabled the discovery of unexpected phenomena in many-body quantum systems, through techniques like quantum state tomography.
    On the experimental side, AI is enhancing the reliability and effectiveness of quantum computing systems.
    Techniques like multiparameter optimization have provided improved control functions for experimental protocols, and AI-based decoding algorithms have been used in state-of-the-art quantum error correction results.

    \item \textbf{Physics of living systems} has identified surprising connections between AI and statistical mechanics.
    As discussed below, the celebrated 2024 Nobel Prize in Physics 
    demonstrates the important relationship between statistical physics and computational neuroscience.
    More recently, the development of diffusion models---which now powers high-fidelity AI image generation---was motivated by studies of the Jarzynzki equality from statistical physics.

    \item \textbf{Mathematical physics} is leveraging AI to find structures in mathematical data and to solve problems in string theory and quantum field theory.
    Recent examples include using AI to compute of metrics on Calabi-Yau manifolds, using neural-network surrogates to find solutions to the $S$-matrix bootstrap equations, and using \glspl{LLM} to predict the structure of scattering amplitudes.
    This research overlaps with AI advances in pure and applied mathematics, where there has been progress in areas like combinatorics and knot theory (see \Sec{sec:AI-advancing-MPS}).
    
\end{itemize}

\subsubsection{How PHY is Advancing AI}
\label{sec:physics_advancing_ai}

Research in physics is helping to advance AI by developing methods to understand and improve AI algorithms.
Some of these innovations were originally driven by the need for AI to meet the necessary targets for scientific applications.
Physicists have also highlighted the ways that neural network-based computational methods can impact computational design, including protocol design and sampling.

As mentioned previously, a notable example of physics advancing AI can be seen in the 2024 Nobel Prize in Physics, which went to Hopfield \& Hinton for the development of deep learning by combining statistical physics with computational neuroscience \cite{NobelPrize2024PhysicsSummary}.
Building on that, the development of diffusion models was inspired by concepts from statistical physics.
Here are additional examples where progress has been made in physics influencing AI: 

\begin{itemize}
    \item \textbf{FastML:} The development of real-time AI for large-scale physics experiments (e.g., at the \gls{LHC} and Fermilab) has led to AI algorithms and systems with applications beyond their original scientific goals.
    This includes advances in high-throughput, low-latency, heterogeneous computing that address big data challenges across domains.
    Some of these techniques were originally developed with particle physics and multi-messenger astrophysics targets in mind, but they also have been relevant for neuroscience, where processing streaming brain activity data in real time has enabling new closed-loop experimental protocols.
    These developments have also contributed to AI systems design, including inference at the edge and hardware/algorithm co-design.

    \item \textbf{Equivariance:}  Physicists have played a key role in developing equivariant or symmetry-preserving neural networks.
    These networks embed physical symmetries---such as rotations, translations, and Lorentz boosts---into neural network architectures.
    These physics-informed architectures not only improve generalization and data efficiency but also help constrain model behavior in ways that align with physical laws.
    As a result, equivariant networks are now being adopted in broader AI contexts, from robotics to drug discovery.

    \item \textbf{Neural scaling laws}:  Understanding the scaling behavior of neural networks---specifically, how performance improves with increasing model size, dataset size, and compute---has become a powerful tool in modern AI.
    Physics-inspired approaches have helped identify regimes in which training converges to stable yet non-linear learning dynamics, enabling researchers to predict model performance and guide the efficient allocation of resources.
    These neural scaling laws provide practical insights into how to optimize architecture design and training procedures across tasks, with particular relevance for \glspl{LLM}.
    Moreover, connections to statistical mechanics and renormalization group evolution have provided theoretical perspectives on why such scaling behaviors emerge.

    \item \textbf{NN-QFT correspondence}: 
    Both \glspl{NN} and \glspl{QFT} can be viewed as ways to sample from a distribution of functions.
    This correspondence offers a promising theoretical framework to understand the inner workings of AI.
    By drawing analogies to Wilsonian effective field theory, researchers can model the behaviors of certain kinds of deep neural networks, with the goal of understanding topics such as initialization, generalization, and \gls{UQ}.

    \item \textbf{Mechanistic interpretability:}
    Physicists are making significant contributions to the emerging field of mechanistic interpretability, which aims to describe how AI models function internally---not just what they predict, but how they compute.
    This includes developing effective theories for representation learning, understanding the role of internal symmetries in neural networks, and classifying phase transitions in learning dynamics.

    \item \textbf{Alternative optimization schemes}:
    Physicists are drawing inspiration from the dynamics of physical systems to develop novel algorithms for optimizing neural network parameters.
    Traditional gradient descent can be interpreted through the lens of classical mechanics, operating in the highly overdamped regime where the loss function plays the role of potential energy.
    By generalizing this perspective and designing systems with custom kinetic and potential energy terms, researchers are creating optimization methods that can achieve faster, more stable, and more reliable convergence than standard techniques.

    \item \textbf{Alternative network architectures:}
    Standard feed-forward neural networks are based on linear matrices and non-linear activation functions.
    While highly effective at learning and highly efficient to implement on \glspl{GPU}, such neural networks are not so easy to interpret.
    Guided by the structure of scientific problems, physicists have developed alternative network architectures, including structures such as learnable activation functions, regularized gradients, and physics-motivated pooling operations.
\end{itemize}

In addition to these specific efforts, physics has a substantial role to play in understanding AI and making it more robust.
Physics can build upon the rich tradition of statistical physics, physics theory, and experimental advancements to develop and understand AI tools.
Also, physics insights, which have developed over centuries as a way to explain the laws of nature, can help enhance our understanding of AI and how it works. 

\subsubsection{Opportunities in AI+PHY}
\label{sec:phy-opportunities}

As physicists increasingly adopt AI techniques, AI+Physics research will undoubtedly be pursued through existing support mechanisms.
With additional dedicated support for AI+Physics, one could further advance physics goals while also providing benefits to MPS and AI more broadly. Some opportunities for further development in AI+Physics include incorporating strategies for uncertainty characterization and quantification, which are needed for robust inference in physics; coordinating on the deployment of AI methods for large-scale physics experiments with collaborations whose computational infrastructure might not be ready to incorporate AI algorithms; and integrating data science and computation as core parts of physics education.
Pursuing these opportunities has the potential to enhance AI tools for physics discovery, as well as galvanize AI innovation and interpretability using physics principles.

As mentioned in \Sec{sec:science-of-ai}, the Science of AI is a research direction that cuts across MPS domains.
Nevertheless, it is worth emphasizing the following research opportunity, which is not currently part of the physics portfolio:

\begin{itemize}
   \item \textbf{Physics of AI:}  The examples in \Sec{sec:physics_advancing_ai} highlight some of the ways that physicists approach understanding AI.
   This physics toolkit is complementary to approaches used in the computer science, mathematics, and statistics communities.
   Just like the Physics of Living Systems approaches biology through a different lens than biologists, a dedicated ``Physics of AI'' research thrust would approach AI problems using physics techniques and perspectives.
\end{itemize}

For domain-specific problems, the following opportunities represent priority research directions to advance the use and application of AI in the Physics domain.
We start with research directions already highlighted in \Sec{sec:techniques} as key AI techniques across MPS. 
\begin{itemize}

    \item \textbf{\glsreset{SBI}\Gls{SBI}:} 
    \Gls{SBI} is ubiquitous in physics, since complex forward simulations are often available but traditional likelihood functions are intractable.
    Advancing \Gls{SBI} methods could enable rigorous inference across domains such as cosmology, high-energy physics, and nuclear physics, particularly in high-dimensional settings where classical statistical tools fail.
    Research opportunities include developing more efficient neural likelihood estimators, uncertainty-aware inference techniques, and hybrid approaches that incorporate physics domain knowledge to improve interpretability and performance.

    \item \textbf{Multi-scale simulations:}
    Physics simulations often involve synthesizing multiple computations that describe disparate physical scales and processes.
    For example, modeling of nuclear and particle physics processes involves combining short-distance interactions of partons with long-distance interactions of hadrons, which are needed to model not only the fundamental processes of interest but also the behavior of experimental detectors.
    Replacing or augmenting these components with AI-based surrogates could both improve the computational efficiency of multi-scale models and better incorporate nuisance parameters.

    \item \textbf{\glsreset{UQ}\Gls{UQ}:}
    Developing reliable and reproducible methods for \gls{UQ} is a critical priority for AI in physics.
    It is essential to combine AI's ability to process vast, complex datasets with the physicist’s expertise in scrutinizing subtle or anomalous features in data.
    Given the longstanding emphasis on uncertainties within the physics community, there is a unique opportunity for physicists to take the lead in advancing \gls{UQ} methodologies.
    Moreover, uncertainty estimates should be a core component of decision-making processes in physics research, particularly when AI models are involved in designing experiments and protocols.

    \item \textbf{Foundation models:}
    The development of Large Physics Models---foundation models trained across diverse physics domains---represents an  opportunity to accelerate data analysis and scientific discovery in experiments such as the \gls{LHC}, \gls{LIGO}, \gls{EIC}, and beyond.
    This could be a single monolithic model (with an interface similar to a commercial general-purpose \gls{LLM}), or a system of interoperable domain-specific models (offering a higher degree of customization for varied scientific workflows).
    By capturing shared structure across datasets, these models could enable transfer learning, improve sample efficiency, and provide unified frameworks for tasks ranging from simulation and reconstruction to anomaly detection and inference.

    \item \textbf{AI for experimental control:} 
    Physicists have an opportunity to explore the use of agentic AI systems for the design, control, and autonomous operation of experiments.
    This includes both near-term applications---such as automating experimental workflows and operations at facilities like the \gls{HL-LHC}---and longer-term opportunities in the conceptual design of future experiments such as the \gls{EIC}, the \gls{LISA}, and the Higgs Factory.
    Agentic AI has the potential to accelerate discovery by enabling adaptive, data-driven experimentation and optimizing complex system performance with minimal human intervention.

    \item \textbf{Data-efficient methods:}  Some small-scale physics experiments face similar challenges to those in the CHE and DMR domains, where devices are bespoke and data sets are small.  These areas would benefit from data-efficient AI methods to make better decisions about what kind of data to collect and to perform more reliable extrapolation from already acquired data.  Large-scale experiments face a deluge of data overall, but would still benefit from data-efficient methods to handle rare processes, especially far-tail and edge cases that are difficult to simulate.
\end{itemize}
Next are research opportunities that are more specific to the Physics domain:
\begin{itemize}
    \item \textbf{Scaling proofs of concept:}
    There have been many proof of concept studies of possible applications of AI in physics, but there is a gap from these methods to full-scale production.
    Streamlining and improving the efficiency of turning candidate AI methods into deployable tools should therefore be a major priority.
    This includes developing more realistic datasets and benchmarks, creating better incentives for building production-grade tools, and fostering more efficient workflows within large scientific collaborations.
    It is also critical to integrate cutting-edge AI methods into upcoming large-scale experiments---such as the \gls{HL-LHC}, Rubin Observatory, \gls{LIGO}, the \gls{DUNE}, and IceCube---including applications like real-time AI, high-dimensional deconvolution, \gls{SBI}, and anomaly detection.

	\item \textbf{Solving inverse problems}:
    Recent advances in AI have led to powerful new tools for solving inverse problems, which are central to many areas of experimental and observational physics. 
    These problems, often high-dimensional and model-based, are notoriously challenging and often ill-posed statistically.
    For sampled data, such as from collider experiments, there has been progress using AI for full phase-space unfolding.
    More generally, automatic differentiation frameworks now make it feasible to apply gradient-based techniques to efficiently infer underlying physical parameters from data.

    \item \textbf{High-fidelity fast simulations:} Fast, high-fidelity simulations are critical across physics domains---for example, enabling rapid and accurate modeling of detector responses in nuclear and particle physics.
    These simulations support scalable generation of synthetic datasets for training and benchmarking AI models in tasks like particle identification, and allow for systematic studies of performance as a function of dataset size.
    Open-sourcing these tools would empower the community to create large, tunable, high-quality datasets, which are essential for probing model generalization, robustness, and representation capacity.

    \item \textbf{Building effective theories:}
    AI can be used to build effective descriptions of physical systems directly from experimental or simulated data.
    A key priority is developing physics-aware AI that respects underlying symmetries, conservation laws, and consistency conditions.
    These effective descriptions could be used to supplement high-fidelity simulations (see previous bullet), capturing multi-scale dynamics and complex detector responses more efficiently, thereby accelerating both theory development and experimental design.

    \item \textbf{Hybrid modeling of complex systems}: Many areas of physics---such as astrophysics, plasma physics, and fluid dynamics---rely on simulations driven by \glspl{PDE} to model complex systems.
    Integrating these mechanistic, interpretable models with data-driven components offers a powerful path forward. 
    Such hybrid systems can address long-standing challenges by bridging gaps between theory and experiment, while preserving the rigor and quantitative precision of mathematical physics.

    \item \textbf{Theoretical frameworks for AI use:}
    Not all physics problems will benefit from the use of AI.
    Developing theoretical frameworks to understand when and how AI can accelerate physics discovery is essential.
    This includes defining the limits of AI models, assessing the reliability of their predictions---particularly through rigorous \gls{UQ}---and identifying where AI is most likely to provide meaningful impact.
    Such frameworks would help prioritize AI integration across physics workflows and guide the development of scientifically grounded applications.

    \item \textbf{AI for Quantum 2.0:}
    Integrating AI into the Quantum 2.0 revolution presents a critical opportunity to accelerate progress in quantum technologies.
    Quantum devices---including both measurement-specific quantum sensors and general-purpose quantum computers---are poised to change the way data is collected, analyzed, and interpreted.
    The current state of the field is often referred to as the \gls{NISQ} era, where quantum devices are good enough to be useful but not so good as to be robustly fault tolerant. For the \gls{NISQ} era and beyond, it is likely that the best performance will be achieved from hybrid quantum/classical devices, where quantum measurements are controlled by classical computers. AI has the potential to further improve performance, since instead of having to directly program \gls{NISQ} devices, one could use AI to learn better protocols.
\end{itemize}
Of course, AI is a rapidly evolving field, as is AI+Physics, so these priorities are expected to evolve in the coming years.

\section{Summary and Conclusions}
\label{sec:summary}

\glsresetall

This white paper has demonstrated that the opportunities in AI+MPS are vast, so now is the time to ensure that the MPS domains capitalize on, and contribute to, the future of AI.
AI has already been used in MPS to accelerate scientific discoveries, though admittedly many of these advances could have been achieved using traditional tools given enough time.
Nevertheless, we anticipate that these initial AI developments will be harbingers for bigger breakthroughs to come, since fundamental research often advances through incremental progress punctuated by serendipitous revolutions.
Eventually, AI could achieve a real moon-shot in an MPS domain, so we encourage MPS researchers to keep sight of the big questions that AI (and perhaps only AI) could potentially answer.

Based on the insights and suggestions presented in this document, we suggest the following as the highest priority, big picture actions to be taken by funding agencies, educational institutions, and individual researchers. 
\begin{itemize}
    \item \textbf{Funding Agencies:}
    Through solicitations and strategic planning, funding agencies play a crucial role in helping shape the direction of future research and providing guidance to the scientific community about outcomes that will be most impactful. To best advance AI+MPS research in the coming years, funding agencies can:
    \begin{itemize}
        \item Support interdisciplinary projects, institutes, and researchers, creating opportunities to bring AI+MPS researchers together (\Secs{sec:funding}{sec:facilitating}); 
        \item Centralize and scale access to computing resources and data management as much as possible (\Sec{sec:infrastructure}); and
        \item Fund research in high priority, cross-cutting areas, including the Science of AI (\Secs{sec:science-of-ai}{sec:techniques}) as well as support domain-specific AI innovations in MPS  (\Sec{sec:domains}).
    \end{itemize}
    \item \textbf{Educational Institutions:} AI has already taken a foothold in education, and educational institutions will play an important role in optimizing the integration of AI into teaching, learning, and training. In order to cultivate innovative, interdisciplinary AI+MPS talent, educational institutions should aim to: 
    \begin{itemize}
        \item Hire and support interdisciplinary researchers who are building connections across domain sciences and with AI (\Secs{sec:faculty}{sec:postdocs});
        \item Establish AI+Science degree programs, including developing modular content for both student education and faculty training (\Secss{sec:graduate}{sec:undergrad}{sec:public}); and
        \item Build relationships with industry, such as through sabbatical exchanges and internship programs, for both knowledge transfer and professional development for early-career researchers (\Secs{sec:faculty}{sec:graduate}).
    \end{itemize}
    \item \textbf{Individual Researchers:} At the heart of scientific discovery and AI innovation are individual researchers---who are directly supported by funding agencies and educational institutions---as they explore and develop new tools, build connections across disciplines, and apply scientific principles to understanding AI. As more researchers move toward conducting research in the burgeoning AI+MPS domain, it will be important for them to:
    \begin{itemize}
        \item Leverage scientific insights to improve, understand, and rigorously test AI tools (\Secs{sec:science-of-ai}{sec:empower-ai});
        \item Explore opportunities for using both off-the-shelf AI tools and bespoke, domain-specific tools (\Secss{sec:techniques}{sec:conducting-research}{sec:domains}); and
        \item Collaborate across disciplines to broaden applications, reduce redundancy, and accelerate scientific discovery (\Sec{sec:facilitating}).
    \end{itemize}
\end{itemize}

The virtuous cycle of AI+MPS has the potential to be transformative for both---offering insight into AI, accelerating the pace of scientific discovery, and developing robust science and AI tools. By developing an intentional strategy, the MPS community is well positioned to be a leader in, and take full advantage of, the coming waves of AI.

\vspace{-0.3em}

\phantomsection
\addcontentsline{toc}{section}{Acknowledgments}
\section*{Acknowledgments}

This workshop was supported by the U.S.~National Science Foundation (NSF) under Award Number 2512945.
Thanks to Sarah Wells for help collating the several rounds of feedback we collected and editing this document. 
Additional thanks to Carol Cuesta-Lazaro and Alex Gagliano (IAIFI Fellows) for help facilitating discussions at the workshop, and Valerie Alleyne and Yesenia Ortiz (MIT Laboratory for Nuclear Science) for logistical support.
Additional funding (and advocacy) was provided by
the MIT School of Science (Nergis Mavalvala) and the MIT Departments of Chemistry (Troy Van Voorhis), Mathematics (Michel Goemans), and Physics (Deepto Chakrabarty).

\vspace{-0.3em}

\printbibliography[heading=bibintoc]

\setglossarypreamble[acronym]{\vspace{-1.2em}}

\phantomsection
\addcontentsline{toc}{section}{Acronyms}

\printglossary[type=\acronymtype]

\end{document}